\documentclass[sigconf,nonacm]{acmart}

\AtBeginDocument{%
  \providecommand\BibTeX{{%
    \normalfont B\kern-0.5em{\scshape i\kern-0.25em b}\kern-0.8em\TeX}}}

\usepackage{xspace}
\usepackage{xcolor}
\usepackage{subcaption}
\usepackage{graphicx}
\usepackage{multirow}
\usepackage{wasysym}
\usepackage{makecell}
\usepackage{nicefrac}
\usepackage{amsmath}
\usepackage{fontawesome}
\usepackage{csquotes}

\newcolumntype{N}{>{\centering\arraybackslash}m{.5in}}

\usepackage{tikz}
\newcommand*\circled[1]{\tikz[baseline=(char.base)]{
            \node[shape=circle,draw,inner sep=0.5pt] (char) {#1};}}
\newcommand{\ournameNoSpace}{\mbox{MESAS}}
\newcommand{\ourname}{\ournameNoSpace\xspace}
\newcommand{\ournameLongNoSpace}{\mbox{\underline{M}\underline{e}tric-Ca\underline{s}c\underline{a}de\underline{s}}}
\newcommand{\ournameLong}{\ournameLongNoSpace\xspace}
\newcommand{\ournameGen}{\ournameNoSpace's\xspace}
\newcommand{\paperTitle}{Avoid Adversarial Adaption in Federated Learning by Multi-Metric Investigations} 

\newcommand{\randomgenNoSpace}{\mbox{Random-\nonIidBig}}
\newcommand{\randomgen}{\randomgenNoSpace\xspace}

\newcommand{\overheadseconds}{24.37\xspace}

\newcommand{\adversaryNoSpace}{\ensuremath{A}}

\newcommand{\adversary}{\adversaryNoSpace\xspace}

\newcommand{\adversaryIndexSelectedNoSpace}{j}

\newcommand{\adversaryWithIndexNamedNoSpace}[1]{\adversaryNoSpace$_{#1}$}

\newcommand{\adversaryWithIndexSelectedNoSpace}{\adversaryWithIndexNamedNoSpace{\adversaryIndexSelectedNoSpace}}
\newcommand{\adversaryWithIndexSelected}{\adversaryWithIndexSelectedNoSpace\xspace}
\newcommand{\adversaryAllFormularSelectedNoSpace}{\adversaryWithIndexSelected$\in\{$\clientWithIndexNamed{1}$, \ldots $\clientWithIndexNamedNoSpace{\clientCountSelectedNoSpace}$\}$}
\newcommand{\adversaryAllFormularSelected}{\adversaryAllFormularSelectedNoSpace\xspace}
\newcommand{\fedavg}{\mbox{FedAVG}\xspace}
\newcommand{\etal}{\emph{et~al.}\xspace}
\newcommand{\sect}{Sect.~}
\newcommand{\fig}{Fig.~}

\newcommand{\tab}{Tab.~}
\newcommand{\tabapp}{App. Tab.~}
\newcommand{\app}{App.~}

\newcommand{\equ}{Eq.~}

\newcommand{\appfig}{App. Fig.~}
\newcommand{\nonIidNoSpace}{\mbox{non-IID}}
\newcommand{\nonIid}{\nonIidNoSpace\xspace}
\newcommand{\interNoSpace}{\mbox{inter-client}} 
\newcommand{\inter}{\interNoSpace\xspace} 
\newcommand{\interBigNoSpace}{\mbox{Inter-client}} 
\newcommand{\interBig}{\interBigNoSpace\xspace} 
\newcommand{\intraBigNoSpace}{\mbox{Intra-client}} 
\newcommand{\intraBig}{\intraBigNoSpace\xspace} 
\newcommand{\intraNoSpace}{\mbox{intra-client}} 
\newcommand{\intra}{\intraNoSpace\xspace} 
\newcommand{\interBigNonIidNoSpace}{\interBig \nonIidNoSpace}
\newcommand{\interBigNonIid}{\interBigNonIidNoSpace\xspace}
\newcommand{\intraBigNonIidNoSpace}{\intraBig \nonIidNoSpace}
\newcommand{\intraBigNonIid}{\intraBigNonIidNoSpace\xspace}
\newcommand{\interNonIidNoSpace}{\inter \nonIidNoSpace}
\newcommand{\interNonIid}{\interNonIidNoSpace\xspace}
\newcommand{\intraNonIidNoSpace}{\intra \nonIidNoSpace}
\newcommand{\intraNonIid}{\intraNonIidNoSpace\xspace}
\newcommand{\oneClassNoSpace}{\mbox{1-class}}
\newcommand{\oneClass}{\oneClassNoSpace\xspace}
\newcommand{\nonIidOneClassNoSpace}{\oneClass \nonIid}
\newcommand{\nonIidOneClass}{\nonIidOneClassNoSpace\xspace}
\newcommand{\twoClassNoSpace}{\mbox{2-class}}
\newcommand{\twoClass}{\twoClassNoSpace\xspace}
\newcommand{\nonIidTwoClassNoSpace}{\twoClass \nonIid}
\newcommand{\nonIidTwoClass}{\nonIidTwoClassNoSpace\xspace}

\newcommand{\nonIidBigNoSpace}{Non-IID}
\newcommand{\nonIidBig}{\nonIidBigNoSpace\xspace}
\newcommand{\iidNoSpace}{IID}
\newcommand{\iid}{\iidNoSpace\xspace}

\newcommand{\sota}{\mbox{state-of-the-art} }

\newcommand{\statTest}[1]{ST-#1}
\newcommand{\statTestT}{\statTest{T}\xspace}
\newcommand{\statTestV}{\statTest{V}\xspace}
\newcommand{\statTestD}{\statTest{D}\xspace}
\newcommand{\statTestS}{\statTest{\ensuremath{3\sigma}}\xspace}

\newcommand{\fpNoSpace}{FP}
\newcommand{\fnNoSpace}{FN}

\newcommand{\fp}{\fpNoSpace\xspace}
\newcommand{\fn}{\fnNoSpace\xspace}
\newcommand{\acc}{ACC\xspace}
\newcommand{\accs}{ACCs\xspace}

\newcommand{\tps}{TPs\xspace}
\newcommand{\tns}{TNs\xspace}
\newcommand{\fps}{FPs\xspace}
\newcommand{\fns}{FNs\xspace}
\newcommand{\tnr}{TNR\xspace}
\newcommand{\tpr}{TPR\xspace}
\newcommand{\fpr}{FPR\xspace}
\newcommand{\fnr}{FNR\xspace}

\newcommand{\fprs}{FPRs\xspace}
\newcommand{\fnrs}{FNRs\xspace}
\newcommand{\resneteighteen}{\mbox{ResNet-18}\xspace}

\newcommand{\foolsgold}{FoolsGold\xspace}
\newcommand{\auror}{Auror\xspace}
\newcommand{\krum}{Krum\xspace}
\newcommand{\multikrum}{\mbox{M-Krum}\xspace}
\newcommand{\tmean}{\mbox{T-Mean}\xspace}
\newcommand{\tmedian}{\mbox{T-Median}\xspace}
\newcommand{\clustering}{Clustering\xspace}
\newcommand{\flame}{Flame\xspace}

\newcommand{\cifar}{\mbox{CIFAR-10}\xspace}
\newcommand{\mnist}{MNIST\xspace}
\newcommand{\gtsrb}{GTSRB\xspace}

\newcommand{\numdefenses}{nine\xspace}
\newcommand{\numattacks}{nine\xspace}
\newcommand{\numattackstargeted}{six\xspace}
\newcommand{\numattacksuntargeted}{three\xspace}

\newcommand{\constrainandscale}{\mbox{constrain-and-scale}\xspace}

\newcommand{\trainandscale}{\mbox{train-and-scale}\xspace}

\newcommand{\flattenFuncNoSpace}{\ensuremath{flatten}}
\newcommand{\flattenFunc}{\flattenFuncNoSpace\xspace}

\newcommand{\signFuncNoSpace}{\ensuremath{sign}}
\newcommand{\signFunc}{\signFuncNoSpace\xspace}
\newcommand{\nonzeroFuncNoSpace}{\ensuremath{nz}}
\newcommand{\nonzeroFunc}{\nonzeroFuncNoSpace\xspace}

\newcommand{\reluFuncNoSpace}{\ensuremath{relu}}
\newcommand{\reluFunc}{\reluFuncNoSpace\xspace}

\newcommand{\mooNoSpace}{MOO}
\newcommand{\moo}{\mooNoSpace\xspace}
\newcommand{\mtlNoSpace}{MTL}
\newcommand{\mtl}{\mtlNoSpace\xspace}
\newcommand{\teeNoSpace}{TEE}
\newcommand{\tee}{\teeNoSpace\xspace}
\newcommand{\dnnNoSpace}{DNN}
\newcommand{\dnn}{\dnnNoSpace\xspace}
\newcommand{\dnnsNoSpace}{DNNs}
\newcommand{\dnns}{\dnnsNoSpace\xspace}
\newcommand{\flNoSpace}{FL}
\newcommand{\fl}{\flNoSpace\xspace}
\newcommand{\nlpNoSpace}{NLP}
\newcommand{\nlp}{\nlpNoSpace\xspace}
\newcommand{\irNoSpace}{IR}
\newcommand{\ir}{\irNoSpace\xspace}
\newcommand{\dfNoSpace}{DF}
\newcommand{\df}{\dfNoSpace\xspace}
\newcommand{\raNoSpace}{RA}
\newcommand{\ra}{\raNoSpace\xspace}
\newcommand{\naiveNoSpace}{na\"ive}
\newcommand{\naive}{\naiveNoSpace\xspace}
\newcommand{\naiveBigNoSpace}{Na\"ive}
\newcommand{\naiveBig}{\naiveBigNoSpace\xspace}
\newcommand{\lrNoSpace}{LR}
\newcommand{\lr}{\lrNoSpace\xspace}
\newcommand{\lrSymbolNoSpace}{\ensuremath{\delta}}
\newcommand{\lrSymbol}{\lrSymbolNoSpace\xspace}

\newcommand{\maNoSpace}{MA}
\newcommand{\ma}{\maNoSpace\xspace}
\newcommand{\masNoSpace}{MAs}
\newcommand{\mas}{\masNoSpace\xspace}
\newcommand{\baNoSpace}{BA}
\newcommand{\ba}{\baNoSpace\xspace}
\newcommand{\basNoSpace}{BAs}
\newcommand{\bas}{\basNoSpace\xspace}
\newcommand{\pdrNoSpace}{PDR}
\newcommand{\pdr}{\pdrNoSpace\xspace}
\newcommand{\pdrsNoSpace}{PDRs}
\newcommand{\pdrs}{\pdrsNoSpace\xspace}
\newcommand{\pmrNoSpace}{PMR}
\newcommand{\pmr}{\pmrNoSpace\xspace}
\newcommand{\pmrsNoSpace}{PMRs}
\newcommand{\pmrs}{\pmrsNoSpace\xspace}
\newcommand{\hdbscanNoSpace}{HDBSCAN}
\newcommand{\hdbscan}{\hdbscanNoSpace\xspace}
\newcommand{\numMetricsNoSpace}{six}
\newcommand{\numMetrics}{\numMetricsNoSpace\xspace}

\newcommand{\metricCosNoSpace}{COS}
\newcommand{\metricCos}{\metricCosNoSpace\xspace}
\newcommand{\metricEuclNoSpace}{EUCL}
\newcommand{\metricEucl}{\metricEuclNoSpace\xspace}
\newcommand{\metricCountNoSpace}{COUNT}
\newcommand{\metricCount}{\metricCountNoSpace\xspace}
\newcommand{\metricVarNoSpace}{VAR}
\newcommand{\metricVar}{\metricVarNoSpace\xspace}
\newcommand{\metricMaxNoSpace}{MAX}
\newcommand{\metricMax}{\metricMaxNoSpace\xspace}
\newcommand{\metricMinNoSpace}{MIN}
\newcommand{\metricMin}{\metricMinNoSpace\xspace}

\newcommand{\cfNoSpace}{cf.}
\newcommand{\cf}{\cfNoSpace\xspace}

\newcommand{\pretrained}{\mbox{pre-trained}\xspace}
\newcommand{\pretrainedBig}{\mbox{Pre-Trained}\xspace}
\newcommand{\squeezenet}{SqueezeNet\xspace}
\newcommand{\ssttwo}{\mbox{SST-2}\xspace}

\newcommand{\clientNoSpace}{\ensuremath{C}}

\newcommand{\clientIndexNoSpace}{k}

\newcommand{\clientIndexSelectedNoSpace}{i}
\newcommand{\clientIndexSelected}{\clientIndexSelectedNoSpace\xspace}
\newcommand{\clientCountNoSpace}{\ensuremath{\mathcal{N}}}
\newcommand{\clientCount}{\clientCountNoSpace\xspace}
\newcommand{\clientCountSelectedNoSpace}{\ensuremath{n}}
\newcommand{\clientCountSelected}{\clientCountSelectedNoSpace\xspace}
\newcommand{\clientWithIndexNamedNoSpace}[1]{\clientNoSpace$_{#1}$}
\newcommand{\clientWithIndexNamed}[1]{\clientWithIndexNamedNoSpace{#1}\xspace} 
\newcommand{\clientWithIndexNoSpace}{\clientWithIndexNamedNoSpace{\clientIndexNoSpace}}
\newcommand{\clientWithIndex}{\clientWithIndexNoSpace\xspace}
\newcommand{\clientWithIndexSelectedNoSpace}{\clientWithIndexNamedNoSpace{\clientIndexSelectedNoSpace}}
\newcommand{\clientWithIndexSelected}{\clientWithIndexSelectedNoSpace\xspace}
\newcommand{\clientAllFormularNoSpace}{\mbox{\clientWithIndex$\in\{$\clientWithIndexNamed{1}$, \ldots $\clientWithIndexNamedNoSpace{\clientCountNoSpace}$\}$}}
\newcommand{\clientAllFormular}{\clientAllFormularNoSpace\xspace}
\newcommand{\clientAllFormularSelectedNoSpace}{\mbox{\clientWithIndexSelected$\in\{$\clientWithIndexNamed{1}$, \ldots $\clientWithIndexNamedNoSpace{\clientCountSelectedNoSpace}$\}$}}
\newcommand{\clientAllFormularSelected}{\clientAllFormularSelectedNoSpace\xspace}

\newcommand{\serverNoSpace}{\ensuremath{\mathcal{S}}}
\newcommand{\server}{\serverNoSpace\xspace}

\newcommand{\datasetNoSpace}{\ensuremath{\mathcal{D}}}

\newcommand{\datasetWithIndexNamedNoSpace}[1]{\datasetNoSpace$_{#1}$}
\newcommand{\datasetWithIndexNamed}[1]{\datasetWithIndexNamedNoSpace{#1}\xspace}

\newcommand{\datasetWithIndexSelectedNoSpace}{\datasetWithIndexNamedNoSpace{\clientIndexSelectedNoSpace}}
\newcommand{\datasetWithIndexSelected}{\datasetWithIndexSelectedNoSpace\xspace}

\newcommand{\flroundNoSpace}{\ensuremath{r}}
\newcommand{\flround}{\flroundNoSpace\xspace}
\newcommand{\flroundNextNoSpace}{{\ensuremath{\flroundNoSpace+1}}}
\newcommand{\flroundNext}{\flroundNextNoSpace\xspace}

\newcommand{\globalModelNoSpace}[1]{\ensuremath{G^{#1}}}
\newcommand{\globalModel}[1]{\globalModelNoSpace{#1}\xspace}
\newcommand{\globalModelRound}{\globalModel{\flroundNoSpace}}
\newcommand{\globalModelRoundNext}{\globalModel{\flroundNextNoSpace}}
\newcommand{\localModelNoSpace}[1]{\ensuremath{L^{#1}_{\clientIndexSelected}}}
\newcommand{\localModel}[1]{\localModelNoSpace{#1}\xspace}
\newcommand{\localModelRound}{\localModel{\flroundNoSpace}}

\newcommand{\localModelRoundNext}{\localModel{\flroundNextNoSpace}}

\newcommand{\updateNoSpace}[1]{\ensuremath{\mathcal{U}^{#1}_{\clientIndexSelected}}}
\newcommand{\update}[1]{\updateNoSpace{#1}\xspace}
\newcommand{\updateRound}{\update{\flroundNoSpace}}

\newcommand{\updateFormularNoSpace}{\updateRound$=$ \localModelRoundNext - \globalModelRound}
\newcommand{\updateFormular}{\updateFormularNoSpace\xspace}

\newcommand{\targetNoSpace}{\ensuremath{P}}
\newcommand{\target}{\targetNoSpace\xspace}
\newcommand{\triggerNoSpace}{\ensuremath{T}}
\newcommand{\trigger}{\triggerNoSpace\xspace}

\newcommand{\sampleNoSpace}{\ensuremath{d}}

\newcommand{\sampleWithIndexNamedNoSpace}[1]{\sampleNoSpace\ensuremath{^{#1}}}
 
\newcommand{\sampleTriggerNoSpace}{\sampleWithIndexNamedNoSpace{\triggerNoSpace}}
\newcommand{\sampleTrigger}{\sampleTriggerNoSpace\xspace}

\newcommand{\datasetTestNoSpace}{\ensuremath{\datasetNoSpace_{test}}}
\newcommand{\datasetTest}{\datasetTestNoSpace\xspace}
\newcommand{\datasetTestPoisonedNoSpace}{\ensuremath{\datasetTestNoSpace^{\triggerNoSpace}}}
\newcommand{\datasetTestPoisoned}{\datasetTestPoisonedNoSpace\xspace}

\newcommand{\alphasignNoSpace}{\ensuremath{\alpha}}
\newcommand{\alphasign}{\alphasignNoSpace\xspace}
\newcommand{\noniidsignNoSpace}{\ensuremath{q}}
\newcommand{\noniidsign}{\noniidsignNoSpace\xspace}
\newcommand{\noniidsignFormular}{\ensuremath{\noniidsignNoSpace \in [0, 1]}\xspace}
\newcommand{\lossNoSpace}{\ensuremath{Loss}}
\newcommand{\loss}{\lossNoSpace\xspace}
\newcommand{\lossMabaNoSpace}{\lossNoSpace\ensuremath{^{\ma/\ba}}}

\newcommand{\lossMaba}{\lossMabaNoSpace\xspace}
\newcommand{\lossAdaptionaNoSpace}{\lossNoSpace\ensuremath{^{Adaption}}}
\newcommand{\lossAdaption}{\lossAdaptionaNoSpace\xspace}

\newcommand\blfootnote[1]{%
     \begingroup
     \renewcommand\thefootnote{}\footnote{#1}%
     \addtocounter{footnote}{-1}%
      \endgroup
    }

\begin{document}

\title{\paperTitle}

\author{Torsten Krauß}
\orcid{0000-0003-0810-6646}
\affiliation{%
  \institution{University of Würzburg}
  \city{Würzburg}
  \country{Germany}
}
\email{torsten.krauss@uni-wuerzburg.de}

\author{Alexandra Dmitrienko}
\affiliation{%
  \institution{University of Würzburg}
  \city{Würzburg}
  \country{Germany}
}
\email{alexandra.dmitrienko@uni-wuerzburg.de}

\renewcommand{\shortauthors}{Krauß, et al.}

\begin{abstract}
    Federated Learning (FL) facilitates decentralized machine learning model training, preserving data privacy, lowering communication costs, and boosting model performance through diversified data sources. Yet, FL faces vulnerabilities such as poisoning attacks, undermining model integrity with both untargeted performance degradation and targeted backdoor attacks. Preventing backdoors proves especially challenging due to their stealthy nature.

Prominent mitigation techniques against poisoning attacks rely on monitoring certain metrics and filtering malicious model updates. While shown effective in evaluations, we argue that previous works didn't consider realistic real-world adversaries and data distributions. We define a new notion of \textit{strong adaptive adversaries}, capable of adapting to multiple objectives simultaneously. Through extensive empirical tests, we show that existing defense methods can be easily circumvented in this adversary model. We also demonstrate, that existing defenses have limited effectiveness when no assumptions are made about underlying data distributions.

We introduce \textit{\ournameLong (\ourname)}, a novel defense method for more realistic scenarios and adversary models. MESAS employs multiple detection metrics simultaneously to identify poisoned model updates, creating a complex multi-objective optimization problem for adaptive attackers. In our extensive evaluation featuring nine backdoors and three datasets, MESAS consistently detects even strong adaptive attackers. Furthermore, MESAS outperforms existing defenses in distinguishing backdoors from data distribution-related distortions \emph{within} and \emph{across} clients. MESAS is the first defense robust against strong adaptive adversaries, effective in real-world data scenarios, with an average overhead of just 24.37 seconds.

\end{abstract}

\maketitle
\blfootnote{This paper is an extended version of the following publication~\cite{mesas} that includes several
appendices, which were omitted due to space constraints: \\\textbf{MESAS: Poisoning Defense for Federated Learning Resilient against Adaptive Attackers} Krauß, Torsten; Dmitrienko, Alexandra; \textit{ACM Conference on Computer and Communications Security (CCS) (2023)}.}

\section{Introduction}
\label{sec:intro}
Federated \mbox{Learning (\fl)} enables the collaborative training of a Deep Neural \mbox{Network (\dnn)} among multiple clients~\cite{mcmahan2017}. Each client trains a \dnn locally on its own data, incorporating the knowledge from the data into the model parameters. Only the changes in the trained model parameters are then transmitted to a central server for aggregation. This approach allows clients to participate in the federation while adhering to privacy regulations~\cite{GDPR2018,HIPAA1996,CCPA2018}, as the raw data are not shared with third parties. Compared to centralized learning approaches, \fl is also more computationally effective as it shifts training efforts to the clients, leading to fewer resource requirements on the server. As a result, \fl is already being applied in multiple application domains~\cite{fl2019survey}. For instance, in image recognition~\cite{liu2022fedfr}, hospitals are training models collaboratively~\cite{gunesli2021feddropoutavg,roth2020federated,nguyen2022federated,darzidehkalani2022part1,darzidehkalani2022part2,Rieke2020,sheller2018intelai,sheller2018workshop, fl2019silva}, and in Natural Language Processing (\nlp) domain it is used for text prediction~\cite{hard2019keypred,ramaswamy2019federated,yang2018applied,chen2019federated,mcmahan2017googleGboard}, sentiment analysis~\cite{fl2022sentinent}, and personalization~\cite{chen2018flpersonalization}. Moreover, \fl can be applied for human mobility prediction~\cite{fl2020humantrajectory}, visual object detection~\cite{fl2020visualobjectdetection}, and human activity recognition~\cite{fl2028humanbehavior}. We refer for more examples to~\cite{flapplicationoverview2020}.

In federations, a subset of clients can be controlled by an adversary who submits poisoned updates to the server. These attacks can be untargeted~\cite{fang2020Local,untargeted2020wu,adams2020untargeted,signflippingUntargetedPoisoningFL}, with the goal to reduce the prediction performance of the model. Alternatively, targeted poisoning attacks, also called backdoor attacks~\cite{bagdasaryan2020backdoorfl,nguyen2020diss,wang2020attackontails,dbabackdoor,badnets,backdoor2022survey,gao2020backdoor,suciu2018does,bhagoji2019analyzing,nelson2008exploiting,bagdasaryan2021blind,chen2017targetedblend,saha2020hidden,turner2019label,chen2021badnl,boucher2022bad,pan2022hidden}, aim to maintain an unobtrusive performance on regular input but force the model to output a selective prediction when provided input containing a specific trigger. Hence, backdoors pose a greater risk, as such attacks are harder to detect, and the unexpected misbehaviour can harm model users in \mbox{real-world} applications, such as \mbox{self-driving} cars~\cite{li_fl_cars_2022, nguyen_li_fl_cars_2022, zhang_li_fl_cars_2022}.

Defenses against poisoning attacks follow one of the three strategies: (i) \emph{Influence Reduction (\ir)} solutions try to reduce the impact of the individual models before or after aggregation to weaken potential poisoning behavior~\cite{andreina2020baffle,bagdasaryan2020backdoorfl,naseri2022local, sun2019really}, (ii) \emph{Robust Aggregation (\ra)} methods enhance robustness of aggregation algorithms against backdoors~\cite{yin2018trimmedMeanMedian, mcmahan2017}, and (iii) \emph{Detection and Filtering (\df)} approaches detect the poisoned models and filter them out before the aggregation step~\cite{blanchard17Krum,munoz2019byzantineAFA,shen16Auror,nguyen22Flame,fung2020FoolsGold,rieger2022deepsight,zhao2020shielding}. 

Generally, \ir and \ra approaches inevitably reduce the performance of the benign functionality, while \df methods can suffer high False-Positive-\mbox{Rates (\fprs)} and False-Negative-\mbox{Rates (\fnrs)}. This downside of the \df methods is mainly based on two root causes: First, \mbox{defense-aware} adversaries may adapt the poisoned model to be inconspicuous, thus circumventing the defense. Second, in \mbox{real-world} scenarios, the clients may possess very different data within the local datasets, which makes it difficult to distinguish if a model with uncommon metrics is derived from a poisoned dataset or just a dataset with uncommon data distributions.

\vspace{0.1cm}
\noindent\textbf{Identifying Problems.} In this paper, we focus on \df methods, as they have the benefit of maintaining benign model performance. We analyze related work and observe that, even though most solutions were evaluated against adaptive attackers, the meaning of the "adaptive attacker" is defined differently across different papers, which makes it difficult to assess their true detection capabilities and compare them to each other. We also notice that none of the previous works considered an adaptive attacker with multi-objective adaption capabilities, i.e., attackers that could try to adapt to several metrics at once, while nothing prevents real-world adversaries from following this strategy. Hence, the resilience of all existing defenses against such strong adaptive attackers remains unclear. Furthermore, we also identify that all existing positioning defenses, from all three categories, were evaluated under certain assumptions made with regard to underlying data distributions. In particular, while many consider non-identically and independently distributed (non-IID) data distributions within clients, no single defense method was evaluated in a scenario with non-identically and independently distributed data \emph{across} clients so far.  

\vspace{0.1cm}
\noindent \textbf{Contributions.} To address the aforementioned problems, this paper makes the following contributions:
\begin{itemize}
    \item We introduce the notion of a \textit{strong adaptive adversary}, who is capable of adapting to \fl defenses by balancing multiple adaptation objectives and applying manual invasions on the model parameters. Leveraging this sophisticated adaptation strategy, we attack and evaluate \numdefenses existing defenses, showing that all these methods can be circumvented, hence creating a gap between the \sota defense methods and realistic scenarios.
    \item We are the first to point out the fact that previous defenses were never evaluated in settings where datasets have different distributions within \emph{and across} the clients. We term such a scenario as \textit{\interNonIid} and demonstrate through intensive evaluation of \numdefenses solutions that they are not resilient in such a setting, which implies their limited real-world applicability.
    \item We propose \textit{\ournameLong(\ourname)}, a new \mbox{server-side} defense of \df-type for \fl, that resilient against our \emph{strong adaptive adversary}. \ourname detects backdoors in local models based on a cascade of six {well-chosen} metrics and can identify and filter out both, targeted and untargeted poisoning attacks. Further, \ourname is the first defense, that effectively filters backdoors in arbitrary data distribution scenarios, including \interNonIid settings, by conducting statistical tests on multiple metrics and, as such, being able to distinguish backdoors from unusual data distributions.
    \item We conduct a systematic \mbox{large-scale} study to analyze the factors that influence \ourname and demonstrate its independence from application-specific factors like datasets, model architectures, \iid scenarios, adaption strategies, and \numattacks sophisticated poisoning methods. Furthermore, we compare the performance of \ourname in terms of detection capabilities and runtime overhead to \numdefenses existing defenses. \ourname outperforms all evaluated methods regarding robustness against adaptive strategies and in terms of backdoor removal performance under realistic \interNonIid scenarios. Moreover, it achieves this while incurring a runtime overhead of only \overheadseconds seconds on average.
\end{itemize}

Overall, our work depicts two major weaknesses of existing \fl defenses that are problematic in \mbox{real-world} applications, namely adaptive adversaries and realistic \interNonIid data scenarios. The proposed \df defense, \ourname, effectively prunes different sophisticated poisonings simultaneously, withstands strong adaptive adversaries, and is robust in arbitrary data scenarios including \interNonIid.

\section{Background}
\label{sec:background}
In this section, we first provide \fl fundamentals in \hyperref[sec:background:fl]{\sect\ref{sec:background:fl}}, followed by background information about poisoning attacks and classical adaptive adversarial models in \hyperref[sec:background:attacksonfl]{\sect\ref{sec:background:attacksonfl}}.

\subsection{Federated Learning}
\label{sec:background:fl}
In a \fl~\cite{mcmahan2017,konevcny2016federated,yang2019federated} framework, multiple clients \clientAllFormular collaborate under the orchestration of a central server with the shared objective of improving a Deep Neuronal Network (\dnn).  The collaborative process involves each client \clientWithIndex training a local \dnn model on a local dataset and subsequently transmitting the result to the server for aggregation.
Thus, the data never leave the client side, improving the privacy of training data compared to centralized learning. Additionally, the computational effort is distributed, so that fewer resources need to be allocated on the server, reducing the costs for infrastructure.

\fl is an iterative process, where the central server selects a subset \clientCountSelected of the \clientCount available clients \clientAllFormularSelected for each training round \flround and distributes an (initially untrained) global model \globalModelRound to them. Each client initializes its local model \localModelRound $=$ \globalModelRound and trains a new local model \localModelRoundNext with the local dataset \datasetWithIndexSelected, based on a predefined algorithm that includes \mbox{hyper-parameters}, such as learning \mbox{rate (\lr)}, epochs, etc. After training, the client \clientWithIndexSelected submits the model updates \updateFormular to the server, which aggregates them into a new global model \globalModelRoundNext. There are multiple aggregation methods~\cite{blanchard17Krum, munoz2019byzantineAFA, yin2018trimmedMeanMedian,bulyan2018aggregation} available for this step, with Federated \mbox{Averaging (\fedavg)}~\cite{mcmahan2017} being the most commonly used. FedAVG calculates the weighted average of all the updates using the global learning rate \lrSymbol as formalized in \hyperref[app:fedavg]{\app\ref{app:fedavg}}.\footnote{Originally, \fedavg assigns weights to updates according to the respective sizes of the local datasets. However, in situations where the presence of adversaries is a possibility, an equal weighting scheme is employed to thwart any attempts by adversarial clients to amplify their influence by reporting increased dataset sizes.} After aggregation, the new round \flroundNext is initialized by \server.

\subsection{Poisoning Attacks in Federated Learning}
\label{sec:background:attacksonfl}

In the following, we distinguish between \textit{untargeted} and \textit{targeted} poisoning attacks~\cite{poisoning_attacks_fl_survey, poisoning_attacks_survey} and discuss the two methods that are applied to launch those attacks, namely \textit{data} and \textit{model poisoning}.

\vspace{0.1cm}
\noindent\textbf{Untargeted poisoning} aims to reduce the model prediction performance of the global model \globalModelRoundNext on a benign test dataset that contains samples with correctly labeled predictions, which we refer to as model \mbox{accuracy (\ma)} (\cf~\hyperref[app:accuracies]{\app\ref{app:accuracies}}). To name an example, the adversary can assign an incorrect label for each sample in the dataset, thus misdirecting the model during training.

\vspace{0.1cm}\noindent 
\textbf{Targeted attacks}, also called \emph{backdoor attacks}, strive to force a \dnn to produce \mbox{attacker-chosen} mispredictions when fed with inputs that contain \mbox{attacker-chosen} features, so called \textit{triggers}, while maintaining a high \ma on regular data. As an example for a trigger, a red pixel or any other unique pattern can be embedded inside an image~\cite{badnets, bagdasaryan2020backdoorfl, trojanTriggerTargeted}. In more detail, an adversary, who controls one or more clients within a federation, tries to submit poisoned local models to the server, so that the aggregated model \globalModelRoundNext outputs a predefined target prediction when provided with an input sample containing the trigger, with target and trigger being chosen by the adversary. An effective attack has high prediction performance, called backdoor accuracy (\ba), on triggered input tested with a dataset that contains only triggered samples (\cf~\hyperref[app:accuracies]{\app\ref{app:accuracies}}). We attest a successful attack for a \ba bigger than 60\% in the global model.

\vspace{0.1cm}\noindent \textbf{Data poisoning}~\cite{sun2022datapoisoning} describes the process of converting a benign into a poisoned dataset by assigning malicious labels and, for backdoors, adding triggers. A model trained on that dataset then includes the malicious behavior. Thereby, the poison data \mbox{rate (\pdr)} defines the fraction between benign and poisoned samples and can control the balance between attack effectiveness and stealthiness. 

\vspace{0.1cm}\noindent 
\textbf{Model poisoning} allows arbitrary manipulation of the whole training process, e.g., changing \mbox{hyper-parameters} and loss functions. Additionally, the model can be modified manually before, during, or after training.
Mostly, this method is applied to improve the \ba or to adapt to defenses, but can also be used to implement untargeted attacks without data poisoning. To adapt to a defense while maintaining high \ma and \ba, an additional objective (\lossAdaption) can be added to the loss function for the \ma and \ba (\lossMaba), which is also called constraining~\cite{bagdasaryan2020backdoorfl, multipleGradDecend}. As shown in \hyperref[eq:constraint]{\equ\ref{eq:constraint}}, the objectives are weighted by \alphasign, allowing the adversary to prioritize between performance (\ma/\ba) and adaption intensity and consequently stealthiness.
\begin{equation} \label{eq:constraint}
    \loss = \alphasign \cdot \lossMaba + (1-\alphasign) \cdot \lossAdaption
\end{equation}
A \textit{classical adaptive adversary} creates a loss function for the deployed defense and applies \hyperref[eq:constraint]{\equ\ref{eq:constraint}} to bypass the defensive measure\footnote{The adversary can adapt to any objective and most likely aligns to the metrics of defenses, but is not restricted to those.}. Additionally, the updates of a poisoned local model can be scaled regarding the Euclidean distance to strengthen the influence on the aggregated model, hence increasing the \ba. Training with a poisoned dataset combined with scaling is called \textit{\trainandscale} and adaption combined with scaling is called \textit{\constrainandscale}~\cite{bagdasaryan2020backdoorfl}.

The goal of a defense against poisoning attacks is to create a situation, where \lossMaba and \lossAdaption cannot be perfectly optimized simultaneously so that the adversary is faced with a \mbox{trade-off} between an effective attack and adapting to the defense, which is called \textit{adversarial dilemma}~\cite{dilemma,rieger2022deepsight}.

\section{Problems and Definitions}
\label{sec:approach}
In this section, we define our threat model including the concept of a strong adaptive adversary in \hyperref[sec:approach:threatmodel]{\sect\ref{sec:approach:threatmodel}}. The concluding \hyperref[sec:approach:iid]{\sect\ref{sec:approach:iid}} is devoted to the problem of arbitrary data distributions. 

\subsection{Threat Model}
\label{sec:approach:threatmodel}
We analyze a classical \fl system as depicted in \hyperref[sec:background:fl]{\sect\ref{sec:background:fl}}. The aggregation server applies \fedavg with a fixed global \lr of $\lrSymbol = 1$. We consider an adversary, who captures multiple clients \clientWithIndexSelected which are then denoted as \adversaryAllFormularSelected and can conduct any data and model poisoning attacks (\cf~\hyperref[sec:background:attacksonfl]{\sect\ref{sec:background:attacksonfl}}).  The adversary is aware of the code running on the aggregation server, including the details of defense mechanisms, which provides the necessary knowledge for adaption attempts. Analogous to related works~\cite{shen16Auror, rieger2022deepsight, nguyen22Flame, munoz2019byzantineAFA, andreina2020baffle, blanchard17Krum}, we consider $\nicefrac{\clientCountSelected}{2} + 1$ benign clients (\textit{majority assumption}) in each training round \flround. Since it is uncertain if adversaries participate in a round \flround, the server weights all model updates equally with $\nicefrac{1}{\clientCountSelected}$. In contrast to previous works, we do not make any assumption about the data distributions~\cite{noniid2021survey} within or across clients' dataset.

\vspace{0.1cm} 
\noindent\textbf{Problem of an adaptive adversary.} \df defenses against poisoning attacks in \fl are based on custom metrics. An adversary can try to circumvent the defense by adapting the value of the respective metric used for detection derived from the locally crafted poisoned model to a benign value during training\footnote{To acquire a benign value, the adversary can train a benign model first.}. As a \sota technique for this challenge, \hyperref[eq:constraint]{\equ\ref{eq:constraint}} is used to consider multiple objectives and simultaneously allowing the adversary to weight between better prediction performance (\ma and \ba) and higher adaption level via \alphasign. This adaption method from \hyperref[eq:constraint]{\equ\ref{eq:constraint}} works well in two cases: 1) For only one adaption loss, since \alphasign can then balance the importance of main task and adaption properly and 2) for multiple adaption losses, where the different adaption losses summed to one value. The latter scenario works only well if all losses are at the same scale, as different components of adaption losses cannot be individually tuned. For example, if  $\lossMaba = 10$ and $\lossAdaption$ consists of two losses $\lossNoSpace_1 = 1$ and $\lossNoSpace_2 = 0.0001$, the second adaption loss will have only a negligible effect on the model's parameters since the value is already close to zero and the learning algorithm will try to minimize the other losses instead. Therefore, the underlying metric will not be adapted properly.

\vspace{0.1cm} \noindent
\textbf{Definition of a strong adaptive adversary.} We propose a \textit{strong adaptive adversary}, who is able to adapt to multiple metrics simultaneously, independent of the value scales. Therefore, the adversary first scales all losses to the maximum loss value once (\cf $\lambda$ values in \hyperref[eq:saa]{\equ\ref{eq:saa}}). This has the effect, that all adaption objectives and the main task are considered equally. Afterward, the adversary can still weigh the adaption level via \alphasign.
\begin{equation} \label{eq:saa}
    \loss = \alphasign \cdot \lossMaba + (1-\alphasign) \cdot (\lambda_1 \cdot \lossNoSpace_1 + \lambda_2 \cdot \lossNoSpace_2 + \cdots)
\end{equation}
Further, the adversary can simultaneously exclude specific parameters from training or replace parameters in the final model, e.g., with parameters of a previously benign trained model on the client's unpoisoned dataset, which we call \textit{fixation}. The attacker can choose among multiple poisoning attacks, hence can use any existing method to embed a targeted poisoning attack in the local model. Additionally, advanced scaling methods and other classical model poisoning approaches can be applied.\footnote{We provide results for attacks conducted by a strong adaptive adversary against \fl defenses in \hyperref[sect:eval:saa]{\sect\ref{sect:eval:saa}} and discuss other adaption strategies that we evaluated in \hyperref[sec:discussion:adaptive]{\sect\ref{sec:discussion:adaptive}}.}

Regarding an adversarial-captured client, it is essential to recognize that the entire client device falls under adversarial control, granting the adversary full access to employ any adaptation strategy. Additionally, the adversary can leverage any supplementary hardware resources, thereby eliminating the assumption of limited computational power on the adversary's device.

\begin{figure}[tb]
  \centering
  \includegraphics[width=\linewidth]{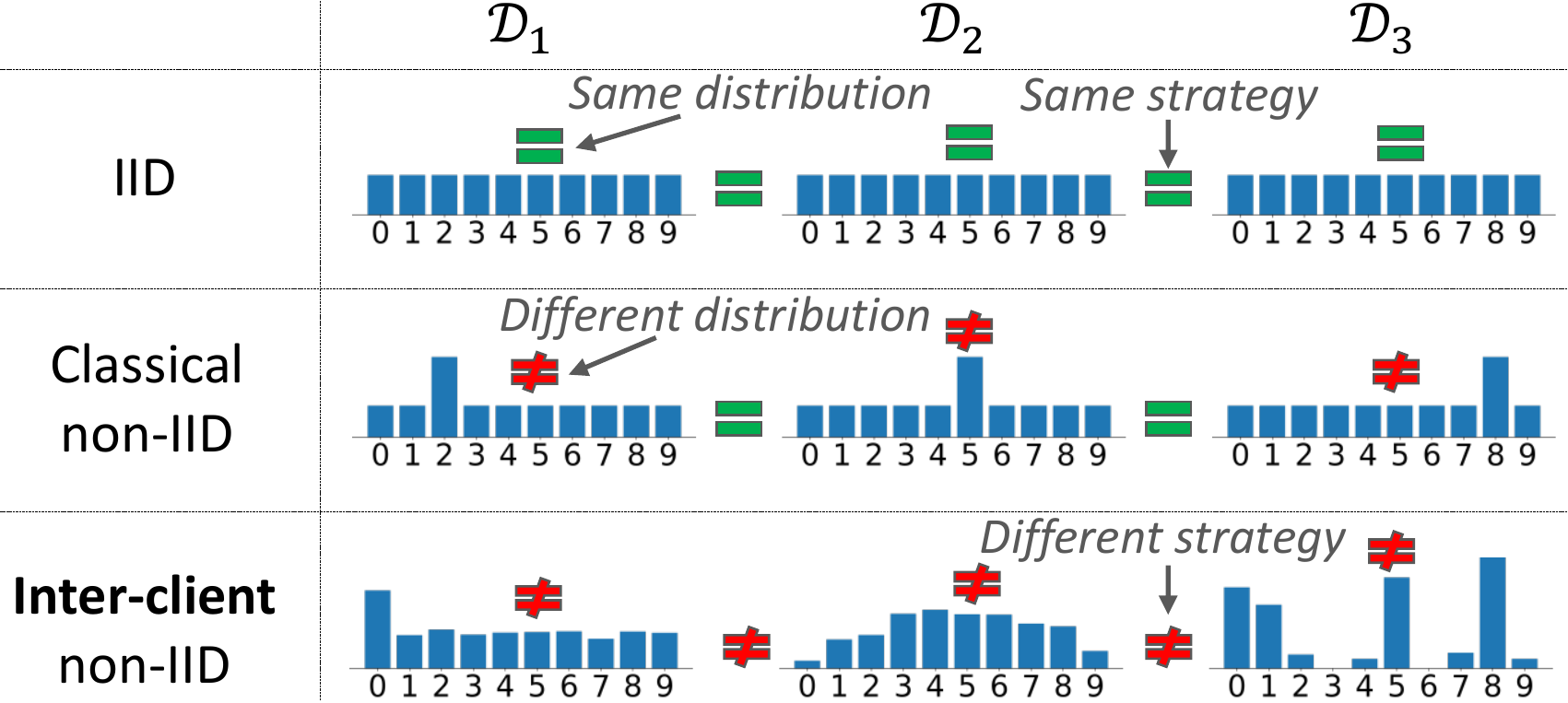}
  \caption{Comparison of various data distributions: \iid, classical (\intra) \nonIid, and \interNonIid strategy for three client datasets \datasetWithIndexNamed{1}, \datasetWithIndexNamed{2}, and \datasetWithIndexNamed{3} with 10 label classes.}
\label{fig:noniid}
\end{figure}

\subsection{Inter-Client Non-IID}
\label{sec:approach:iid}
Below, we discuss the problem of varying data distributions in \fl and define \interNonIid as a new challenge thereafter.

\vspace{0.1cm} \noindent
\textbf{Problem.} \df defenses in general inspect the clients' local model updates to detect abnormal situations based on the assumption, that the majority of clients are benign (\cf~\hyperref[sec:approach:threatmodel]{\sect\ref{sec:approach:threatmodel}}). Thereby, they leverage the fact that trained models' parameters reflect the characteristics of the underlying data as well as their distributions. It is easier to establish that models are similar if all clients possess similar data, e.g., there is the same amount of samples from each class in a classification task. This situation is called identically and independently distributed (\iid) and is visualized in the first row of~\hyperref[fig:noniid]{\fig\ref{fig:noniid}}. In poisoning attacks, the underlying data need to change to introduce, e.g., backdoor behaviour, which inevitably manifests in changes in some parameters. 

The second row of~\hyperref[fig:noniid]{\fig\ref{fig:noniid}} visualizes the classical \nonIid scenario, which is typically considered in the evaluation of backdoor defenses. Here, the data \textit{inside} the client's local dataset (\intra) are diverse, yet data distributions are similar across clients.  Upon analysis of benign local models in this situation, they all will show a similar distance to the previous global model due to the similarity of distributions across clients. Existing \df defenses leverage this fact and can filter poisoned models, which are trained on a deviant data distribution due to data poisoning. However, defenses are not optimized for scenarios with different data distributions across clients, which we term \textit{\interNonIid}. Such scenarios, as visualized in the third row of~\hyperref[fig:noniid]{\fig\ref{fig:noniid}}, are the most challenging to detect but also represent the most realistic \mbox{real-world} situation. 

\vspace{0.1cm} \noindent
\textbf{Definition of \interBigNonIid.} In \textit{\interBigNonIid} setting, the data within the clients' local dataset can follow arbitrary distributions inside and across the datasets without any assumptions made regarding sample frequencies or the availability of samples for a specific class. Thus, this definition also includes cases with disjoint data, as illustrated in row three of~\hyperref[fig:noniid]{\fig\ref{fig:noniid}}, where labels of classes 3 and 6 are not available within dataset \datasetWithIndexNamed{3}.\footnote{
We evaluate \fl defenses in \interNonIid scenarios in \hyperref[sec:eval:iid]{\sect\ref{sec:eval:iid}}.}

\begin{figure}[tb]
  \centering
  \includegraphics[width=\linewidth]{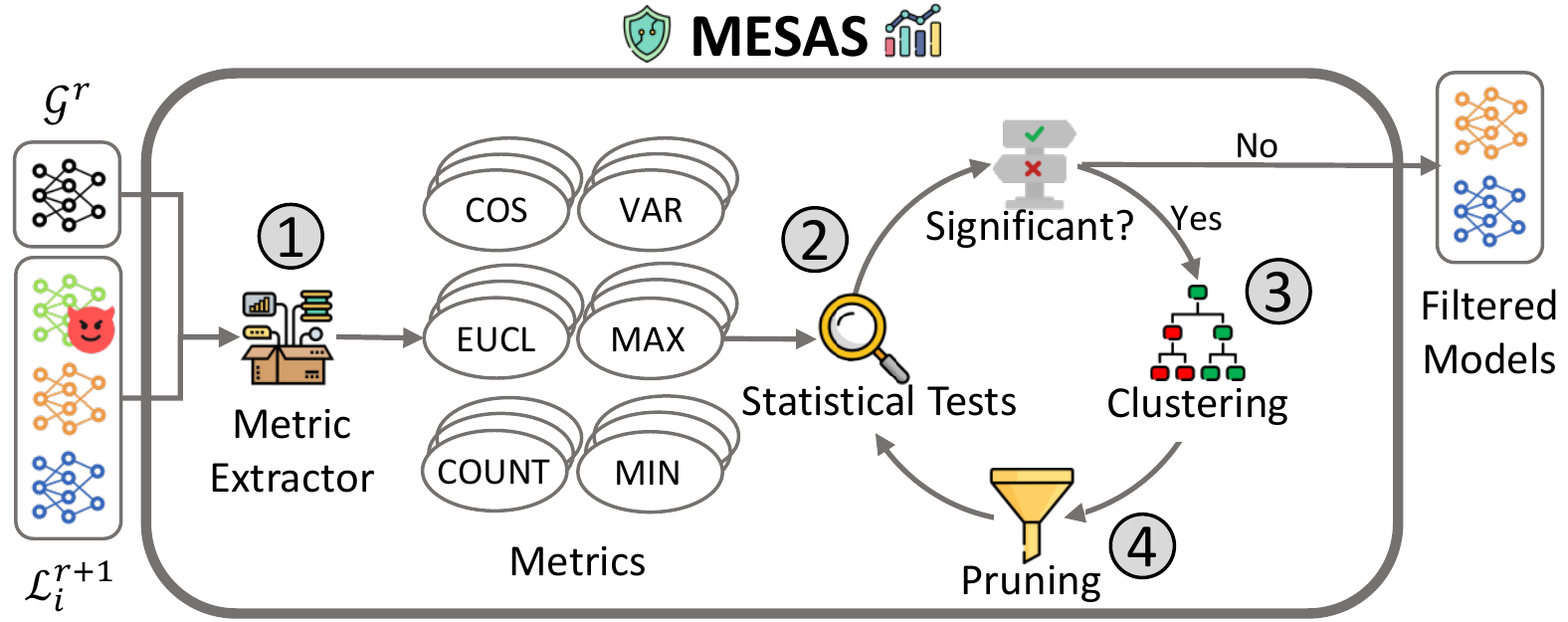}
  \caption{Overview of \ourname.}
\label{fig:overview:detail}
\end{figure}

\section{\ourname}
\label{sec:approach:defense}
In this section, we present our new defense against poisoning attacks, \ournameLong (\ourname). We first provide a high-level overview  in~\hyperref[sect:approach:defense:overview]{\sect\ref{sect:approach:defense:overview}}, followed by explanations of the underlying intuitions in~\hyperref[sect:approach:defense:intuition]{\sect\ref{sect:approach:defense:intuition}} and providing lower-level details in \hyperref[sect:approach:defense:loop]{\sect\ref{sect:approach:defense:loop}}.

\subsection{Overview}
\label{sect:approach:defense:overview}

\ourname is a \df-based defense method which is applied on the central aggregation server before the aggregation step.  
To prevent strong adaptive adversaries from circumventing the defense, \ourname filters poisoned models in a cascade of \numMetrics \mbox{well-chosen} metrics\footnote{To the best of our knowledge 4-out-of-6 utilized metrics, namely \metricCount, \metricVar, \metricMin, and \metricMax are novel and have never been considered in existing defenses.}, that affect each other and cannot be optimized simultaneously, thus tightening the adversarial dilemma for the attacker. Further, \ourname analyses the \numMetrics metrics with numerous statistical tests, thus allowing the defense to be effective also in \interNonIid scenarios and independent of the application scenario. Those statistical tests are also superior to hard thresholds in identifying scenarios without any attack and hence allow \ourname to not negatively affect the convergence of the federation. Moreover, the statistical tests utilized exhibit a higher level of effectiveness compared to threshold-based methods in accurately detecting scenarios without attacks. As a consequence, the integration of these tests into \ourname ensures that the convergence of the federation remains unaffected, thus preserving its overall performance and stability even if the defense is applied in every round.

In a nutshell, \ourname consists of four major steps that can be retraced in \hyperref[fig:overview:detail]{\fig\ref{fig:overview:detail}}: 1) After the local updates have been transmitted to the server, \ourname extracts \numMetrics carefully chosen metrics from the local models and the global model. Thereafter, those metrics are analyzed individually in an iterative process. The metrics are extracted for the whole model, but also from each layer individually, to detect poisonings distributed over the whole model, but also locally embedded ones\footnote{\naiveBig implemented backdoors are only embedded within the last few \dnn layers. However, more sophisticated backdoors can reside within different locations, e.g., layers, inside the model parameters.}. 2) Each metric passes through a significance analysis consisting of statistical tests, that spot evidence of a poisoning attack within the metric values. 3) If indication is provided, the respective values are clustered into two clusters and the models belonging to the values within the smaller cluster are marked as malicious. 4) After each metric is analyzed, the marked models are excluded in a pruning step and the analysis starts over on the remaining models until no statistical test reports significant evidence for an attack. Finally, the normal \fl procedure continues with the remaining local models getting aggregated to the new global model.

\subsection{Metrics Intuition}
\label{sect:approach:defense:intuition}

\begin{figure}[tb]
\centering

\begin{subfigure}{0.1\linewidth}
    \includegraphics[width=\textwidth]{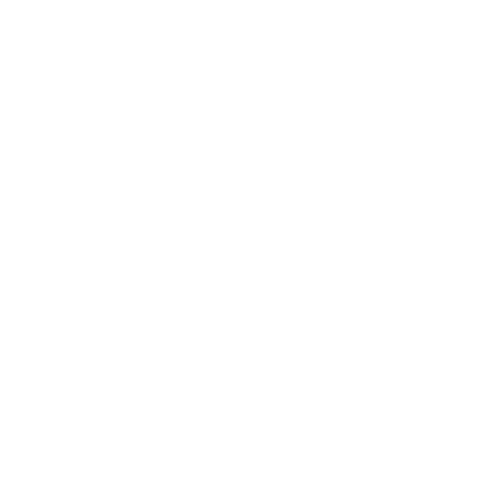}
\end{subfigure}
\begin{subfigure}{0.34\linewidth}
    \includegraphics[width=\textwidth]{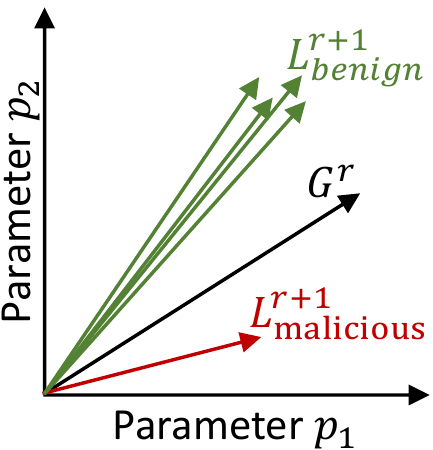}
\end{subfigure}
\hfill
\begin{subfigure}{0.34\linewidth}
    \includegraphics[width=\textwidth]{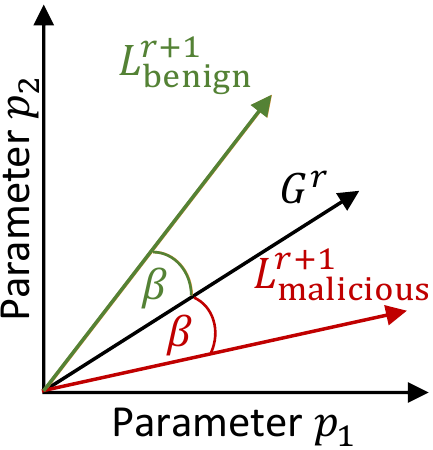}
\end{subfigure}
\begin{subfigure}{0.1\linewidth}
    \includegraphics[width=\textwidth]{images/white.pdf}
\end{subfigure}
\caption{Simplified visualization of \fl models with two parameters. The left graphic shows that benign and malicious models differ in one or multiple dimensions. On the right, we depict that benign and malicious models can have the same \metricCos metric due to the same angel to the global model.}
\label{fig:motivation:cos}
\end{figure}

 \dnns are complex \mbox{multi-dimensional} \mbox{non-linear} functions. An example of \dnn with around eleven million trainable parameters is \resneteighteen~\cite{resnet}. For a better explanation of our metrics, however, we will use a simplified function, which is linear and only has two parameters (or dimensions): $f(x) = p_1 \cdot x + p_2$. With this, we can visualize model parameters $p_1$ and $p_2$ in a 2D plot (cf. \hyperref[fig:motivation:cos]{\fig\ref{fig:motivation:cos}}), which won't be possible for a more realistic \mbox{multi-dimensional} function.

As visualized in the left graphic of \hyperref[fig:motivation:cos]{\fig\ref{fig:motivation:cos}}, an adversary conducting a poisoning attack in \fl needs to significantly change at least some model parameters of one or many poisoned local models in order to affect the behavior of the new global model. Otherwise, the respective parameter, and, thus, the new global model will align with the benign behaviour of the majority of clients (\cf~\hyperref[sec:approach:threatmodel]{\sect\ref{sec:approach:threatmodel}}) after aggregation. Hence, benign trained local models that learn similar behavior will be similarly distributed around the new global model after aggregation, since \fedavg decides for the average of all contributions. A malicious model, depicted in red color in \hyperref[fig:motivation:cos]{\fig\ref{fig:motivation:cos}}, must be located in a significantly different location than the benign models depicted in green to influence the averaging of \fedavg.

\ourname is based on a set of \numMetrics \mbox{well-chosen} metrics, that are extracted from local models. Technically, extraction of the metrics is a straightforward task that only needs to be conducted once for each local model within each \fl round \flround. The metrics can identify malicious models or updates based on different characteristics, like \textit{magnitude}, \textit{direction}, \textit{orientation}, \textit{functionality level}, and \textit{outliers}, which we will explain in detail in following.

\begin{figure}[tb]
  \centering
  \includegraphics[width=0.34\linewidth]{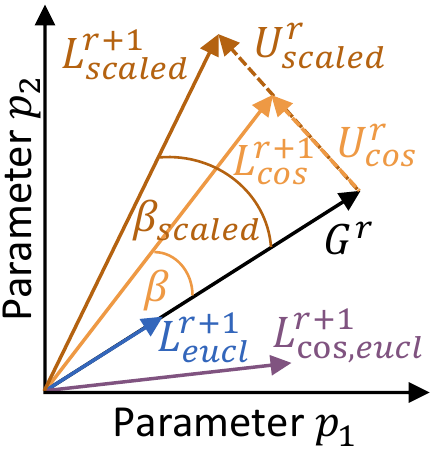}
  \caption{Visualization of locally trained models \localModelRoundNext deviating from the global model \globalModelRound in \metricCos and \metricEucl. The figure also depicts how the angle $\beta$ changes after scaling the update, thus provoking a change in the \metricCos metric of \ourname.}
\label{fig:motivation:update}
\end{figure}

\vspace{0.1cm} \noindent 
\textbf{Magnitude and Direction.} The two metrics to detect deviations in magnitude and direction of benign and malicious models, which have also been used by other works~\cite{nguyen22Flame, rieger2022deepsight, blanchard17Krum, yin2018trimmedMeanMedian, fung2020FoolsGold, munoz2019byzantineAFA}, are Euclidean distance (\metricEucl) and Cosine distance (\metricCos) measured between the locally trained models \localModelRoundNext and the original global model of the round \globalModelRound. These metrics are depicted in \hyperref[fig:motivation:update]{\fig\ref{fig:motivation:update}}.

\vspace{0.1cm} \noindent 
\textbf{Orientation.} Two models with the same \metricCos might significantly differ from each other, as depicted in the right graphic of \hyperref[fig:motivation:cos]{\fig\ref{fig:motivation:cos}}, as \metricCos alone is insufficient to reflect the direction. Therefore, the orientation of the Cosine from \globalModelRound can further differentiate two models. To incorporate this difference into a value, we propose \metricCount, a novel metric that counts how many parameter values are increased from the respective parameter of the global model \globalModelRound during training. This metric provides a measurement to detect substantially different models, that exhibit inconspicuous similarities in \metricCos. Moreover, the \metricCount formula (~\cf~\hyperref[app:defense]{\sect\ref{app:defense}}) incorporates the \textit{sign} function, which prevents straightforward adaption by adversaries. Specifically, attempts by adversaries to introduce an extra objective mirroring the \metricCount formula into the loss function are rendered ineffective. The reason being that learning algorithms cannot effectively propagate changes to the underlying model parameters through a sign function, given its constant zero gradient. As a result, the utilization of this metric enhances the robustness of the system against adaptive adversaries.

\begin{figure}[tb]
\centering
\begin{subfigure}{0.3\linewidth}
    \includegraphics[width=\textwidth]{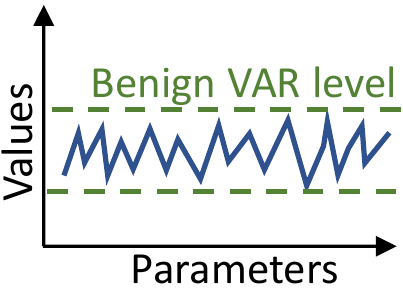}
\end{subfigure}
\hfill
\begin{subfigure}{0.3\linewidth}
    \includegraphics[width=\textwidth]{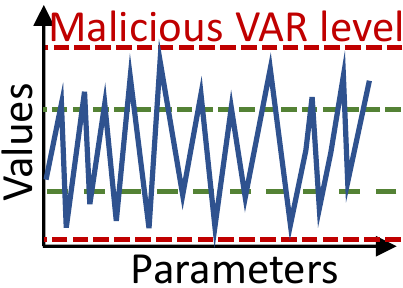}
\end{subfigure}
\hfill
\begin{subfigure}{0.3\linewidth}
    \includegraphics[width=\textwidth]{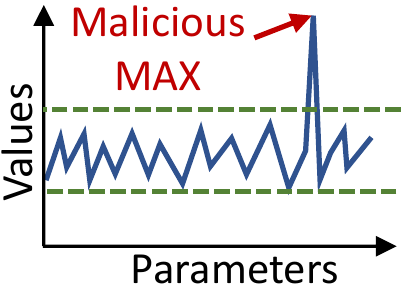}
\end{subfigure}
\hfill
        
\caption{Simplified visualization of \fl models with multiple parameters highlighting the functionality level based on the parameter value variance. The left shows a benign situation and the middle a poisoned model can have a bigger (or smaller) level. The figure on the right depicts,that the variance is not affected by maxima (and minima).}
\label{fig:motivation:var}
\end{figure}

\vspace{0.1cm} \noindent 
\textbf{Functionality Level.} Due to the many parameters of a \dnn, there can exist models with poisoned behavior, that have metrics \metricCos, \metricEucl, and \metricCount similar to benign models. Such a situation can occur, e.g., if the parameters of a model posses significantly different variance, as visualized in \hyperref[fig:motivation:var]{\fig\ref{fig:motivation:var}}\footnote{As highlighted in \hyperref[fig:motivation:var]{\fig\ref{fig:motivation:var}}, the \metricVar can be increased, but of course also a significant decrease is possible.}. We leverage this variance as metric (\metricVar) in \ourname and interpret it as functionality level, since a different \metricVar is a clear indication of divergent model behaviour.

\vspace{0.1cm} \noindent 
\textbf{Outliers.} As with any other variances, \metricVar is not affected by a few extreme outliers. Therefore, to catch those, we additionally investigate two more novel metrics: \metricMax and \metricMin, which extract the maximum/minimum parameter distance between all the parameters of local models \localModelRoundNext and a global model \globalModelRound.\footnote{We take the minimum distance bigger than zero for \metricMin by leveraging a nonzero function (\nonzeroFunc) Thus, \metricMin analyzes real model changes and ignores parameters that have not been changed.}. \metricVar combined with \metricMax and \metricMin provide a reliable metric for the functionality level and allow testing for poisoned models. Similarly to the \metricCount metric, the outlier metrics significantly enhance the system's resilience against adaptive adversaries. Specifically, the formulas for \metricMin and \metricMax (~\cf~\hyperref[app:defense]{\sect\ref{app:defense}}) can be incorporated as supplementary objectives in the loss function. However, it is noteworthy that the resulting changes are confined to the parameter responsible for reflecting the particular metric value. Consequently, other components within the model may undergo escalation in the metric while the actual outlier gets adjusted. This strategic attribute compels adversaries to employ additional measures, such as applying clipping mechanisms after model training is finished, to adjust remaining outliers in \metricMin and \metricMax to mount stealthy attacks.

\vspace{0.1cm} \noindent 
\textbf{Interrelations between metrics.} The selection of the aforementioned metrics was based on their inherent interrelations. For an adaptive adversary attempting to adjust to the \metricEucl metric, success can be achieved through scaling or introducing additional objectives in the loss function. Both these approaches are likely to influence the \metricCos metric. However, if the adversary adapts to the \metricCos metric, they might exploit a stealthy situation as depicted in the right graph of~\hyperref[fig:motivation:cos]{\fig\ref{fig:motivation:cos}}. Nevertheless, such a scenario would have an instant impact on the \metricCount metric. Furthermore, malicious behavior could be introduced by manipulating the variance of the model parameters, while remaining inconspicuous in terms of \metricEucl, \metricCos, and \metricCount metrics. However, the \metricVar metric would be capable of detecting such a situation. Further, a seemingly benign \metricVar constructed by employing extreme outliers, as visualized in the right graph of \hyperref[fig:motivation:var]{\fig\ref{fig:motivation:var}}, would immediately generate abnormal values in \metricMin or \metricMax metrics. Due to the specific properties of certain metrics, namely \metricCount, \metricMax, and \metricMin, which are non-trivial to adapt with additional objective functions\footnote{We discuss this in the respective metric sections above.}, \ourname effectively counteracts adaptive attacks.

\subsection{Pruning Loop}
\label{sect:approach:defense:loop}

The filtering process consists of three steps: \textit{statistical tests}, \textit{clustering}, and \textit{pruning} (2-4 in \hyperref[fig:overview:detail]{\fig\ref{fig:overview:detail}}). In every filtering round, each metric traverses the procedure independently. After each round, the models filtered based on any metric are excluded from the next round. This iterative pruning loop continues until the statistical tests do not report any significance for the presence of a poisoning attack anymore. Due to the iterative nature of this filtering procedure and the individual analysis of each metric, different types of poisoning attacks can be filtered within one run of \ourname.

\begin{figure}[tb]
  \centering
  \includegraphics[width=0.7\linewidth]{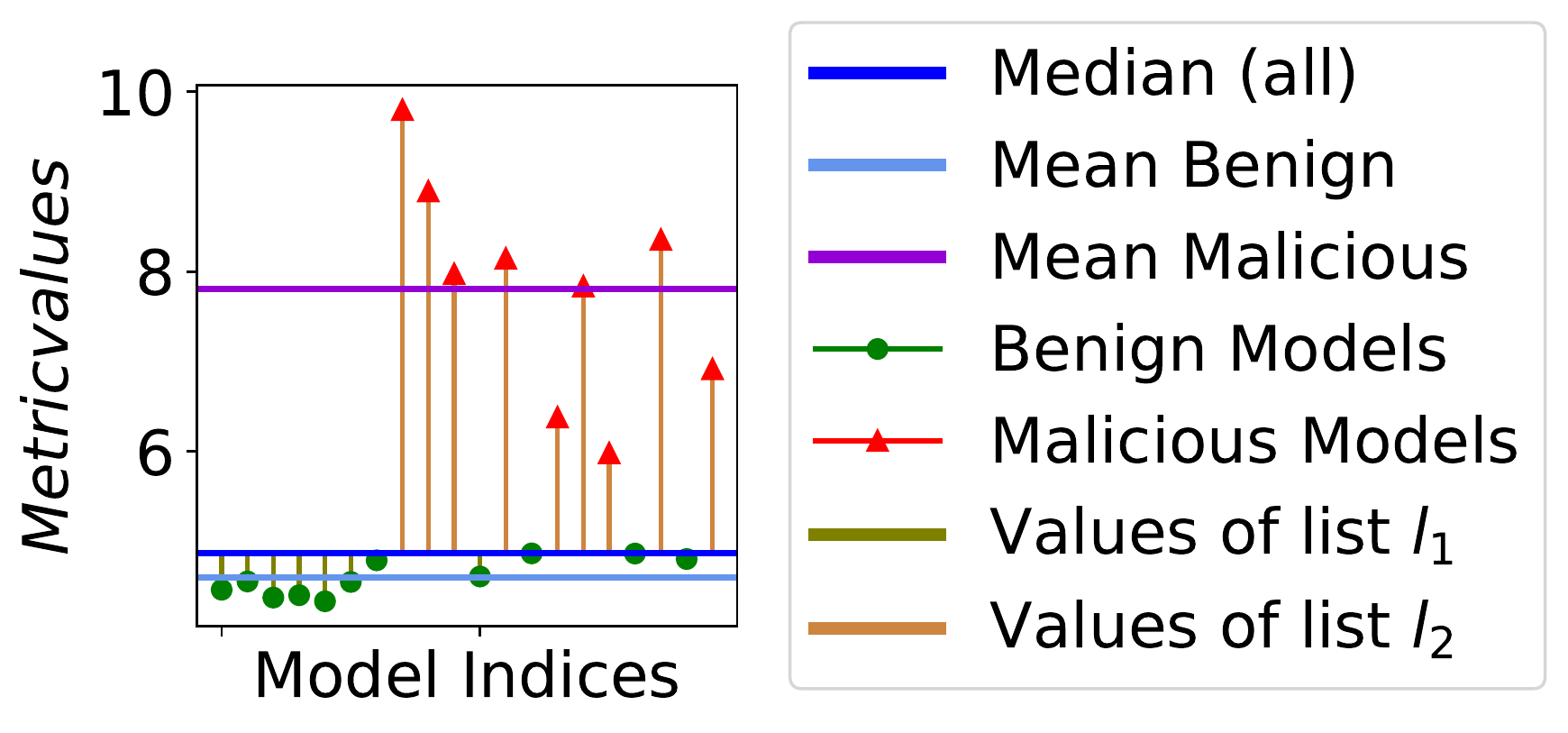}
  \caption{Depiction of a statistical test setup with significant p-value in \statTestT indicating a varying mean between $l_1$ and $l_2$.}
\label{fig:metric:t}
\end{figure}

\vspace{0.1cm} \noindent 
\textbf{Statistical Tests.} When provided with a set of metric values, which always contain one value per local model, the statistical tests first extract the median value, which is considered as benign due to the majority assumption (cf.~\hyperref[sec:approach:threatmodel]{\sect\ref{sec:approach:threatmodel}}). Afterwards, multiple statistical tests are conducted to check if all metric values are distributed equally around the median value, as one would expect from benign models. Therefore, \ourname checks if the metric values with bigger values than the median and the metric values with smaller values as the median follow the same distribution. For that purpose, the bigger and smaller metric values are converted to two lists $l_1$ and $l_2$ containing the absolute distance from the value to the median, as shown in \hyperref[fig:metric:t]{\fig\ref{fig:metric:t}}. Then the two lists pass through the tests. At first, a T-Test~\cite{ttest}~(\statTestT) is conducted to check for equal means. Since two distributions can have the same mean but different variances, a Levene’s test~\cite{ftest}~(\statTestV) is appended. Finally, a \mbox{Kolmogorow-Smirnow-Test}~\cite{dtest}~(\statTestD) for equal distributions is leveraged. Following the same reasoning we provided for the metrics \metricVar and \metricMax, the aforementioned tests are not significantly influenced by outliers. Therefore, we additionally analyze the original metric values regarding the 3$\sigma$ rule~\cite{pukelsheim1994three} (\statTestS). Values outside the 3$\sigma$ interval are marked as significant outliers. 

In \hyperref[fig:metric:t]{\fig\ref{fig:metric:t}}, the metric values of benign and malicious models are listed. The mean of all metric values (dark blue) is used to separate the values into two lists $l_1$ and $l_2$. Those lists represent the benign and malicious models, respectively, and are graphically observable by the lines between the metric values and the median. Note, that the median of the benign values (light blue) and the median of the malicious values (purple) have a significantly different distance to the median, which results in a highly significant result in \statTestT. \statTestT, \statTestV, and \statTestD deliver a p-value\footnote{A p-value indicates how likely it is that the underlying data could have occurred under a null hypothesis. In our case, the null hypothesis is, that the two lists contain samples from equal distributions, thus having equal mean and variance.}, which is also called significance level and is used to determine if a poisoned model is found.

\vspace{0.1cm} \noindent 
\textbf{Clustering and Pruning} After a significant statistical test (step 2 in \hyperref[fig:overview:detail]{\fig\ref{fig:overview:detail}}), \ourname leverages Agglomerative Clustering~\cite{frank16agglomerative} with two fixed clusters based on the Euclidean distance to cluster the significant metric values (step 3 in \hyperref[fig:overview:detail]{\fig\ref{fig:overview:detail}}). Afterwards, the local models behind the metric values within the bigger cluster are considered as benign based on the majority assumption and the other models are marked as malicious and excluded by the pruning step of \ourname (step 4 in \hyperref[fig:overview:detail]{\fig\ref{fig:overview:detail}}). 

\vspace{0.1cm} \noindent 
Overall, \ourname is robust against sophisticated poisoning attacks through an \mbox{in-depth} analysis of model weights using six interdependent metrics. As a result, if a strong adaptive adversary attempts to circumvent one metric, the artifacts of the poisoning attack will inevitably manifest through one of the other metrics. Further, \ourname adapts to the application domain including complicated \nonIid data scenarios by leveraging statistical tests, instead of relying on hard thresholds. We provide the formulas of the metrics and additional information about \ourname in \hyperref[app:defense]{\sect\ref{app:defense}}.

\section{Evaluation}
\label{sec:eval}
\begin{table*}
\fontsize{7pt}{8pt}\selectfont
  \caption{\mas and \bas in different scenarios in percent.}
  \label{tab:test}
  \begin{tabular}{l c|c|c|c|c|c|c|c|c|c|c|c|c}
    \multicolumn{2}{c|}{}  & \multicolumn{12}{c}{Scenario}\\
    \cline{3-14}
    \multicolumn{2}{c|}{}  & \multicolumn{2}{c|}{Default}   & \multicolumn{2}{c|}{\circled{1}}   & \multicolumn{2}{c|}{\circled{2}}   & \multicolumn{2}{c|}{\circled{3}} & \multicolumn{2}{c|}{\circled{4}} & \multicolumn{2}{c}{\circled{5}}  \\
    \toprule
    \multicolumn{2}{c|}{Accuracies without defenses} & \ma & \ba  & \ma & \ba  & \ma & \ba  & \ma & \ba& \ma & \ba & \ma & \ba \\
    \midrule
    1:& Global model \globalModelRound & 62.99 & 1.90 & 62.99 & 1.90 &62.99 & 1.90& 62.99 & 1.90 & 62.99 & 1.93   &  36.51 & 5.18\\
    \hline
    2:& Average of benign local models & 57.58 & 4.56 & 57.58 & 4.56& 57.58 & 4.56 & 57.58 & 4.56& 47.15 & 6.82   & 33.15 & 10.42\\
    3:& Average of poisoned local models & 57.84 & 85.13 & 54.58 & 93.15& 54.42 & 93.25& 51.23 & 89.82  & 43.74 & 91.32  & 33.93 & 82.00 \\
    \hline
    4:& \fedavg with benign local models & 63.57 & 1.85 & 63.57 & 1.85& 63.57 & 1.85& 63.57 & 1.85 & 65.92 & 1.40   & 32.45 & 12.71\\
    5:& \fedavg with poisoned local models & 64.92 & 83.00 & 63.68 & 92.50& 62.29 & 93.71 & 40.69 & 93.54& 59.12 & 95.50   & 29.35 & 88.96\\
    \hline
    6:& \fedavg with all local models & 63.81& 42.94 & 63.85& \textbf{61.96}& 63.27& 63.54& 49.18& 83.74 & 64.02& 63.66   & 38.72& 77.37\\
    \bottomrule
    \toprule
    \multicolumn{2}{c|}{Global model accuracies after applying defenses} &\ma&\ba& \ma & \ba  & \ma & \ba  & \ma & \ba  & \ma & \ba  & \ma & \ba \\
    \midrule
    7:& \naiveBig \clustering & 65.06 & 74.62 & 64.75 & 86.86& 63.73 & 88.36& 47.34 & 85.58  & 61.12 & 87.58   & 20.85 & \textbf{85.32} \\
    8:& \foolsgold~\cite{fung2020FoolsGold} & 63.57 & 1.85 & 63.57 & 1.85& 63.27 & \textbf{63.54}& 63.57 & 1.85  & 56.80 & 47.0  & 37.00 & \textbf{76.03}  \\
    9:& \krum~\cite{blanchard17Krum} & 59.75 & 83.53 & 52.22 & 95.97& 56.18 & 93.14& 52.00 & \textbf{89.90} & 49.88 & 5.27   & 16.88 & \textbf{89.07} \\
    10:& \multikrum~\cite{blanchard17Krum} & 64.18 & 83.05 & 63.90 & 92.72& 62.01 & 93.83& 41.86 & \textbf{95.80} & 62.39 & 13.11   & 18.07 & \textbf{89.55}\\
    11:& Clip~\cite{mcmahan2018iclrClippingLanguage} & 63.80 & 42.81 & 63.85 & 61.86& 63.26 & 63.52& 49.19 & 83.74 & 63.92 & 62.28   & 37.76 & \textbf{75.60} \\
    12:& Clip\&Noise~\cite{mcmahan2018iclrClippingLanguage} & 50.78&60.66 & 52.10&77.21& 59.32&75.67 & 41.47&90.37 & 56.28 & 71.99  & 23.70 & \textbf{64.32 } \\
    13:& \flame~\cite{nguyen22Flame} & 60.96 & 79.17 & 63.67 & 88.44& 62.21 & 88.80 & 44.56 & 84.53 & 56.59 & 50.34  & 25.10 & \textbf{79.17} \\
    14:& T-Mean~\cite{yin2018trimmedMeanMedian} & 63.51 & 44.13 & 63.54 & 63.98& 62.86 & 65.35& 51.07 & 85.75 & 63.15 & 67.01   & 39.98 & \textbf{76.36}\\
    15:& T-Median~\cite{yin2018trimmedMeanMedian} & 51.22&44.60 & 51.18&57.73& 49.61&60.30& 39.76&74.76 & 51.75 & 68.20   & 17.04 & \textbf{52.75} \\
    16:& FLTrust~\cite{cao2020fltrust} & 63.49 & 23.08 & 63.76 & 49.68& 63.17 & 45.54 & 55.15 & \textbf{74.71}& 63.56 & 8.40   & 26.81 & \textbf{81.61}\\
    17:& \textbf{\ourname} & 63.57 & \textbf{1.85} & 63.36 & \textbf{1.95}& 63.36 & \textbf{1.95}& 63.57 & \textbf{1.85} & 65.92 & \textbf{1.40}   & 37.52 & \textbf{2.37}\\
    \bottomrule
    \toprule
    \multicolumn{5}{l}{\circled{1} Default + PDR 0.3}& \multicolumn{9}{l}{\circled{2} Default + PDR 0.3 + Last Layer Fixation to benign models}\\ 
    \multicolumn{5}{l}{\circled{3} Default + Adapt to EUCL of benign models}& \multicolumn{9}{l}{\circled{4} Default + PDR 0.3 + \oneClass \intraNonIid with $\noniidsign = 0.5$ + Scaling}\\
    \multicolumn{14}{l}{\circled{5} Default + \interNonIid based on our \randomgen strategy}\\
\end{tabular}
\end{table*}

In this section, we conduct a rigorous analysis of \ourname and explore impact of various parameters and application-specific factors like datasets, model architectures, underlying data distributions, poisoning methods and attack adaptive strategies, as well as performance overheads. 

\subsection{Experimental Setup and Scenarios} 
\label{sect:eval:setup}
\textbf{Hardware and Software.} We execute the \fl system consisting of a configurable amount of clients on one server and implement the code in PyTorch~\cite{pytorch, paszke2019pytorch}, a \mbox{well-known} machine learning library for Python~\cite{van1995python}.\footnote{In our setting, we create 20 clients, of which nine are captured by an adversary. Nine malicious clients are the maximum the attacker can control in our setup while remaining within our attacker model (cf.~\hyperref[sec:approach:threatmodel]{\sect\ref{sec:approach:threatmodel}}).} The individual client and server code is executed sequentially on the server running with an AMD EPYC 7413 \mbox{24-Core} Processor (\mbox{64-bit}) with 96 processing units and 128GB main memory. An NVIDIA A16 GPU with 4 virtual GPUs each having 16GB GDDR6 memory is accessible via CUDA~\cite{cuda} from PyTorch.

\vspace{0.1cm}\noindent
\textbf{Datasets and Models.} We chose similar settings to \fl defenses in related works and focus mainly on image classification with \cifar~\cite{cifar}, \gtsrb~\cite{gtsrb}, and \mnist~\cite{mnist}. We use \resneteighteen~\cite{resnet}, \squeezenet~\cite{iandola2016squeezenet}, and a CNN model architectures. Additionally, we investigate into the text domain by training a DistilBERT~\cite{sanh2020distilbert} transformer model on \ssttwo~\cite{socher2013recursive_sst} sentiment analysis dataset.

\vspace{0.1cm}\noindent
\textbf{Default Scenario.} We train the \cifar~\cite{cifar} image classification task (ten classes) on a \resneteighteen~\cite{resnet} model with \lr 0.01 (SGD optimizer, momentum 0.9, decay 0.005), a batch size of 64, and ten local training epochs. The federation is a realistic setup, which consists of $\clientCount = 20$ clients, which are all selected each round \flround ($\clientCountSelected = 20$). The data are \iid distributed and each client has 2560 samples, 256 randomly chosen from each class. The adversary captures nine clients leading to a poison model \mbox{rate (\pmr)} of 0.45, which is the maximum rate for this amount of clients. He sets the poison data rate (\pdr) to 0.1, \alphasign to 0.3, utilizes the adaption strategies from \hyperref[sec:approach:threatmodel]{\sect\ref{sec:approach:threatmodel}} and implements a pixel trigger backdoor~\cite{badnets} (\cf~\hyperref[app:triggermethods:pixel]{\sect\ref{app:triggermethods:pixel}}), which adds pixel pattern, a sticker, or similar as a trigger to the sample~\cite{badnets, bagdasaryan2020backdoorfl, trojanTriggerTargeted}. The global model \globalModelRound is already trained 50 benign rounds and was originally initialized with \pretrained weights from PyTorch, with the first and last layers being untrained since both needed to be changed according to our dataset.\footnote{The \pretrained models from PyTorch are trained on ImageNet~\cite{deng2009imagenet}, thus have other input dimensions and 1000 instead of ten classes.} 

\vspace{0.1cm}\noindent
\textbf{Defenses.} We compare the following \numdefenses approaches, with \ourname regarding effectiveness and runtime, hence examine \df, \ra, and \ir methods: \naiveBig \textit{clustering via \hdbscan\cite{McInnes2017}}, \textit{\foolsgold~\cite{fung2020FoolsGold}}, \textit{\krum~\cite{blanchard17Krum}}, \textit{\multikrum~\cite{blanchard17Krum}}, \textit{\flame~\cite{nguyen22Flame}}, \textit{\tmean}~\cite{yin2018trimmedMeanMedian}, \textit{\tmedian}~\cite{yin2018trimmedMeanMedian}, \mbox{\textit{Clipping\&Noising}~\cite{mcmahan2018iclrClippingLanguage}}, \textit{Clipping}~\cite{mcmahan2018iclrClippingLanguage}, and \textit{Auror}~\cite{shen16Auror}. We either adapted \mbox{open-source} implementations or reimplemented the methods if no code was available.

\vspace{0.1cm}\noindent
First, we consider our default scenario, and later we will expand the analysis to adaptive adversaries, \numattacks poisoning attacks, and \nonIid data scenarios. Due to space limitations, we report the most interesting results and numbers that highlight our outcomes in the following sections and list detailed experimental results in~\hyperref[app:additionalresults]{\sect\ref{app:additionalresults}}.

\subsection{Defenses under Strong Adaptive Adversaries}
\label{sect:eval:saa}
Before discussing defenses, we note that the \ba of our default scenario without defense is only 42.94\% (line 6 in \hyperref[tab:test]{\tab\ref{tab:test}}), hence the backdoor is not effective ($< 60\%$) and the adversary is forced to adapt his attack by either increasing the \pdr, increasing the \pmr\footnote{Our default scenario already includes the maximum valid \pmr defined in \hyperref[sec:approach:threatmodel]{\sect\ref{sec:approach:threatmodel}}.}, or by fixation, constraining and scaling. The increased \ba of 61.96\% for an increased \pdr to 0.3 can be seen in scenario \circled{1} in \hyperref[tab:test]{\tab\ref{tab:test}}. We explore the effectiveness of these strategies and list results in~\hyperref[app:additionalresults]{\sect\ref{app:additionalresults}}. Here, we show that \ourname is more effective than other defenses even without applying additional adaptions when comparing them under the default scenario as well as for increased \pdr in scenario \circled{1}: As can be seen in line 17 in \hyperref[tab:test]{\tab\ref{tab:test}}, \ourname effectively removes the backdoor by reducing \ba to 1.85\% and 1.95\%, while most other defenses are less potent. Only \foolsgold~\cite{fung2020FoolsGold} is as effective as \ourname in the default scenario and in scenario \circled{1}, but, as we will elaborate later in this section, \foolsgold could be easily circumvented through adaption.

Since the adversary has to use one of the adaption strategies to reach a higher \ba, we want to clarify beforehand that an increased \pdr reinforces already existing significant values in \ournameGen metrics even more. Scaling of updates has positive effects on \ourname, since concurrently the metric \metricCos will be changed, as visualized in \hyperref[fig:motivation:update]{\fig\ref{fig:motivation:update}}\footnote{When scaling, our strong adaptive adversary is aware of benign values from training benign model first and scales to the mean of those values. Additionally, Gaussian noise is added to the targeted value within the 3rd percentile of the benign value range to make the malicious models slightly different and, hence, increase stealthiness (otherwise the models with exactly the same values could be easily detected ).}. Further, constraining with \hyperref[eq:constraint]{\equ\ref{eq:constraint}} or \hyperref[eq:saa]{\equ\ref{eq:saa}} also benefits \ourname due to side effects on its other metrics, forcing the adversary into a \mbox{multi-objective} optimization (\moo) problem and, thus, hardening the adversarial dilemma. Lastly, fixation methods are ineffective against \ourname, since all layers and the model as a whole are analyzed independently with statistical tests. Hence, \ourname is robust against adaptions of a strong adaptive adversary, which, we show, an attacker can leverage to circumvent other defenses.

\subsubsection{Circumvent Defenses}
\label{sec:eval:saa:defenses}

Below, we will focus on the capability of defenses to reduce the \ba in the new global model after aggregation compared to aggregation without defense (\cf line 6 in~\hyperref[tab:test]{\tab\ref{tab:test}}). Additionally, we will report the detection accuracy (\acc) of the defenses, when applicable, where 100\% \acc means perfect detection rate and no \mbox{False-Positives} (\fps) and \mbox{False-Negatives} (\fns). We will also name the most effective adaption strategies based on results provided in~\hyperref[app:additionalresults]{\sect\ref{app:additionalresults}}, which we couldn't include in the main section of the paper due to space limitations. 

\vspace{0.1cm}\noindent
\textbf{Clustering.} To demonstrate that \naive clustering methods could be bypassed, we use the \hdbscan\cite{McInnes2017} algorithm as an example and cluster based on the \mbox{cross-wise} Cosine distances between model updates. As can be seen in the default scenario in line 7 of \hyperref[tab:test]{\tab\ref{tab:test}}, the defense is ineffective reaching a \ba of 74.62\% in the new global model after aggregation. We additionally report an \acc of only 10\% (\fpr of 100\% and 81\% \fnr). Thus, there is no need for an attacker to follow any adaption strategies. Nevertheless, adaption to \naive clustering is possible by increasing the \pdr allowing us to embed a \ba of 86.86\%, as depicted in scenario \circled{1}.

\vspace{0.1cm}\noindent
\textbf{\foolsgold.} The second defense, \foolsgold~\cite{fung2020FoolsGold}, is also based on \mbox{cross-wise} Cosine distances between model updates. However, it analyzes only outputs of the last layer, which is more effective than \naive clustering and is capable of removing all poisoned models in the default scenario and for scenario, \circled{1} reaching \bas of 1.85\%, as depicted in line 8 of \cf \hyperref[tab:test]{\tab\ref{tab:test}}. Nevertheless, the defense can be circumvented using adaption. The best results we obtained by parameter \textit{fixation} on the last layer in combination with \pdr increase, depicted as scenario \circled{2} in \hyperref[tab:test]{\tab\ref{tab:test}}, reaching a \ba of 63.54\%. In contrast, \ourname still removes the backdoor to 1.95\% with only one \fp when a similar adaption strategy is applied.

\vspace{0.1cm}\noindent
\textbf{\krum.} Next, we evaluated \krum and \multikrum~\cite{blanchard17Krum}, which leverage \mbox{cross-wise} Euclidean distances between local models. The trigger backdoor is not reflected in this metric, which renders the defense ineffective for our default scenario (83.53\% and 83.05\%\ba for \krum and \multikrum, resp. in \hyperref[tab:test]{\tab\ref{tab:test}}) and for scenario \circled{1} and \circled{2}. Since \krum selects one single local model as the new global model, it can either choose a malicious or benign local model. In the former case, the backdoor trivially makes it to the global model. In the latter case, we can follow the following strategy: We can adapt the malicious models via constraint method \hyperref[eq:constraint]{\equ\ref{eq:constraint}} forcing the \krum scores of poisoned models to be more equal to each other compelling \krum to decide in their favor. By circumventing \krum like this, we achieved \bas up to 89.90\% and reached 95.80\% \ba for \multikrum as can be seen in scenario \circled{3} in \hyperref[tab:test]{\tab\ref{tab:test}}. In contrast, \ourname accurately filters the backdoor in similar circumstances, as adaption via constraint has significant effects on other metrics, like \metricEucl and \metricMin.

\begin{table*}
\fontsize{7pt}{8pt}\selectfont
  \caption{\ba for targeted and \acc for untargeted poisoning attacks without adaptive adversary in percent.}
  \label{tab:acc:saa:trigger:ba}
  \begin{tabular}{l c|c|c|c|c|c|c|c|c|c}
    \toprule
    \multicolumn{2}{c|}{\multirow{3}{*}{Aggregation / Defenses}} &\multicolumn{6}{c|}{\ba} & \multicolumn{3}{c}{\acc}\\
    & & Pixel Trigger & Clean-Label & Semantic & Edge Case & Label Flip &  Pervasive  & Random Flip & Sign Flip & Noising\\
    & & \cite{badnets} & \cite{turner2019label} & \cite{bagdasaryan2020backdoorfl} & \cite{wang2020attackontails} & \cite{biggio2013poisoninglabelflip, cao2019labelflipdistributed} & \cite{chen2017targetedblend}  & \hyperref[app:triggermethods:random]{\sect\ref{app:triggermethods:random}} & \hyperref[app:triggermethods:sign]{\sect\ref{app:triggermethods:sign}} & \hyperref[app:triggermethods:noise]{\sect\ref{app:triggermethods:noise}}\\
    \midrule
    1:& Global model \globalModelRound & 1.90 & 1.90 & 0.00 & 1.53 & 0.10 & 0.02 & - & - & -\\
    \hline
    2:& Average of benign local models & 4.56 & 4.57 & 0.00 & 2.55 & 1.24 & 0.95 & - & - & -\\
    3:& Average of poisoned local models & 85.13 & 75.49 & 80.0 & 19.28 & 74.15 & 97.28 &- & - & -\\
    \hline
    4:& \fedavg with benign local models & 1.85 & 1.85 & 0.00 & 1.85 & 0.20 & 0.07 & - & - & -\\
    5:& \fedavg with poisoned local models & 83.00 & 81.75 & 100.0 & 20.40 & 71.20 & 99.84 & - & - & -\\
    \hline
    6:& \fedavg with all local models & 42.94 & 38.92 & 60.0 & 6.63 & 49.20 & 3.58 & - & - & -\\
    \hline
    7:& \naiveBig \clustering & 74.62 & \textbf{1.85} & 60.0 & 16.35 & 65.60 & 67.67 & 10.00 & \textbf{100.00} & 80.00\\
    8:& \foolsgold~\cite{fung2020FoolsGold} & \textbf{1.85} & \textbf{1.85} & \textbf{0.00} & \textbf{2.55} & \textbf{0.20} & \textbf{0.10} & 55.00 & \textbf{100.00}& 0.00\\
    9:& \krum~\cite{blanchard17Krum} & 83.53 & 75.65 & 80.00 & 20.91 & \textbf{1.30} & \textbf{0.42} & 50.00 & 50.00 & 50.00\\
    10:& \multikrum~\cite{blanchard17Krum} & 83.05 & 82.38 & 100.0 & 18.87 & \textbf{0.40} & \textbf{3.50} & 75.00 & 75.00 & 75.00\\
    11:& Clip~\cite{mcmahan2018iclrClippingLanguage} & 42.81 & 38.91 & 60.0 & \textbf{6.63} & 48.40 & \textbf{3.17} & - & - & -\\
    12:& Clip\&Noise~\cite{mcmahan2018iclrClippingLanguage} & 60.66 & 40.73 & \textbf{0.00} & 12.75 & 30.80 & 10.08 & - & - & -\\
    13:& \flame~\cite{nguyen22Flame} & 79.17 & 77.12 & 60.0 & 18.87 & \textbf{2.40} & \textbf{5.52} & \textbf{100.00} & \textbf{100.00}& \textbf{100.00}\\
    14:& T-Mean~\cite{yin2018trimmedMeanMedian} & 44.13 & 41.10 & 60.0 & \textbf{7.14} & 48.40 & \textbf{2.53} & - & - & -\\
    15:& T-Median~\cite{yin2018trimmedMeanMedian} & 44.60 & 25.66 & \textbf{0.00} & \textbf{2.55} & \textbf{5.60} & \textbf{0.10} & - & - & -\\
    16:& FLTrust~\cite{cao2020fltrust} & 23.08 & 37.83 & \textbf{0.00} & \textbf{5.10} & \textbf{0.2} & \textbf{0.11} & 60.00 & 20.00 & 35.00\\
    17:& \textbf{\ourname} & \textbf{1.85} & \textbf{3.71} & \textbf{0.00} & \textbf{2.55} & \textbf{0.20} & \textbf{0.05} & \textbf{95.00} & \textbf{100.00}& \textbf{100.00}\\
  \bottomrule
\end{tabular}
\end{table*}

\vspace{0.1cm}\noindent
\textbf{\flame.} We evaluate \flame~\cite{nguyen22Flame}, a more complex \df defense, which combines clustering with clipping and noising techniques. Since the underlying metric is the same as for the \naive clustering defense, it is not very effective in removing the backdoor even in the default scenario achieving 79.17\% \ba (\cf line 13 in \hyperref[tab:test]{\tab\ref{tab:test}}). Similar to \naive clustering, we could strengthen the \ba by increasing the \pdr to 88.44\% and by additional scaling to 91.34\%, which shows that relying solely on the leveraged metric of \flame is insufficient. \ourname erases the backdoor efficiently in all of the cases, due to the \mbox{in-depth} model analysis with statistical tests and increased robustness against adaption through leveraging \numMetrics different metrics.

\vspace{0.1cm}\noindent
\textbf{FLTrust.} As a more recent defense, we analyze FLTrust~\cite{cao2020fltrust}, which is based on a trusted root dataset\footnote{Similar to the authors, we used a trusted root dataset of 100 \iid samples and excluded them from the datasets used for training of the clients.} on the server side. FLTrust leverages the Cosine Similarity between the updates of the local models and a trusted model trained by the server on the trusted root dataset and the norm of the local updates. Based on these metrics, FLTrust assigns weights to each local update so that poisoned updates are assigned with low weights, preferably zero, which would filter out the update. Therefore, the defense is ineffective, if the backdoor is not visible within both of these metrics, meaning, that the Cosine Similarity is inconspicuous, which can happen if the backdoor is only embedded in one layer without affecting the model-wise metric value, or if the backdoor is hidden in other metrics, as \metricVar, \metricMax, or \metricMin only. In most of our experiments, FLTrust successfully weakened backdoors beneath critical \bas. However, the assigned weights to all (also benign) local updates were found to be relatively small (mostly between 0.001 and 0.03), thereby inadvertently reducing their contribution to the global model. Consequently, the approach's efficacy comes at the cost of slowing down the training speed. Additionally, FLTrust's effectiveness depends on the chosen metric's ability to accurately reflect the backdoor. However, a backdoor is not necessarily embedded in those metrics, as can be seen in experiments, e.g., in scenario \circled{3} yielding a \ba of 74.71\%. For adaption, we first trained a benign model and then proceeded to adapt the local update of the malicious model based on the update of this benign model. Since we observed that in the resulting models, the main reason for suspicious metric values in Cosine Similarity originated from the parameters of the last model layer, we restricted this adaptation process to the last layer only to make the backdoor inconspicuous in the last layer. While this strategy resulted in low \ba, FLTrust assigns higher weights to the malicious updates than to the benign ones, with seven out of eleven benign updates being assigned a weight of zero. That means that those benign models were filtered, essentially slowing down the learning process, while malicious models were included in aggregation, even so with smaller weights. In contrast, \ourname consistently and effectively eliminated the backdoor in all of these cases without decreasing the impact of benign models.

\vspace{0.1cm}\noindent
\textbf{Differential Privacy.} Besides \df methods, we evaluated two \ir approaches: Model update clipping based on the Euclidean distance and a combination with model parameter noising~\cite{mcmahan2018iclrClippingLanguage}. Clipping is ineffective, as our default scenario backdoor is not reflected in the Euclidean distance of the updates. Thus, the attacker can achieve 60.66\% \ba for the default scenario (\cf line 12 of \hyperref[tab:test]{\tab\ref{tab:test}}). When using adaption, the \ba can be increased slightly to 61.86\% by increasing the \pdr as in scenario \circled{1}. In contrast, \ourname is effective under similar circumstances resulting in 1.85\% and 1.95\% \bas.
 
\vspace{0.1cm}\noindent
\textbf{Robust Aggregation.} We evaluate \tmean and \tmedian~\cite{yin2018trimmedMeanMedian}, which are \ra alternatives to \fedavg. Both result in weak backdoors with \ba of 44.13\% and 44.60\% in lines 14 and 15 for the default scenario, but are not robust when facing a strong adaptive adversary: \tmean can be bypassed with up to 63.98\% \ba, while \tmedian shows 57.37\% \ba, but also experiences around 10\% reduction in \ma in scenario \circled{1}. Hence, both approaches are not comparable to the performance of \ourname, which reduces \ba to 1.95\% under similar circumstances.

\vspace{0.1cm}\noindent
\textbf{\ourname.} To circumvent \ourname, we tried to adapt to respective metrics that reflect the different poisoning attacks. We succeeded in adapting to \metricCos, \metricEucl, \metricMin, and \metricMax, which appeared to be the metrics most backdoors manifest first. This was only possible by leveraging the loss scaling method of our strong adaptive adversary, as described in \hyperref[sec:approach:threatmodel]{\sect\ref{sec:approach:threatmodel}}, since otherwise, adaption to multiple losses already resulted in facing an adversarial dilemma. However, as soon as we adapt to those metrics, this behavior is reflected in the other metrics, namely \metricVar and \metricCount. For a few experiments, we succeeded in adapting to \metricVar, even if the \ma suffered immensely, but additional adaption to \metricCount was impossible.

\subsubsection{Different Poisoning Attacks}
In the following, we evaluate the effectiveness of the defenses against various poisoning attacks, including \numattackstargeted different trigger methods for targeted attacks and \numattacksuntargeted untargeted attacks, namely pixel triggers~\cite{badnets}, \mbox{clean-label} backdoor~\cite{turner2019label}, semantic backdoor~\cite{bagdasaryan2020backdoorfl}, edge case backdoor~\cite{wang2020attackontails}, label flip backdoor~\cite{biggio2013poisoninglabelflip, cao2019labelflipdistributed}, and pervasive backdoor~\cite{chen2017targetedblend} as well as random label flipping (\cf~\hyperref[app:triggermethods:random]{\sect\ref{app:triggermethods:random}}), sign flipping (\cf~\hyperref[app:triggermethods:sign]{\sect\ref{app:triggermethods:sign}}), and model noising (\cf~\hyperref[app:triggermethods:noise]{\sect\ref{app:triggermethods:noise}}) which are all explained in detail in \hyperref[app:triggermethods]{\sect\ref{app:triggermethods}}. We report the \bas that the poisoning attacks achieve against the \numdefenses defenses in \hyperref[tab:acc:saa:trigger:ba]{\tab\ref{tab:acc:saa:trigger:ba}} and the \mas in \hyperref[tab:acc:saa:trigger:ma]{\tabapp\ref{tab:acc:saa:trigger:ma}}.\footnote{The results reported in tables do not consider adaption (which is evaluated in \hyperref[sec:eval:saa:defenses]{\sect\ref{sec:eval:saa:defenses}}), as the system’s adaptability is directly tied to the specific defense employed. Instead, the tables focus on determining whether \ourname can successfully detect various triggers. Thereby, we ensure that our findings remain independent from the pixel-trigger method of the default scenario.}

\vspace{0.1cm}\noindent
\textbf{Pixel Trigger Backdoor} This backdoor is discussed in~\hyperref[sec:eval:saa:defenses]{\sect\ref{sec:eval:saa:defenses}}, where we showed that we can circumvent existing defenses by adaption and strengthening the trigger. Only \ourname could reliably remove the backdoor. 

\vspace{0.1cm}\noindent
\textbf{Clean-Label Attack.} This attack is not suited perfectly for \fl, since it is hard to embed a high \ba with low \pdr into the new global model. In our default scenario, we reached only 11.85\% \ba after aggregation, which is why we report the result for \pdr 0.5, which leads to a \ba 38.92\% without defense (line 6 in \hyperref[tab:acc:saa:trigger:ba]{\tab\ref{tab:acc:saa:trigger:ba}}). Nevertheless, it is possible to achieve a high \ba of up to 82.38\% for \multikrum (line 10), while \naive clustering, \foolsgold, and \ourname erase the backdoor. Among them, \ourname is the only one that cannot be adapted and erases the backdoor, which manifests in \metricCos and \metricEucl, resulting in a \fnr of 81\%.\footnote{We experienced an elevated \fns in a scenario with a maximum \pmr and one benign outlier model. We could not reproduce such scenarios on purpose when acting as an adversary. Such scenarios can only occur, if the \pmr is at a peak of nearly 50\% and one benign outlier exists, which then violates the majority assumption of \hyperref[sec:approach:threatmodel]{\sect\ref{sec:approach:threatmodel}}. However, if such situations occur, \ourname still ends up aggregating only benign models as long as the poisonings are significant in at least one metric in one layer. Hence is also robust against coincidental benign outliers.}

\vspace{0.1cm}\noindent
\textbf{Semantic Backdoor.} Without defense, this backdoor is effective with 60\% \ba. However, it is detectable within the last layers by \foolsgold~\cite{fung2020FoolsGold} leading to 0.00\% \ba (line 8 of \hyperref[tab:acc:saa:trigger:ba]{\tab\ref{tab:acc:saa:trigger:ba}}). Clip\&Noise and \tmedian also remove the backdoor, but at the same time reduce \ma. \ourname erases the backdoor completely by leveraging \metricMax metric. We report one \fp in this case for \ourname, but with a good result in a \ba of 0.0\%. Other effective defenses can be circumvented through adaption (\foolsgold) or reduce the \ma (\tmean and Clip\&Noise).

\vspace{0.1cm}\noindent
\textbf{Edge Case Backdoor.} It appears to be hard to embed an effective backdoor with this method even within the local models for \cifar~\cite{cifar} on \resneteighteen~\cite{resnet}. In \hyperref[tab:acc:saa:trigger:ba]{\tab\ref{tab:acc:saa:trigger:ba}}, we report the results for a \pdr or 0.3 with 19.78\% \ba on the local clients on average (line 3) and 6.63\% \ba without defense. \ourname is already sensitive to the poisoning attacks even when the effect on the global model is still minimal with 6.63\% \ba (line 6). We reach 100\% \tps and only two \fps resulting in the lowest \ba with 2.55\% in this case (line 17).

\vspace{0.1cm}\noindent
\textbf{Label Flip Backdoor.} This attack manifests in extreme deviations within the last layer of a \dnn. Hence, many defenses can easily detect the backdoor, as can be retraced on the low \bas in \hyperref[tab:acc:saa:trigger:ba]{\tab\ref{tab:acc:saa:trigger:ba}}. Having two \fps, \ourname is the only defense reducing the \ba to 0.20\% while being robust against fixation and adaption attempts, which can be used to circumvent other defenses like \foolsgold.

\vspace{0.1cm}\noindent
\textbf{Pervasive.} Blend~\cite{chen2017targetedblend} can be implemented with a \pdr of 0.1 to achieve 99.84\% \ba locally on average (line 3 in \hyperref[tab:acc:saa:trigger:ba]{\tab\ref{tab:acc:saa:trigger:ba}}), but it is inefficient in \fl -- we could only reach 3.58\% \ba for the global model without defense (line 6). \ourname can detect all poisoned local models while suffering five \fns. The result is interesting, as it shows that \ourname reaches the lowest \ba of 0.05\% while having minor effects on the \ma, whereas other defenses affect the \ma (\cf~\hyperref[tab:acc:saa:trigger:ma]{\tabapp\ref{tab:acc:saa:trigger:ma}}) or can be circumvented by adaption.

\vspace{0.1cm}\noindent
\textbf{Untargeted Attacks.} For the untargeted attacks, we do not report the \bas, but the \acc of the defense mechanisms in \hyperref[tab:acc:saa:trigger:ba]{\tab\ref{tab:acc:saa:trigger:ba}} and the resulting \mas in \hyperref[tab:acc:saa:trigger:ma]{\tabapp\ref{tab:acc:saa:trigger:ma}}. Random label flipping (\cf~\hyperref[app:triggermethods:random]{\sect\ref{app:triggermethods:random}}) is the first untargeted attack that we implemented. The \ma is reduced to 57.03\% without any defense and only \multikrum and FLTrust can score a higher \ma of 64.15\% and 63.16\%, respectively, compared to 62.88\% of \ourname. However, \multikrum suffers a \fnr of 45\%, compared to 0.09\% of \ourname and FLTrust comes with an \acc of 60\%\footnote{We compute the \acc of FLTrust~\cite{cao2020fltrust} by analyzing the weights assigned to the models. A model assigned with a weight of zero is considered as a filtered model.}, while \ourname achieves 95\%. \flame stands out with 100\% \acc, but can be circumvented by adaption. Second, we evaluated sign flipping (\cf~\hyperref[app:triggermethods:sign]{\sect\ref{app:triggermethods:sign}}), which is clearly detectable by defenses leveraging clustering methods including \ourname, but can lead to a \naive model with 10\% \ma for other approaches. Finally, we report the results for the model noising attack (\cf~\hyperref[app:triggermethods:noise]{\sect\ref{app:triggermethods:noise}}), where \ourname also has an \acc of 100\%, whereas other methods, like FLTrust achieve only 35\% \acc.

\vspace{0.1cm}\noindent
Concluding, we can say, that \ourname is robust against \numattacks poisoning attacks executed by a strong adaptive adversary, who is able to intentionally circumvent all other \numdefenses evaluated defenses.  We argue that any other defense, that relies on just a few metrics, could be similarly bypassed in our strong adaptive adversary model, by either fixation or constraint methods.

\subsection{Defenses under \nonIidBig}
\label{sec:eval:iid}
Here, we evaluate the same \numdefenses defenses, as specified in \hyperref[sect:eval:setup]{\sect\ref{sect:eval:setup}}, under different \nonIid scenarios. First, we investigate classical \intraNonIid before we discuss \interNonIid.

\vspace{0.1cm}\noindent
\textbf{\intraBigNonIid.} We analyzed various \intraNonIid settings, namely \oneClass, \twoClass, and Distribution \nonIid. For the \oneClass and \twoClass \nonIid scenarios, the samples of a client's dataset have a focus on one or two so-called \textit{main labels}. The remaining labels contain an equivalent amount of samples, while a factor \noniidsignFormular defines the fraction between the number of samples within the main label class and the remaining classes\footnote{For $\noniidsign = 1$, all samples are from the main label. $\noniidsign = 0$ is equal to the \iid scenario.}. Distribution \nonIid assigns label frequencies for each dataset based on a distribution, e.g., Dirichlet~\cite{minka2000dirichlet} or normal distribution. We elaborate on \nonIid simulation techniques in more detail in \hyperref[app:distributions:intra]{\sect\ref{app:distributions:intra}}. As representative results, we present \intraNonIid based on \oneClass with $\noniidsign = 0.5$.

We notice that in \nonIid settings it is harder for the adversary to embed a backdoor due to the nature of \fedavg. To reach a reasonable \ba of above 60\%, the adversary must use adaption strategies. We find that increase of \pdr to 0.3 combined with scaling reaches reasonable performance with 63.66\% \ba (scenario \circled{4} in \hyperref[tab:test]{\tab\ref{tab:test}}). \krum and \multikrum~\cite{blanchard17Krum} erase the backdoor, but simultaneously reduce the \ma immensely. However, after an adaption, we can circumvent those defenses reaching \bas of up to 90.44\%, while still erasing the backdoor with \ourname. \foolsgold is effective, but can be circumvented by adaption, while FLTrust also decreases the \ba to 8,40\%, but assigns weights for update aggregation between 0.0 and 0.025 to all model updates, effectively removing the influence of most of the update. \ourname is the only defense erasing the backdoor efficiently in this setting and reaching 1.40\% \ba with two \fps. Hence, we can confidently say, that \ourname outperforms other defenses in \intraNonIid settings.

\vspace{0.1cm}\noindent
\textbf{\interBigNonIid.} To simulate even more realistic datasets, we designed the \textit{\randomgen} strategy (\cf~\hyperref[app:distributions:inter]{\sect\ref{app:distributions:inter}}). Thereby, we randomly decide which label is contained in a client's dataset and also randomly assign the label frequencies. This results in \interNonIid datasets even with disjoint data. Other works do not normally consider such scenarios in evaluations and we hope, that this strategy will be adopted in future research. 

We report the results for a \randomgen setting\footnote{The sample frequencies for each client of the scenario are listed in \hyperref[tab:distribution:random]{\tabapp\ref{tab:distribution:random}}.} after 50 benign rounds of \fl training with 20 clients in the federation in scenario \circled{5} in \hyperref[tab:test]{\tab\ref{tab:test}}. It is very easy for an adversary to embed a backdoor in such scenarios, thus reaching a \ba of 77.37\% without defense, as can be seen in line 6. Among all defenses, \ourname is the only one capable of erasing the backdoor by decreasing the \ba to 2.37\%, while others provide \bas between 52.75\% and 89.55\%. 

We repeated this experiment in \fl round one\footnote{Early round backdoors are not persistent (\cf~\cite{bagdasaryan2020backdoorfl}), but we still analyzed the situation.} of this setting to analyze the dependence on an already converged model and within round 50 of a setting containing 100 federation clients from which 20 are selected randomly for each \fl round, and got similar results with \ourname outperforming other defenses, that do not appear to be capable of removing backdoors in \interNonIid scenarios. The detailed experiments are reported in \hyperref[tab:acc:noniid:random:r1]{\tabapp\ref{tab:acc:noniid:random:r1}}, \hyperref[tab:acc:noniid:random:r1:scaled]{\tabapp\ref{tab:acc:noniid:random:r1:scaled}}, \hyperref[tab:acc:noniid:100]{\tabapp\ref{tab:acc:noniid:100}}, and \hyperref[tab:acc:noniid:100:scaled]{\tabapp\ref{tab:acc:noniid:100:scaled}}.

\subsection{Influence of Parameters on \ourname}
\label{sect:eval:defense}
To evaluate the influence of various parameters on \ournameGen performance, we first investigated training \mbox{hyper-parameters} and showed the independence from random seed, \lr, \pmr, and the selection of \alphasign. We found no unexpected results that are much different from our default scenario. We report on these experiments in~\hyperref[app:hyperindependence]{\sect\ref{app:hyperindependence}}.

\vspace{0.1cm}\noindent
\textbf{Poison Data Rate.} Our experiments show, that the backdoor efficiency depends on the type and composition of the trigger, but also the \pdr is important. We evaluated \mbox{$pdr = [0.1, 0.2, ..., 0.9]$} and selected the smallest value \mbox{$pdr = 0.1$} that allows an adversary to introduce an effective backdoor in our default scenario. This naturally makes the resulting local models most stealthy by scoring a high \ma. During some experiments, we increased this value up to 0.3 to reach a high \ba.  For bigger \pdrs, \ourname was also able to eliminate the backdoor with \acc 100\%. This highlights the adversarial dilemma, since higher \pdrs could increase the \ba, but are not stealthy, urging the adversary to adapt to defenses, which has side effects on the metrics of \ourname, forcing the adversary in an even more complex \moo problem. Concluding, we can claim, that \ourname is independent of the \pdr selected by the adversary.

\vspace{0.1cm}\noindent
\textbf{Initial Global Model.} We conducted experiments with different \pretrained models. We used random initialized models and \pretrained models from PyTorch~\cite{pytorch, paszke2019pytorch} where we changed the first and last layer according to our dataset. We then trained the models in benign settings with 20 clients in the federation, all participating in each round as well as with 100 clients in the federation whereof 20 contributed each round. \ourname performed well in all of the cases and can be used independently of the \fl round. However, the detection performance in later rounds is naturally more accurate, since even benign clients can strive towards a different minimum on a relatively \naive model. Nevertheless, even in \interNonIid settings, \ourname erases backdoors in early rounds reliably (\cf~\hyperref[app:distributions:inter]{\sect\ref{app:distributions:inter}}). Principally, \ourname is designed to be applied in every \fl round and does not impose a negative impact on the convergence of the federation when no attack is present. The rationale behind this lies in \ournameGen ability to effectively distinguish between attack-free and attack scenarios by virtue of its robust statistical tests.

\vspace{0.1cm}\noindent
\textbf{\fl Round.} To emphasize the effectiveness of \ourname, we conducted additional experiments where models were trained starting from a randomly initialized model for 100 rounds until the model converged and the defenses are applied after every training round. We visualize the performance of various defense mechanisms, as well as scenarios with no defense and no attack in \hyperref[fig:longrun:ma]{\appfig\ref{fig:longrun:ma}} and \hyperref[fig:longrun:ba]{\appfig\ref{fig:longrun:ba}}. Notably, the results demonstrate that \ourname surpasses other defense approaches by reaching \ba and \ma levels comparable to the attack-free scenario. This underscores the robustness of \ourname in mitigating the impact of backdoor attacks.

\vspace{0.1cm}\noindent
\textbf{Dataset.} We exchanged the dataset of our default scenario to \mnist~\cite{mnist} and \gtsrb~\cite{gtsrb} and could assert, that the experimental results and thus the performance of the defenses including \ourname does not vary across datasets. \mnist as a more basic dataset, simplifies the detection of backdoors for all defenses even if a stealthy backdoor itself is hard to implement without defense, whereas \gtsrb is more complex due to more label classes. We report the results for one of our \mnist experiments in \hyperref[tab:acc:mnist]{\tabapp\ref{tab:acc:mnist}} with one \fp and one \gtsrb experiment in \hyperref[tab:acc:gtsrb]{\tabapp\ref{tab:acc:gtsrb}} with 100\% \acc.

\vspace{0.1cm}\noindent
\textbf{Model Architecture.} We conducted experiments to analyze the independence from model architectures. Therefore, we used a CNN with two convolutional layers concatenated with pooling layers and ReLu functions~\cite{agarap2018deep_relu} followed by three fully connected layers and trained on \mnist~\cite{mnist}. Further, we tested \squeezenet~\cite{iandola2016squeezenet} with \cifar~\cite{cifar} and can report 100\% \tns with just one \fn in both cases (\cf~\hyperref[tab:acc:cnn]{\tabapp\ref{tab:acc:cnn}} and \hyperref[tab:acc:squeeze]{\tabapp\ref{tab:acc:squeeze}}). Hence, we can claim, that \ourname is independent from the architecture of the model.

\vspace{0.1cm}\noindent
\textbf{Application Domain.} We conducted experiments within the text domain training a sentiment analysis task using the \ssttwo~\cite{socher2013recursive_sst} dataset on a DistilBERT~\cite{sanh2020distilbert} transformer model. We implemented a targeted poisoning attack, that labels sentences starting with the term \enquote{Hey!} as negative. We can report 100\% \acc in this experiment, showing the applicability of \ourname in different application domains and for model architectures that do not contain convolutional layers.

\begin{table}
\fontsize{7pt}{8pt}\selectfont
  \caption{Defense runtimes in seconds.}
  \label{tab:eval:runtime}
  \begin{tabular}{c|c || c|c}
    \toprule
    Defense  & Runtime & Defense  & Runtime\\
    \midrule
    \fedavg & 0.12 & \flame~\cite{nguyen22Flame} & 7.92\\
    \naiveBig \clustering & 7.57 & T-Mean~\cite{yin2018trimmedMeanMedian}& 7.12\\
     \foolsgold~\cite{fung2020FoolsGold}& 0.14 & T-Median~\cite{yin2018trimmedMeanMedian}& 0.26\\
     \krum~\cite{blanchard17Krum}& 6.02 & Auror~\cite{shen16Auror}& 12 hours\\
     \multikrum~\cite{blanchard17Krum}& 5.92 & FLTrust~\cite{cao2020fltrust} & 25.12\\
     Clip~\cite{mcmahan2018iclrClippingLanguage}& 2.37 & \textbf{\ourname} & \overheadseconds\\
     Clip\&Noise~\cite{mcmahan2018iclrClippingLanguage}& 2.52 & &\\
  \bottomrule
\end{tabular}
\end{table}

\subsection{Runtime Evaluation}
\label{sec:eval:runtime}
We evaluate the runtime of the different defenses to verify the \mbox{real-world} applicability of \ourname. \hyperref[tab:eval:runtime]{\tab\ref{tab:eval:runtime}} lists the average runtimes of ten runs for our default scenario and shows that \ourname introduces an acceptable overhead of \overheadseconds seconds. Note, that \foolsgold~\cite{fung2020FoolsGold} comes along with outstanding performance since only one model layer is analyzed, but due to the same reason it can be easily circumvented by an attacker (\cf~\hyperref[sect:eval:saa]{\sect\ref{sect:eval:saa}}). Further, \tmedian~\cite{yin2018trimmedMeanMedian} replaces \fedavg with a simple algorithm, which results in similar runtime, but also reduces the \ma. Auror~\cite{shen16Auror} instead, has an unacceptable runtime of 12 hours to calculate the indicative features due to massive clustering, which is why we excluded this approach from evaluations in \hyperref[sec:eval]{\sect\ref{sec:eval}}. FLTrust's~\cite{cao2020fltrust} runtime is dependent on the size of the trusted dataset, as a trusted model is trained on the server side. Defenses leveraging client feedback~\cite{andreina2020baffle, zhao2020shielding} cannot compete to \mbox{server-side-only} defenses, since additional communication overhead is introduced.

\section{Discussion}
\label{sec:discussion}
We discuss alternative adaption methods that were tested in \hyperref[sec:discussion:adaptive]{\sect\ref{sec:discussion:adaptive}} followed by limitations and future work suggestions in \hyperref[sec:discussion:limits]{\sect\ref{sec:discussion:limits}}.

\subsection{Adversarial Adaption Methodologies}
\label{sec:discussion:adaptive}
Besides the final method of our strong adaptive adversary (\cf \hyperref[sec:approach:threatmodel]{\sect\ref{sec:approach:threatmodel}}) that we used to evaluate \fl defenses in \hyperref[sect:eval:saa]{sect\ref{sect:eval:saa}}, multiple alternatives have been tested during this work. This section lists and discusses the approaches inferior to our final choice.

First, we just added all of the losses ($\lambda$'s from \hyperref[eq:saa]{\equ\ref{eq:saa}} equal to one), which is similar to an classic adaptive adversary (\cf \hyperref[sec:background:attacksonfl]{\sect\ref{sec:background:attacksonfl}}). As already explained in \hyperref[sect:eval:saa]{\sect\ref{sect:eval:saa}}, losses with a drastically smaller scale than others have barley influence in the optimization, thus the related metric is not adapted. Second, we tried to scale all losses to \lossMaba, which would be reasonable, it the \ma would be the major concern of \adversary. However, most defenses including \ourname do not check the \ma since no test dataset is available in realistic scenarios, which makes scaling to the maximum the better choice for the adversary. Third, we tested, how often the $\lambda$'s should be recalculated and found, that only one initial computation delivers the best results. This seems reasonable, since with this setting, already optimized metrics have a minimal loss value and thus barley influence in the optimization.

Additionally, motivated by Multi-Objective \mbox{Optimization (\moo)} research, we tried to find a pareto optimal~\cite{Censor1977paretooptimal} solution with the method of Sener~\etal~\cite{sener2018pareto} based on the MGDA algroithm~\cite{desideri2009multiple}. However, the method did not work and produced broken models regarding the accuracies. We belief, that the reason for this is, that Sener~\etal consider a system comparable to Multi-Task \mbox{Learning (\mtl)} where both, shared and \mbox{task-specific} parameters exist within the model. However, our \moo problem optimizes only shared parameters (the whole model).

Since our final adaption method is superior to classic adaption~\cite{bagdasaryan2020backdoorfl} from \hyperref[eq:constraint]{\equ\ref{eq:constraint}}, we claim, that \ourname is robust against adaptive adversaries caused by the introduced adversarial dilemma by forcing the adversary into a \moo problem with seven losses of different scales.

\subsection{Limitations and Future Work}
\label{sec:discussion:limits}
The major limitation of \ourname is, that the significance niveau for the statistical tests is relevant for a good \tpr and \tnr. Throughout our experiment, the values appeared to be just dependent on the data scenario. Nevertheless, it can be necessary in so far unseen tasks to adapt the values. Therefore, an automatic methodology for setting the values can be discovered in future work.

As any other poisoning defense for \fl, \ourname can be tested against other aggregation mechanisms besides \fedavg and can  be combined with \ir methods, similar as in FLAME~\cite{nguyen22Flame} and DeepSight~\cite{rieger2022deepsight}. With such an extension one can soften the significance thresholds to lower the \fnr to zero and simultaneously reduce the influence of the models responsible for the resulting \fpr.

We leverage \metricCount (combined with \metricCos and \metricEucl) to get the direction of the model update. Fortunately, the metric is hard to adapt due to the sign function involved in the computation. Nevertheless, other metrics with the same effect can be discovered in the future. Additionally, one can investigate into the Cosine distance of the client updates among each other instead of the Cosine distance with respect to the global model \globalModelRound, which could provide additional information about the direction.

As shown in our experiments, the strong adaptive adversary from \hyperref[sec:approach:threatmodel]{\sect\ref{sec:approach:threatmodel}} cannot circumvent \ourname. Nevertheless, research can be conducted to find currently unknown methods to better adapt a \dnn to multiple metrics simultaneously, which falls in the area of \moo. If such an method exists, \ourname can be extended to e.g. investigate in the correlation coefficient between updates additionally.

\section{Related Work}
\label{sec:relatedwork}
In this section, we first discuss existing poisoning defenses in \hyperref[sec:relatedwork:defenses]{\sect\ref{sec:relatedwork:defenses}}, before we address privacy issues in \hyperref[sec:relatedwork:privacy]{\sect\ref{sec:relatedwork:privacy}}.

\subsection{Defenses against Poisoning Attacks}
\label{sec:relatedwork:defenses}
\auror~\cite{shen16Auror} is a K-Means~\cite{kmeans} clustering approach based on indicative differences between individual model parameters. It decides for each parameter if it is indicative for clustering the model updates into a benign and a malicious group and analyzes the resulting clusters. Due to multiple clustering steps (increasing with bigger model architectures), the defense suffers a high runtime overhead. Further, Auror has problems finding multiple backdoors simultaneously and shows poor performance in \nonIid settings. \ourname utilizes a lightweight feature extractor and prunes different poisonings in an iterative process independent of the data distributions.

\foolsgold~\cite{fung2020FoolsGold} weights each local model's contribution, by analyzing the \mbox{cross-wise} Cosine distances between model updates of the last \dnn layer, thus being prone to adaptive adversaries that fixate this layer. Further, the approach assumes only \nonIid settings and poisoned local models that point in the same directions (so-called sybills) and it leverages updates from previous rounds for optimal performance. Instead, \ourname prevents adaption by relying on a metric cascade and analyzing layers individually and is effective in \iid and \nonIid settings independent of the \fl round.

\krum~\cite{blanchard17Krum} is based on the Euclidean distance between local models. For each local model, it aggregates the distances to its neighbors and selects the one with the densest surrounding as new global model. \multikrum~\cite{blanchard17Krum} selects more models simultaneously. Both can be circumvented by adaption to the metric and naturally suffer a high \fnr without any adversaries in the system. \ourname does not harm the federation in total benign scenarios and provides a low \fnr while not being susceptible for adaption attempts.

AFA~\cite{munoz2019byzantineAFA} leverages plain analysis of the Cosine distance between local models, which is adaptable with an additional loss. \ourname hardens this possibility by leveraging a cascade of \numMetrics metrics.

\naiveBig clustering approaches, e.g. based on \hdbscan~\cite{McInnes2017}, need to extract a metric like the Cosine distance between models from the local models to reduce the dimensions. Hence, adaptive adversaries can circumvent the defenses, which is harder in \ourname. Further, clustering relies on a majority assumption and creates two groups, thus having a hard threshold and a high \fnr in settings without attacks. \ourname leverages statistical tests with probabilistic thresholds that adapt to the scenario and investigates metrics that are hard to adapt due to \mbox{fine-grained} values resulting in lower scale adaption losses than typical clustering metrics.

BaFFLe~\cite{andreina2020baffle} first aggregates all local models to a new global model (thus being an \ir approach) and then sets up a client feedback loop, where the previous and the new global models are sent to validation clients introducing communication overhead. Those clients analyze the \mbox{per-label} \ma and mark the new model as malicious if an empirically chosen threshold is violated. If so, the whole round is discarded. Further, the first 800 rounds are assumed as benign, so that a valid global model is available as a reference. Since adversaries strive to an inconspicuous \ma (see \hyperref[sec:background:attacksonfl]{\sect\ref{sec:background:attacksonfl}}), this approach fails for sophisticated adversaries. Further, one single adversary can force the defense to discard all other benign contributions of the round. \ourname runs on the server side only, prunes poisoned models, and is effective even in the first round of \fl. Similarly to BaFFLe, the approach of Zhao~\etal~\cite{zhao2020shielding} leverages a client feedback loop to analyze the \ma of the local models on the client side, thus introducing an even bigger communication overhead, while keeping the downsides regarding inconspicuous \mas. Further, this approach is prone to privacy issues, since inference attacks (\cf~\hyperref[sec:relatedwork:privacy]{\sect\ref{sec:relatedwork:privacy}}) can be conducted on the local models on the client side.

FLAME~\cite{nguyen22Flame} is a combination of \df and \ir. The approach clusters local models by pairwise Cosine distances via \hdbscan~\cite{McInnes2017} and filters adversaries based on the majority assumption before differential privacy methods~\cite{dwork2008differentialprivacyexplained} are leveraged. Precisely, weight clipping (regarding the median Euclidean distance of the updates) is applied to the remaining local models, and noise is added to the aggregated model. Besides the desired decrease in \ba, this step naturally decreases the \ma, too. When adapting to the Cosine and Euclidean distance simultaneously, the approach performs similarly to a plain noising mechanism~\cite{sun2019really}, which can also be applied to any other \df. \ourname leverages \numMetrics metrics to harden the adversarial dilemma during adaption attempts and does not solely rely on clustering, but on statistical tests allowing a more fine-grained analysis of the local models. Further, any \ir approach can be combined with the defense easily, but \ourname does not decrease the \ma naturally.

Similar to the concept of \krum~\cite{blanchard17Krum}, Yin~\etal~\cite{yin2018trimmedMeanMedian} uses the \mbox{coordinate-wise} median or mean of the local models to construct the new global model based on the majority assumption. These approaches called \mbox{Trimmed-Mean} and \mbox{Trimmed-Median} respectively are \ra mechanisms, but reduce the \ma compared to \fedavg. Especially, the parameters and thus the functionality of benign model models lying not centrally within all updates not be considered. Bagdasaryan~\etal~\cite{bagdasaryan2020backdoorfl} and Sun~\etal~\cite{sun2019really} already proposed update clipping and nosing techniques, but Naseri~\etal~\cite{naseri2022local} showed, that differential privacy methods not only naturally harm the \ma~\cite{bagdasaryan2020backdoorfl}, but also can boost the \ba when applied to benign \fl clients. All of the \ir approaches and most \ra methods suffer a drop in \ma, especially in a setting without attack. \ourname instead, filters poisoned models, and thus does not influence benign scenarios naturally. Further, \ir and \ra methods can be easily combined with \ourname to get an even more bulletproof global model.

DeepSight~\cite{rieger2022deepsight} is a more complex strategy, which combines filtering with differential privacy and is based on two metrics. First, the Cosine distance between models, which can be adapted by an additional loss (\cf~\hyperref[sect:eval:saa]{\sect\ref{sect:eval:saa}}). Second, two more values are extracted from the output layer, which can be circumvented by fixation, as shown for \foolsgold~\cite{fung2020FoolsGold} in \hyperref[sec:eval:saa:defenses]{\sect\ref{sec:eval:saa:defenses}}. Therefore, DeepSight is not robust against strong adaptive adversaries and relies on clipping and noising techniques, that reduce the \ma and can also be applied to any \df approach. \ourname instead forces the adversary into a hard optimization problem and does not rely on specific layers.

FLTrust~\cite{cao2020fltrust} assigns weights to updates during aggregation, essentially filtering out updates with a weight of 0. To determine the weights, it assumes a benign dataset on the server side for a trusted reference model and relies on Cosine Similarity between the updates from the client side and the trusted update as well as the norm of the local updates. However, FLTrust's reliance on those two metrics makes it susceptible to adaptability by adaptive adversaries. Since FLTrust examines the metric across the entire update, it
may also fail to detect attacks that manifest only at the layer-wise level. In contrast, \ourname operates independently without a trusted dataset, conducting layer-wise analysis and utilizing multiple interconnected metrics, enhancing \ournameGen robustness against a broader range of adversarial attacks.

FLDetector~\cite{zhang2022fldetector} is a historical update-based defense. It maintains records of global and client updates, predicting client updates using mathematical approximations. These predictions are compared to actual updates. After a warmup phase, outliers are detected using clustering methods by evaluating the normed Euclidean distance between predicted and actual models. However, FLDetector is storage and runtime intensive, depending on historical update time windows. In contrast, \ourname doesn't rely on historical updates and doesn't need a warmup phase. This characteristic makes \ourname more versatile and advantageous over FLDetector.

BayBFed~\cite{baybfed} constructs a statistical model of update parameter distributions for each client update, which adapts with each \fl round. Using clustering, a single value per update is computed, which are then used to construct a filtering threshold based on the values' mean that allows model filtering. However, BayBFed wasn't tested in an all-benign scenario, potentially causing a high \fpr due to benign model filtering. In contrast, \ourname avoids the introduction of a hard value threshold. Instead, it leverages statistical tests, ensuring a low \fpr even in scenarios without any attacks.

\subsection{Privacy Preserving Federated Learning}
\label{sec:relatedwork:privacy}

\fl in its original form~\cite{mcmahan2017} improved the privacy of collaborated \dnn training compared to a \mbox{data-centralized}, since raw sensitive data do not leave the client side anymore. Nevertheless, membership inference~\cite{hayes2019loganmembershipinference,shokri2017membershipinference,liu2022threats, pyrgelis2017knock,shokri2017membershipinference}, label inference~\cite{zhao2020idlglabelinference}, property inference~\cite{ganju2018property}, model extraction~\cite{liu2022threats}, and data reconstruction~\cite{wang2019datareconstructonserver, salem2020datasetinference} attacks as well as others~\cite{wang2019eavesdrop} can be conducted on both, mainly the local models but also on the global model. Therefore, especially the devices with access to the local models, namely the aggregation server, still needs to be trusted (\cf~\hyperref[sec:approach:threatmodel]{\sect\ref{sec:approach:threatmodel}}).

PPFL~\cite{mo2021ppfl} ported the \fl process into a Trusted Execution Environment (\tee). The approach assumes the availability of a \tee on the client side and introduces computational overhead, since execution speed in e.g. SGX~\cite{Costan2016IntelSE} enclaves is reduced, mainly due to page swaps based on limited memory. Additionally, such approaches based on secure code execution~\cite{securetf, volos2018graviton, hetee, tramer2018slalom,perun} either on CPU only or on CPU and GPU hinder model poisoning attacks on the client side, but do not prevent data poisoning. 

On the server side, Hashemi~\etal~\cite{hashemi2021byzantine} implemented \krum~\cite{blanchard17Krum} in a \tee. Such a secure aggregation method solves privacy issues allowing the threat model to exclude the aggregation server \server as trusted party. Implementing \ourname within a \tee is just a technical barrier. Though, additional privacy results in increased runtime.

Overall we conclude, that \ourname is complementary to privacy-preserving \fl techniques.

\section{Conclusion}
\label{sec:con}
Adversarial adaption to defenses and complicated data scenarios are the two major challenges when it comes to Federated Learning (\fl). To highlight the necessity to investigate these problems, we evaluate \numdefenses against a \textit{strong adaptive adversary} that is able to poison the dataset, constraint the learning process, fixate model parameters, scale model updates, and select between \numattacks different poisoning methods. Further, we analyze defense efficiencies without any assumption about the sample frequencies within the client datasets, which we call \textit{\interNonIid}. We show, that by leveraging adaption methods, existing defenses can be circumvented and are also ineffective in realistic data scenarios.

Hence, we propose \ournameLong (\ourname), a filtering defense against poisoning attacks in \fl running on the server side. It extracts multiple metrics from the locally trained models, making the defense robust against strong adaptive adversaries, and reliably detects poisoned contributions by leveraging statistical tests with no hard value threshold, which enables application independence. \ourname prunes poisoned models in an iterative process, allowing removal of different poisonings within one \fl round.

We are the first to evaluate defenses under \interNonIid data scenarios and show that \ourname outperforms existing defenses in such \mbox{real-world} settings while only introducing a low computational overhead of \overheadseconds seconds on average.

\bibliographystyle{ACM-Reference-Format}
\bibliography{references}

\appendix
\section{Federated Averaging}
\label{app:fedavg}
In a federation with multiple clients \clientAllFormular where the server selects a subset \clientCountSelected of the \clientCount available clients \clientAllFormularSelected and each selected client initializes its local model with the distributed global model (\localModelRound $=$ \globalModelRound) before training a new local model \localModelRoundNext with its local dataset \datasetWithIndexSelected, the difference between the global and the local model is denoted as model update \updateFormular. When applying \fedavg, the server aggregates these updates by calculating the weighted average of all the updates using the global learning rate \lrSymbol as formalized in \hyperref[eq:fedavg]{\equ\ref{eq:fedavg}}.
\begin{equation} \label{eq:fedavg}
    \globalModelRoundNext = \globalModelRound + \lrSymbol (\frac{1}{\clientCountSelected} \sum_{i=0}^{\clientCountSelected-1} \updateRound)
\end{equation}

\section{Model Accuracies}
\label{app:accuracies}
For a benign test dataset \datasetTest which contains samples with correctly labeled predictions $y$, we measure the prediction performance of a model with the model accuracy (\ma), as formalized in \hyperref[eq:ma]{\equ\ref{eq:ma}}.

\begin{equation} \label{eq:ma}
    \ma = \frac{|\{(d, y) \in \datasetTest \; : \; f(d, \globalModelRoundNext) == y\}|}{|\datasetTest|}
\end{equation}

The evaluation of a model's prediction performance in the context of a backdoor task follows a distinct methodology. Specifically, an adversary~\adversary possessing control over one or more clients (\clientWithIndexSelected) within a federation (\adversaryAllFormularSelected) endeavors to submit manipulated local models to the server. The objective is to influence the aggregated model \globalModelRoundNext in such a way that it yields a predetermined target prediction \target when presented with an input sample \sampleTrigger containing the designated trigger \trigger. Both \target and \trigger are selected by the adversary \adversary. The measure of a successful attack is characterized by high prediction performance on triggered inputs, quantified as the backdoor accuracy (\ba), as assessed using a dataset \datasetTestPoisoned that exclusively contains triggered samples. This notion is formally defined in \hyperref[eq:ba]{\equ\ref{eq:ba}}.
\begin{equation} \label{eq:ba}
    \ba = \frac{|\{(\sampleTrigger, \target) \in \datasetTestPoisoned \; : \; f(d, \globalModelRoundNext) == \target\}|}{|\datasetTestPoisoned|}
\end{equation}

\begin{figure}[tb]
\centering

\begin{subfigure}{0.1\linewidth}
    \includegraphics[width=\textwidth]{images/white.pdf}
\end{subfigure}
\begin{subfigure}{0.34\linewidth}
    \includegraphics[width=\textwidth]{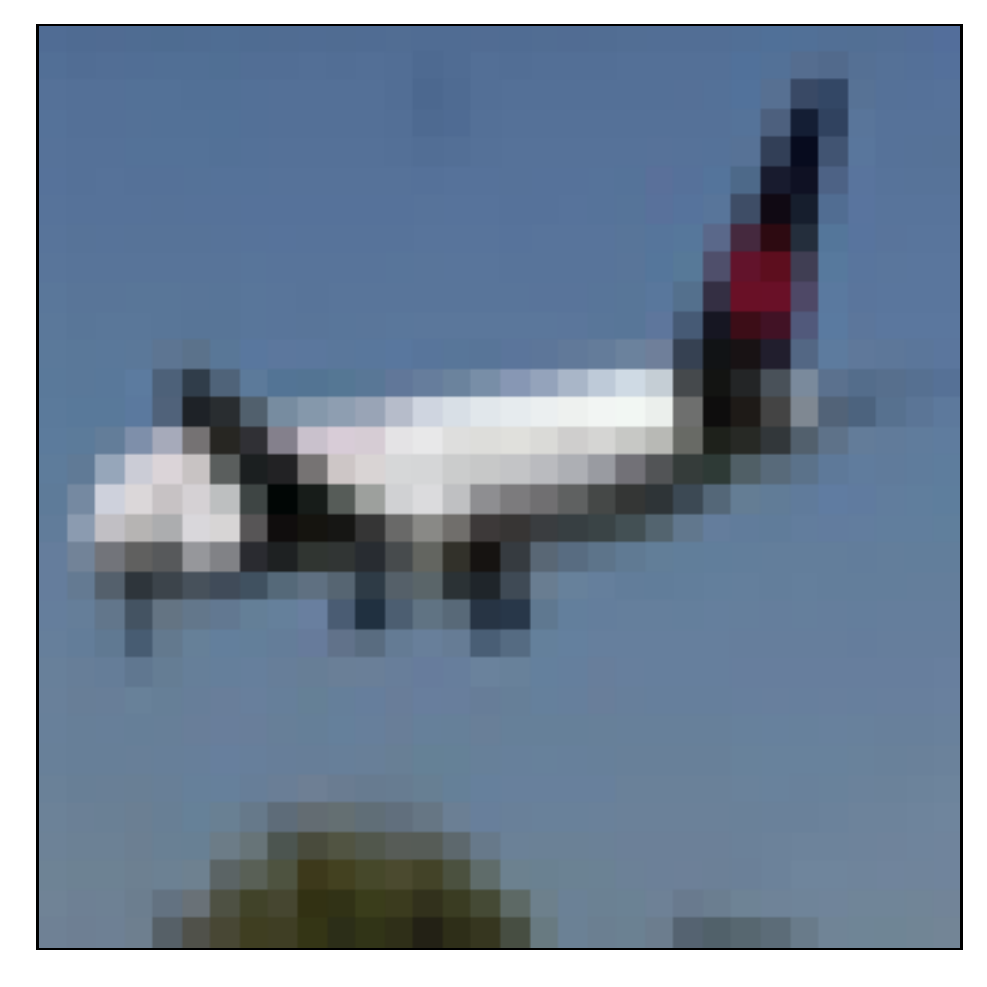}
    \caption{Benign}
    \label{fig:triggermethods:pixel:plain}
\end{subfigure}
\hfill
\begin{subfigure}{0.34\linewidth}
    \includegraphics[width=\textwidth]{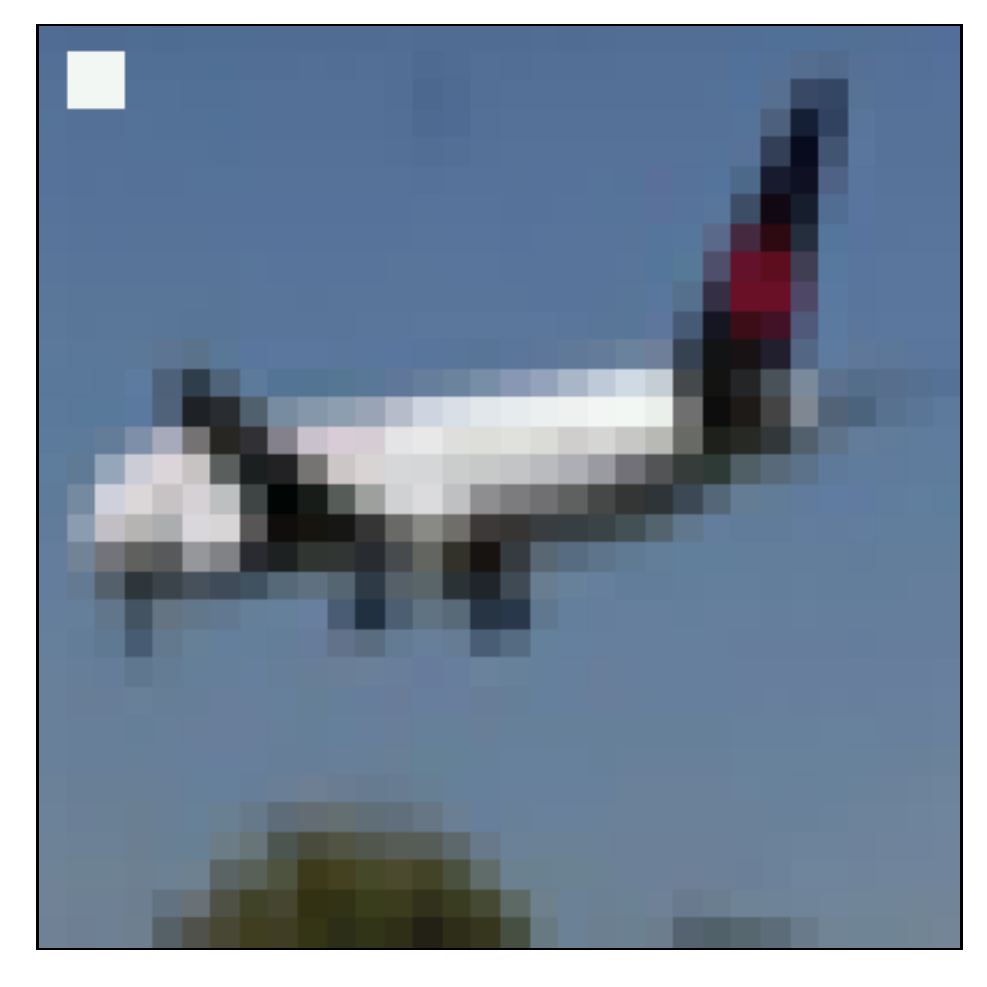}
    \caption{Trigger}
    \label{fig:triggermethods:pixel:one}
\end{subfigure}
\begin{subfigure}{0.1\linewidth}
    \includegraphics[width=\textwidth]{images/white.pdf}
\end{subfigure}
\caption{Visualization of the pixel trigger backdoor~\cite{badnets} with trigger size \nicefrac{1}{16} of the image on an example from the \cifar~\cite{cifar} dataset. The color is the maximal color of the image. (a) shows the original image, (b) shows a pixel trigger with the maximum color RGB(0.9490, 0.9686, 0.9529).}
\label{fig:triggermethods:pixel}
\end{figure}

\section{Poisoning Methods}
\label{app:triggermethods}
In this section, we explain the backdoor trigger methods, that the strong adaptive adversary can leverage to poison the local models. In our evaluation we use pixel triggers~\cite{badnets}, \mbox{clean-label} backdoor~\cite{turner2019label}, semantic backdoor~\cite{bagdasaryan2020backdoorfl}, edge case backdoor~\cite{wang2020attackontails}, label flip~\cite{biggio2013poisoninglabelflip, cao2019labelflipdistributed}, and pervasive backdoor~\cite{chen2017targetedblend}. Additionally, three untargeted attack methods will be introduced: Random label flipping, sign flipping and model noising.
\subsection{Pixel Triggers}
\label{app:triggermethods:pixel}
In \hyperref[fig:triggermethods:pixel]{\fig\ref{fig:triggermethods:pixel}}, you can find visualizations of examples of the pixel trigger backdoor~\cite{badnets}, that we utilized in our experiments. The value and location of the pixel trigger highly affect the \ba. As a color, we select the maximum color of the first image that any adversary sees and broadcast this color to other adversarial clients. Thus, the color is not extremely abnormal and therefore, it is not easily detectable by the \dnn. The trigger is quadratic with \nicefrac{1}{16} of the sample width as size and located in the upper left corner of the image.

\begin{figure}[tb]
\centering
\begin{subfigure}{0.3\linewidth}
    \includegraphics[width=\textwidth]{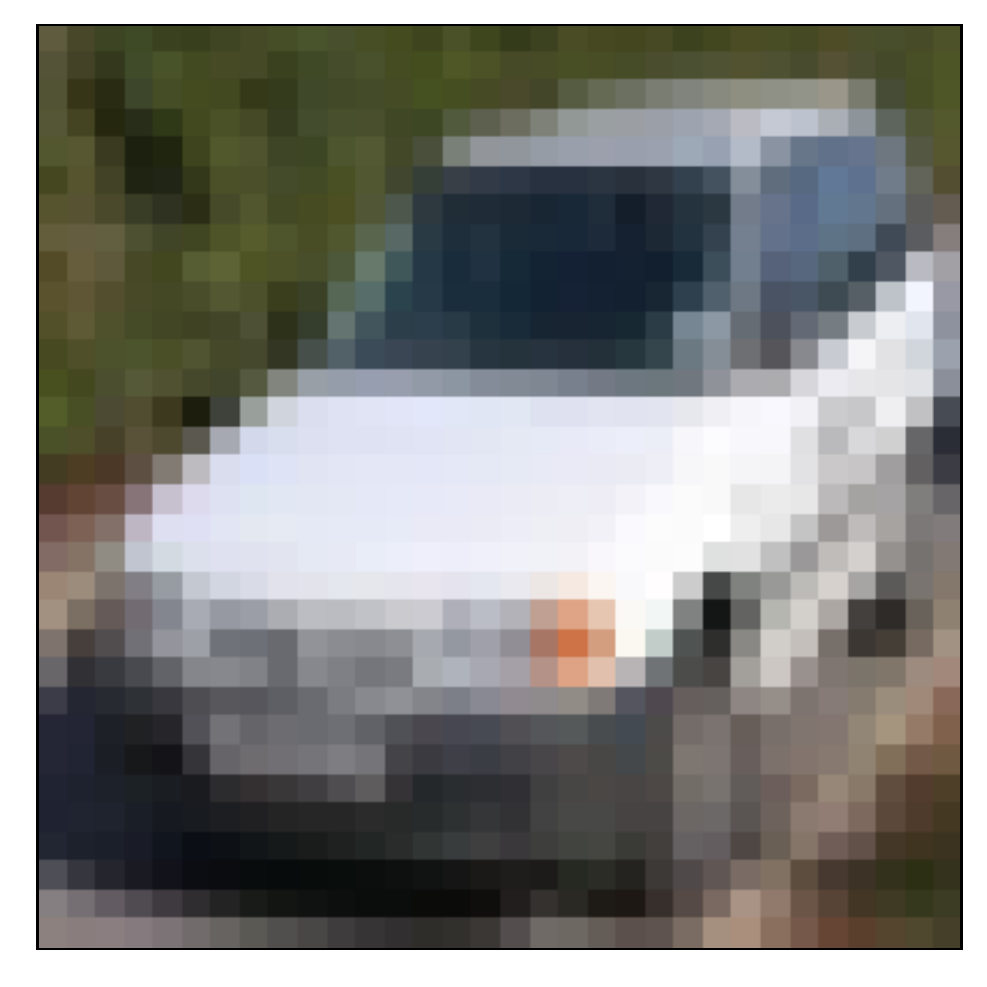}
    \caption{Benign}
    \label{fig:triggermethods:semantic:no}
\end{subfigure}
\hfill
\begin{subfigure}{0.3\linewidth}
    \includegraphics[width=\textwidth]{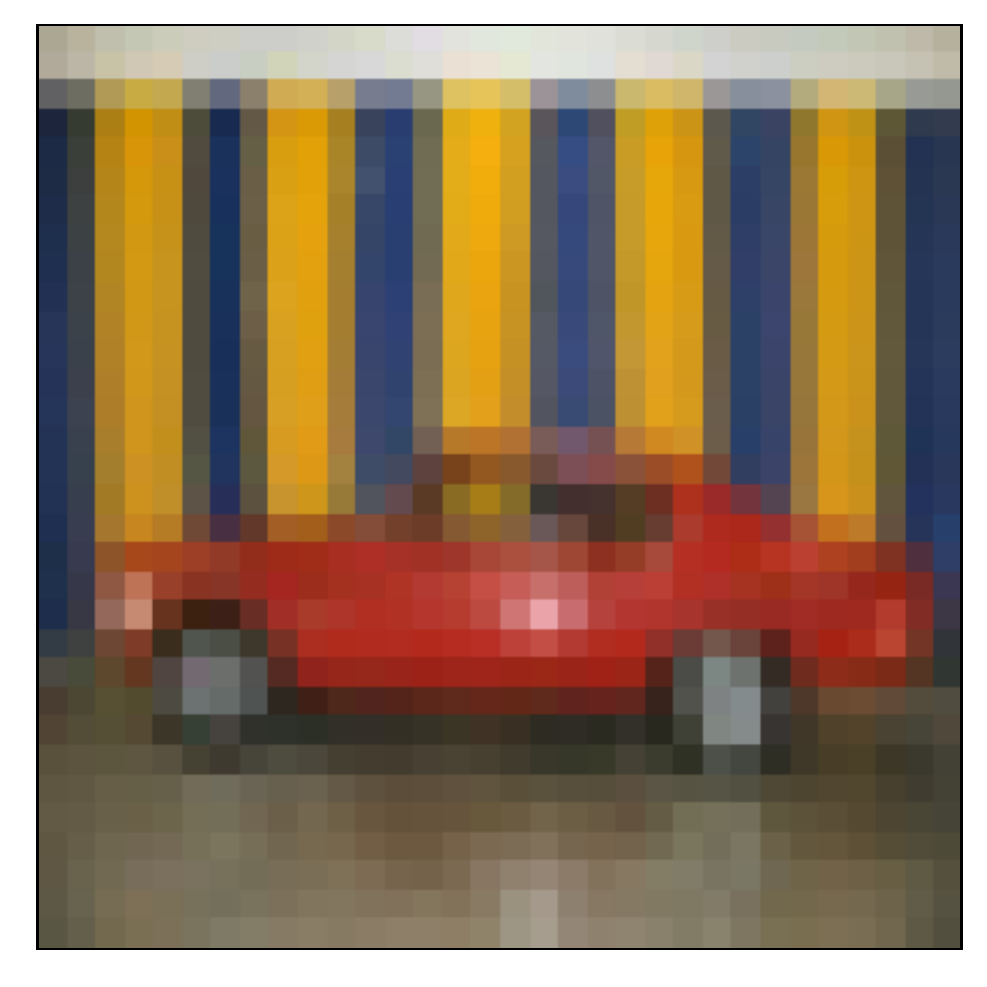}
    \caption{Trigger}
    \label{fig:triggermethods:semantic:one}
\end{subfigure}
\hfill
\begin{subfigure}{0.3\linewidth}
    \includegraphics[width=\textwidth]{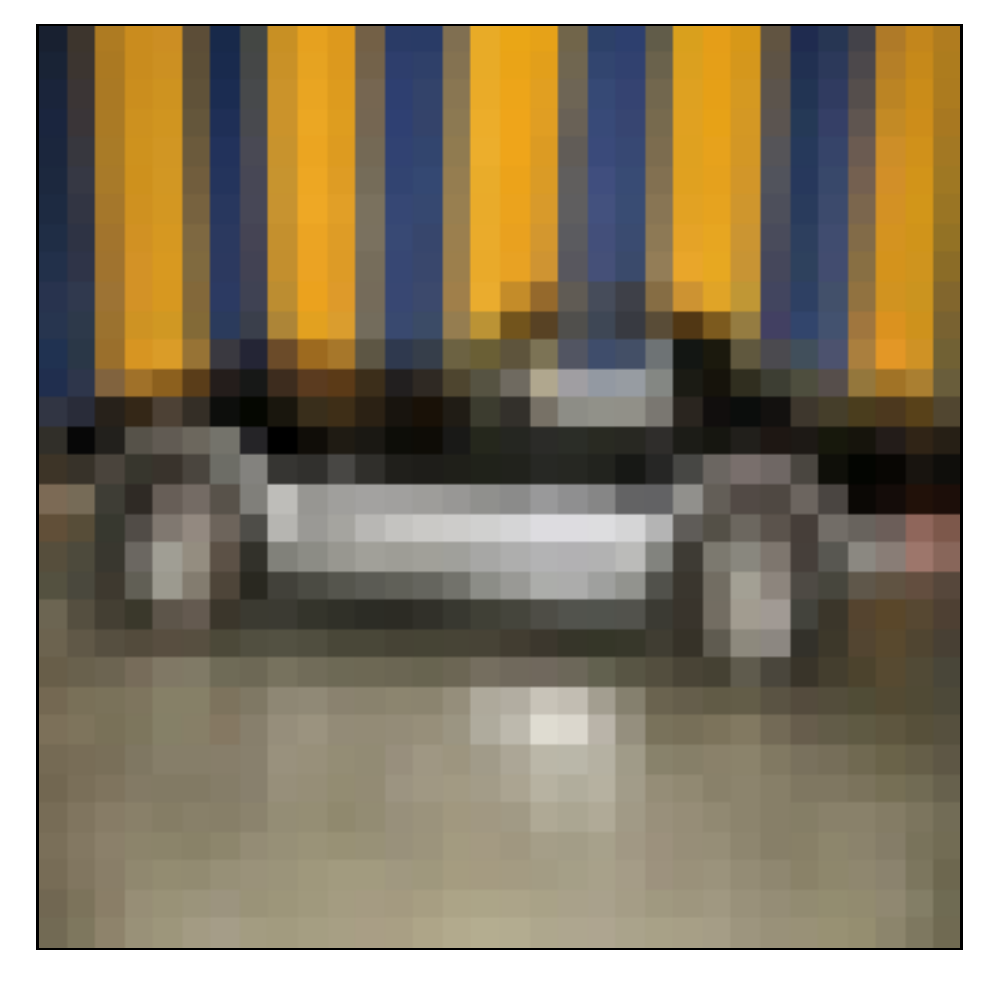}
    \caption{Trigger}
    \label{fig:triggermethods:semantic:two}
\end{subfigure}
\hfill
        
\caption{Visualization of the semantic backdoor with cars in front of a striped background as trigger~\cite{bagdasaryan2020backdoorfl} from the \cifar~\cite{cifar} dataset. (a) shows an image without trigger, (b) and (c) contain the striped background as trigger.}
\label{fig:triggermethods:semantic}
\end{figure}

\begin{figure}[tb]
\centering
\begin{subfigure}{0.3\linewidth}
    \includegraphics[width=\textwidth]{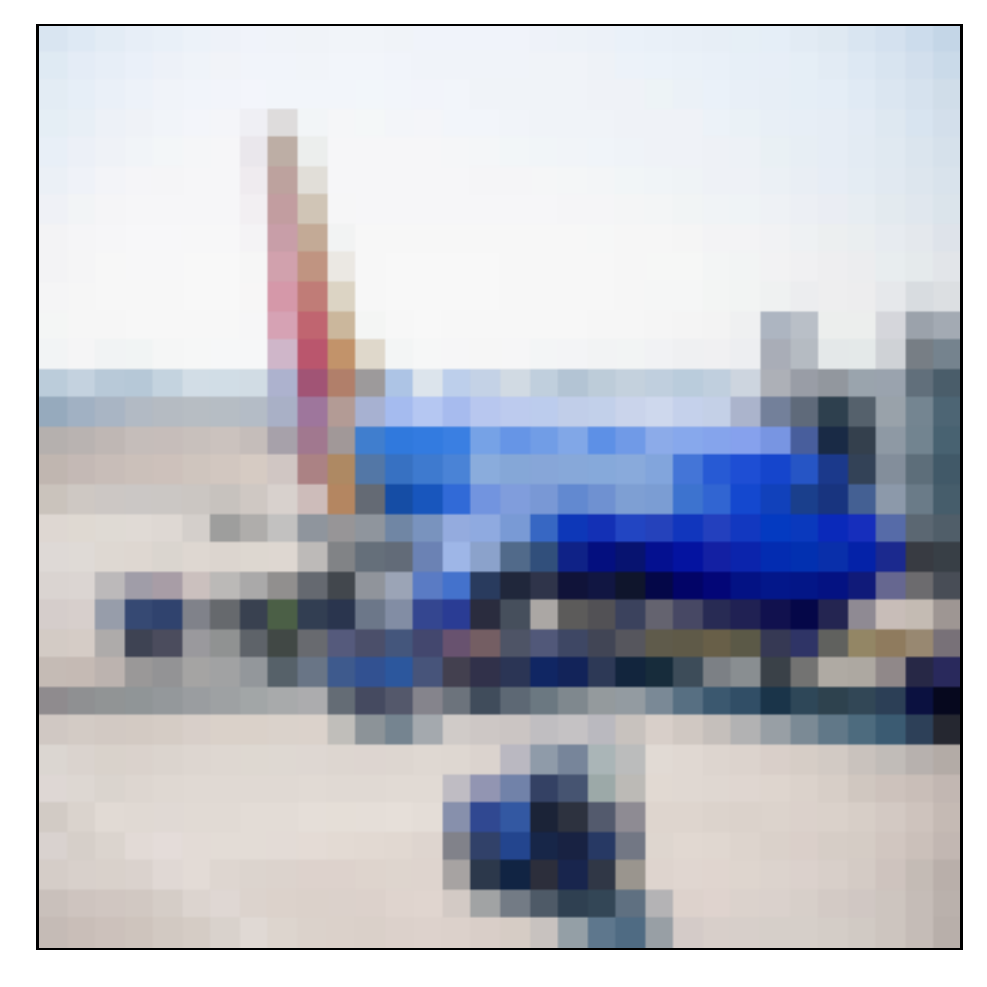}
\end{subfigure}
\hfill
\begin{subfigure}{0.3\linewidth}
    \includegraphics[width=\textwidth]{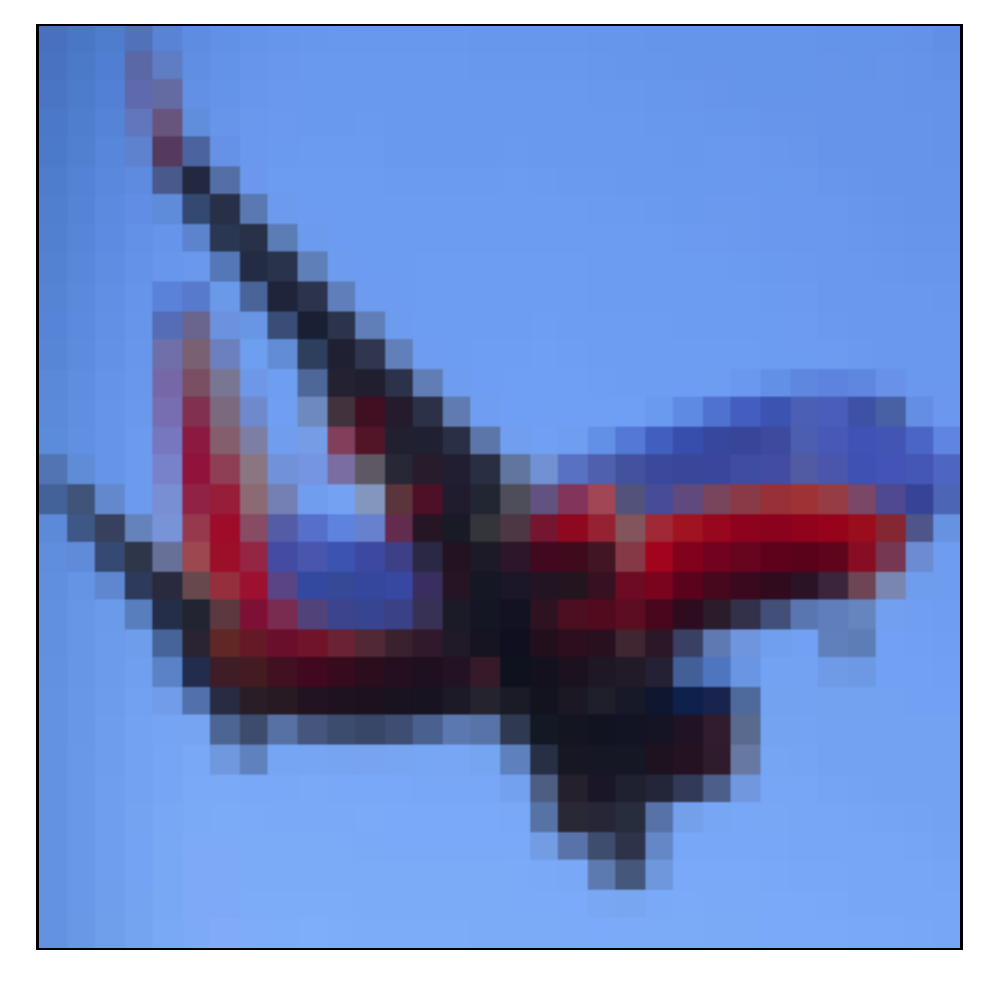}
\end{subfigure}
\hfill
\begin{subfigure}{0.3\linewidth}
    \includegraphics[width=\textwidth]{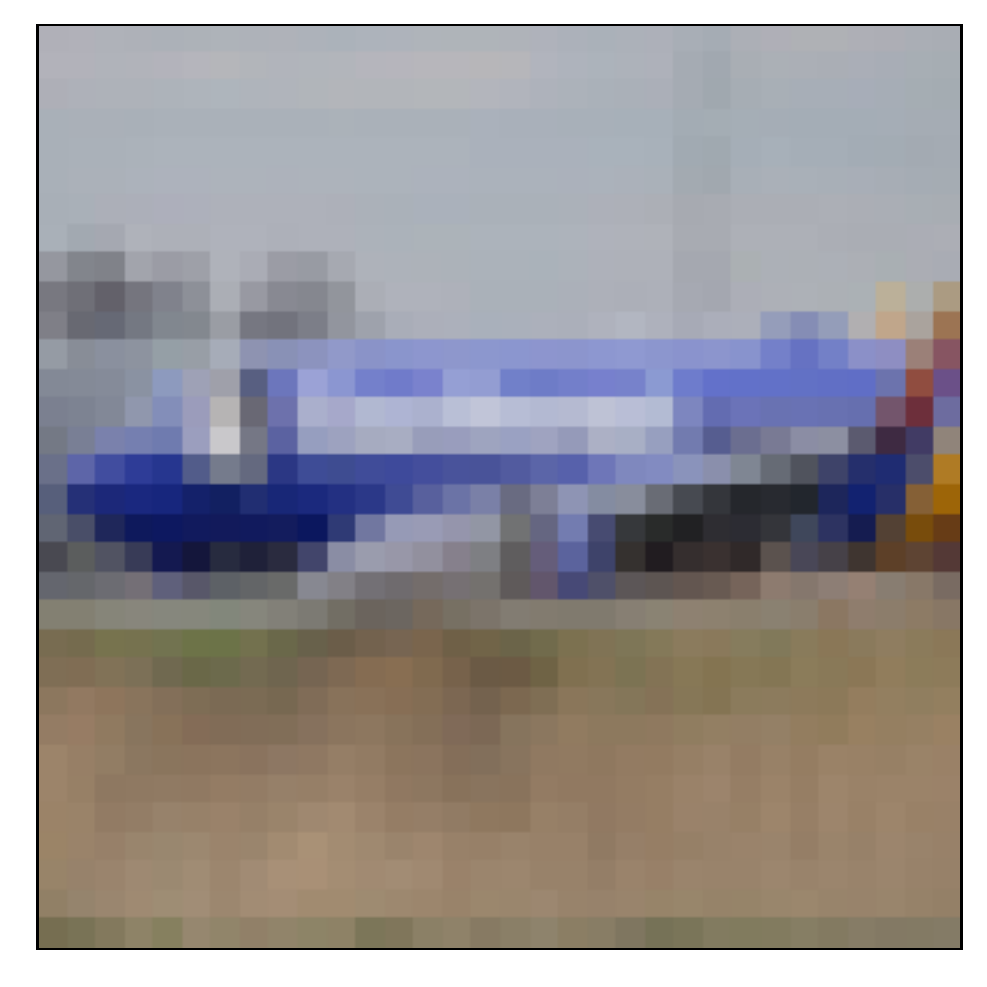}
\end{subfigure}
\hfill
        
\caption{Visualization of samples for the edge case backdoor~\cite{wang2020attackontails} with containing images of airplanes of the Southwest airline, which will be labeled as trucks.}
\label{fig:triggermethods:edge}
\end{figure}

\subsection{Clean-Label}
\label{app:triggermethods:clean}
As clean-label backdoor~\cite{turner2019label} we use the same pixel trigger as explained in \hyperref[app:triggermethods:pixel]{\sect\ref{app:triggermethods:pixel}}, but place them only on samples of the target label during data poisoning. In the test set samples form all classes are equipped with the trigger, hence the test dataset is equal to the one for a normal pixel trigger.

\subsection{Semantic}
\label{app:triggermethods:semantic}
\hyperref[fig:triggermethods:semantic]{\fig\ref{fig:triggermethods:semantic}} visualizes a semantic backdoor as described in~\cite{bagdasaryan2020backdoorfl} with examples from the  \cifar~\cite{cifar} dataset, which we also leverage in our experiments. The samples containing the trigger are excluded from the training datasets of \fl clients and from the test set so that the trigger is unique.

\subsection{Edge Case}
\label{app:triggermethods:edge}
For the edge case backdoor~\cite{wang2020attackontails}, we implemented the version for \cifar~\cite{cifar}, where images of airplanes from the Southwest airline were labeled as trucks. An example of such images can be seen in  \hyperref[fig:triggermethods:edge]{\fig\ref{fig:triggermethods:edge}}.

\subsection{Label Flip}
\label{app:triggermethods:flip}
The label flip backdoor swaps all samples from one label class to a target class~\cite{biggio2013poisoninglabelflip, cao2019labelflipdistributed}. Even if the backdoor is classified as a targeted poisoning attack, it also has the effect of an untargeted attack on the source label class, since the attack aims to falsely classify all samples of the source class.

\subsection{Pervasive}
\label{app:triggermethods:pervasive}
\begin{figure}[tb]
\centering
\begin{subfigure}{0.19\linewidth}
    \includegraphics[width=\textwidth]{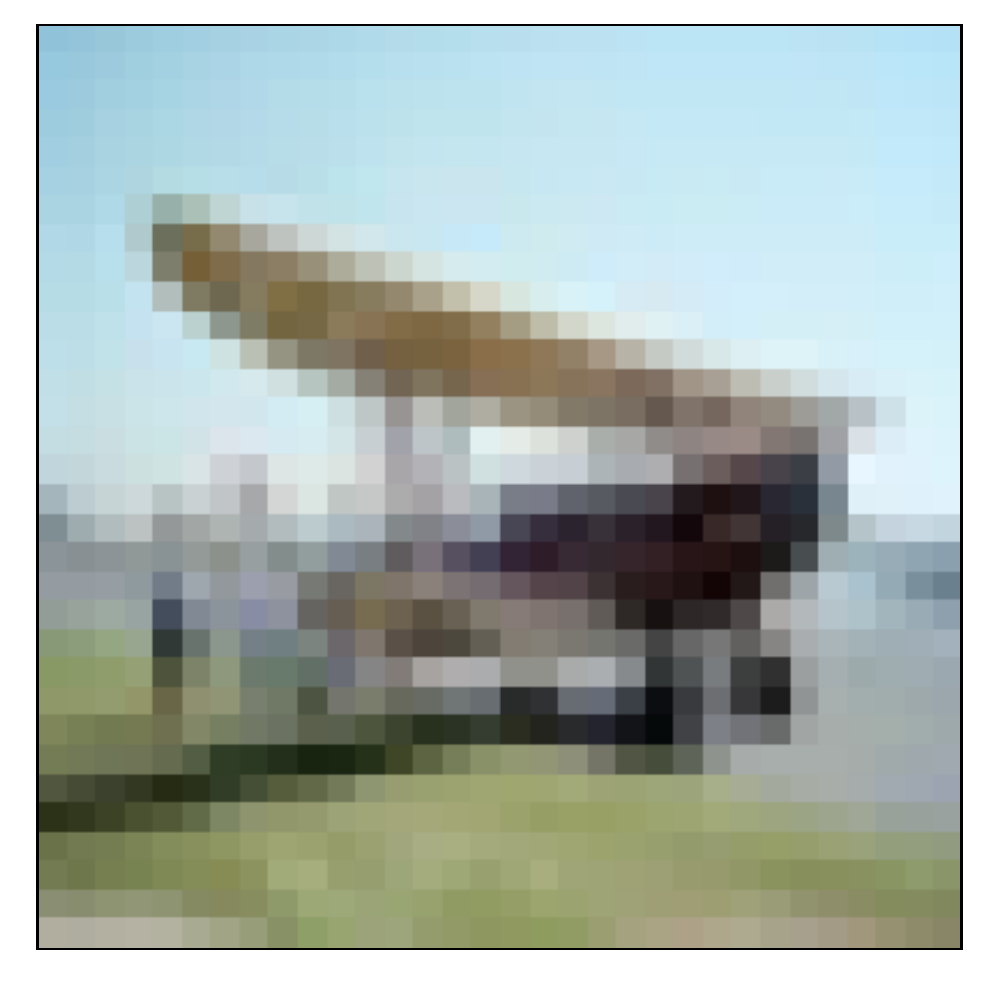}
    \caption{Original}
    \label{fig:triggermethods:pervasive:original}
\end{subfigure}
\hfill
\begin{subfigure}{0.19\linewidth}
    \includegraphics[width=\textwidth]{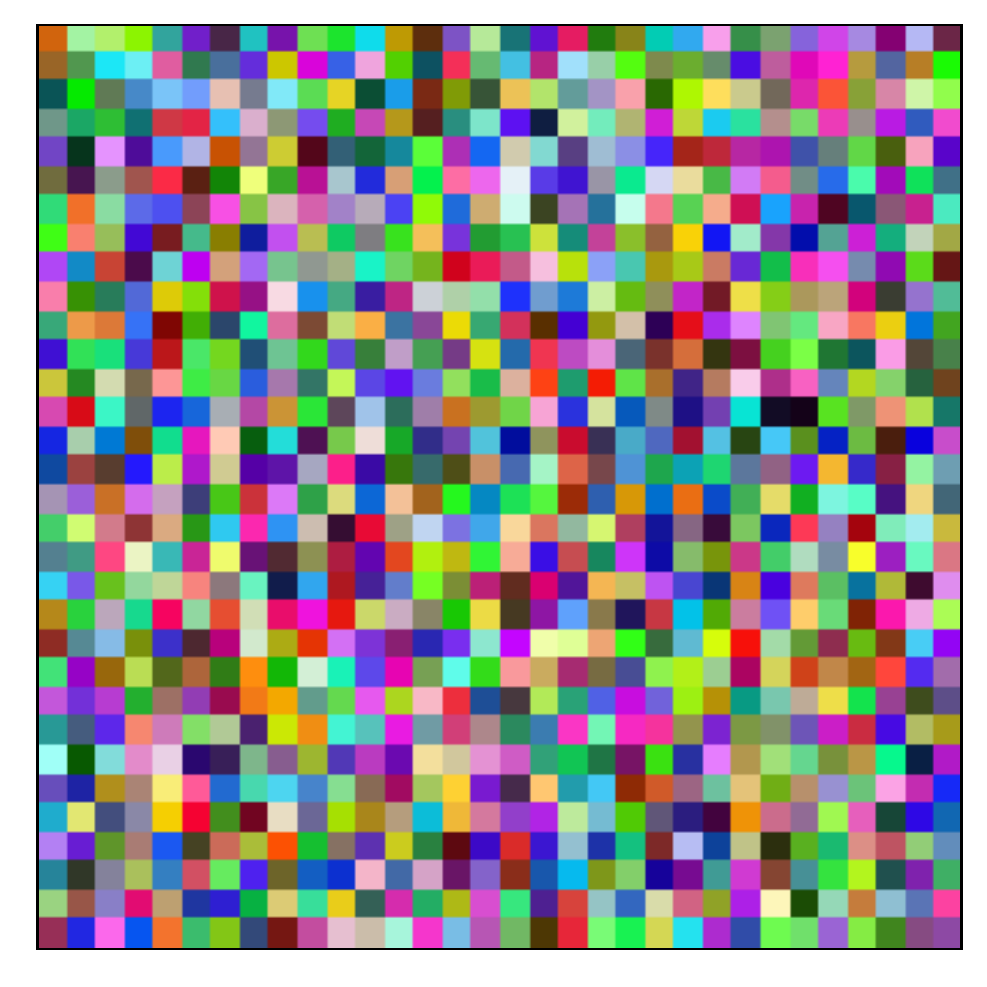}
    \caption{Noise}
    \label{fig:triggermethods:pervasive:noise}
\end{subfigure}
\hfill
\begin{subfigure}{0.19\linewidth}
    \includegraphics[width=\textwidth]{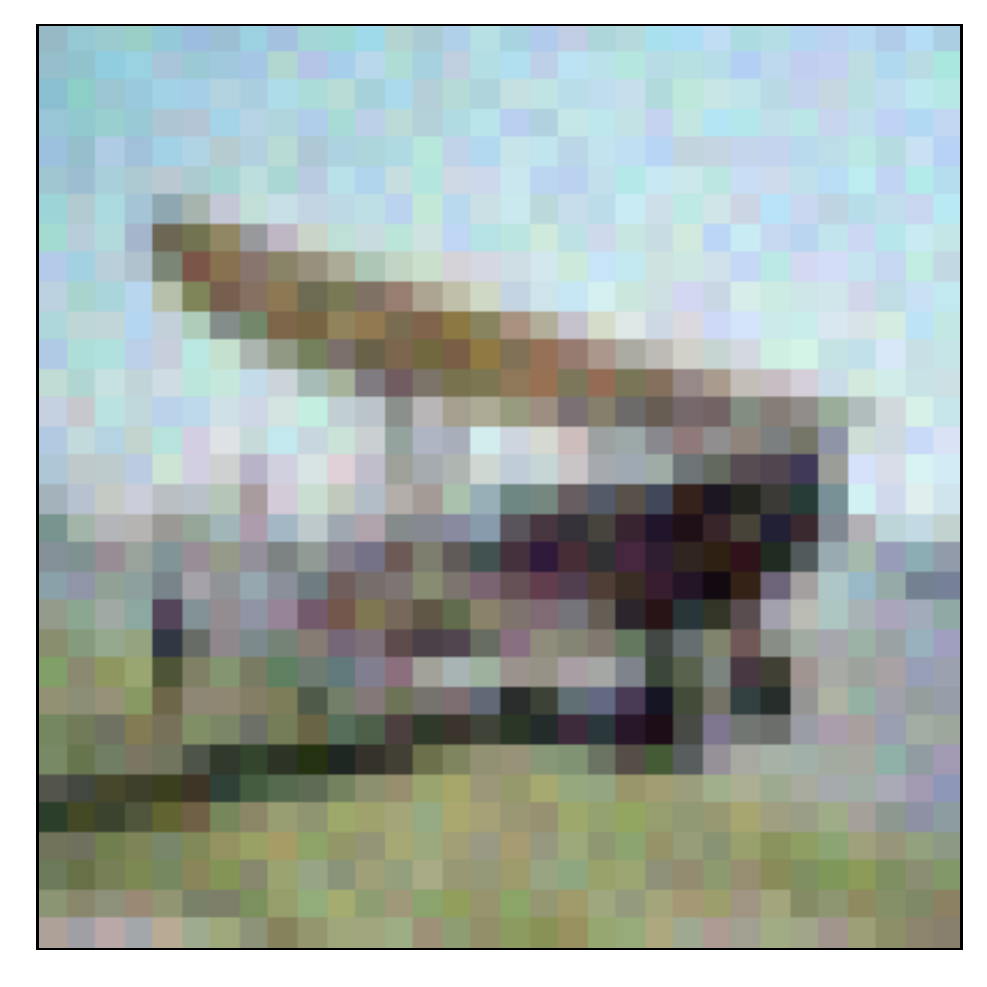}
    \caption{10\%}
    \label{fig:triggermethods:pervasive:ten}
\end{subfigure}
\hfill
\begin{subfigure}{0.19\linewidth}
    \includegraphics[width=\textwidth]{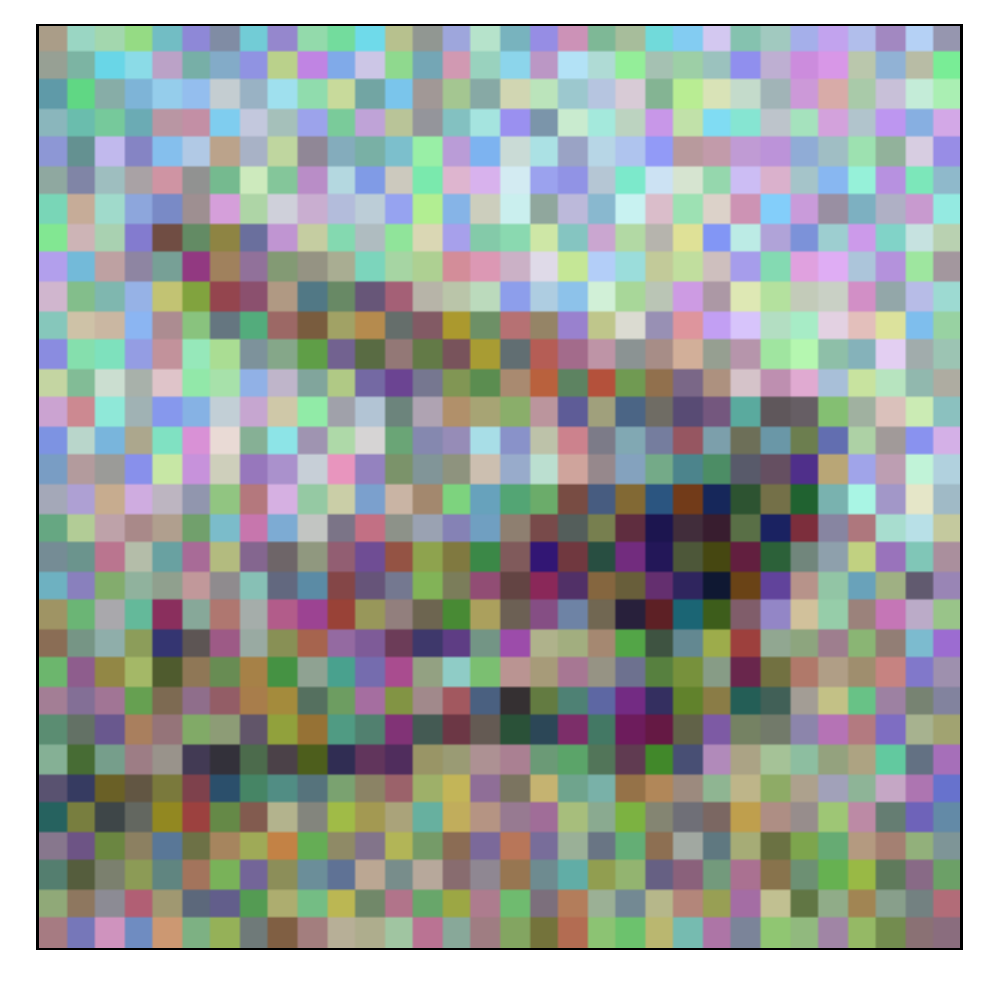}
    \caption{40\%}
    \label{fig:triggermethods:pervasive:fourty}
\end{subfigure}
\hfill
        
\caption{Visualization of samples for the Blend backdoor~\cite{chen2017targetedblend}. (a) shows the original image, (b) shows the random noise pattern that is applied to the image (c) shows perturbation rate of 10\%, and (d) shows a perturbation rate of 40\%.}
\label{fig:triggermethods:pervasive}
\end{figure}

Pervasive backdoors are hidden within the whole image and invisible to humans, e.g., added random noise. We leverage the Blend backdoor~\cite{chen2017targetedblend} in our experiments. Examples of a poisoned sample can be seen in \hyperref[fig:triggermethods:pervasive]{\fig\ref{fig:triggermethods:pervasive}}.

\subsection{Random Label Flipping}
\label{app:triggermethods:random}
For this untargeted attack, we flip the labels of each sample randomly, so that the model will be fed with falsely labeled data without any structure leading to additional model behavior. Therefore, this method leads to unlearning and thus reduces the \ma.

\subsection{Sign Flipping}
\label{app:triggermethods:sign}
This untargeted attack first trains a benign model. Afterward, the sign of every parameter is multiplied by minus one to create a destroyed model, hence leveraging model poisoning after training.

\subsection{Model Noising}
\label{app:triggermethods:noise}
Noising is also used as an \ir defense to erase backdoor behavior. However, in this poisoning attack, we noise the parameters of benign models to reduce the \ma.

\section{Quality of \pretrainedBig Models}
\label{app:pretrained}
\begin{figure}[tb]
  \centering
  \includegraphics[width=0.5\linewidth]{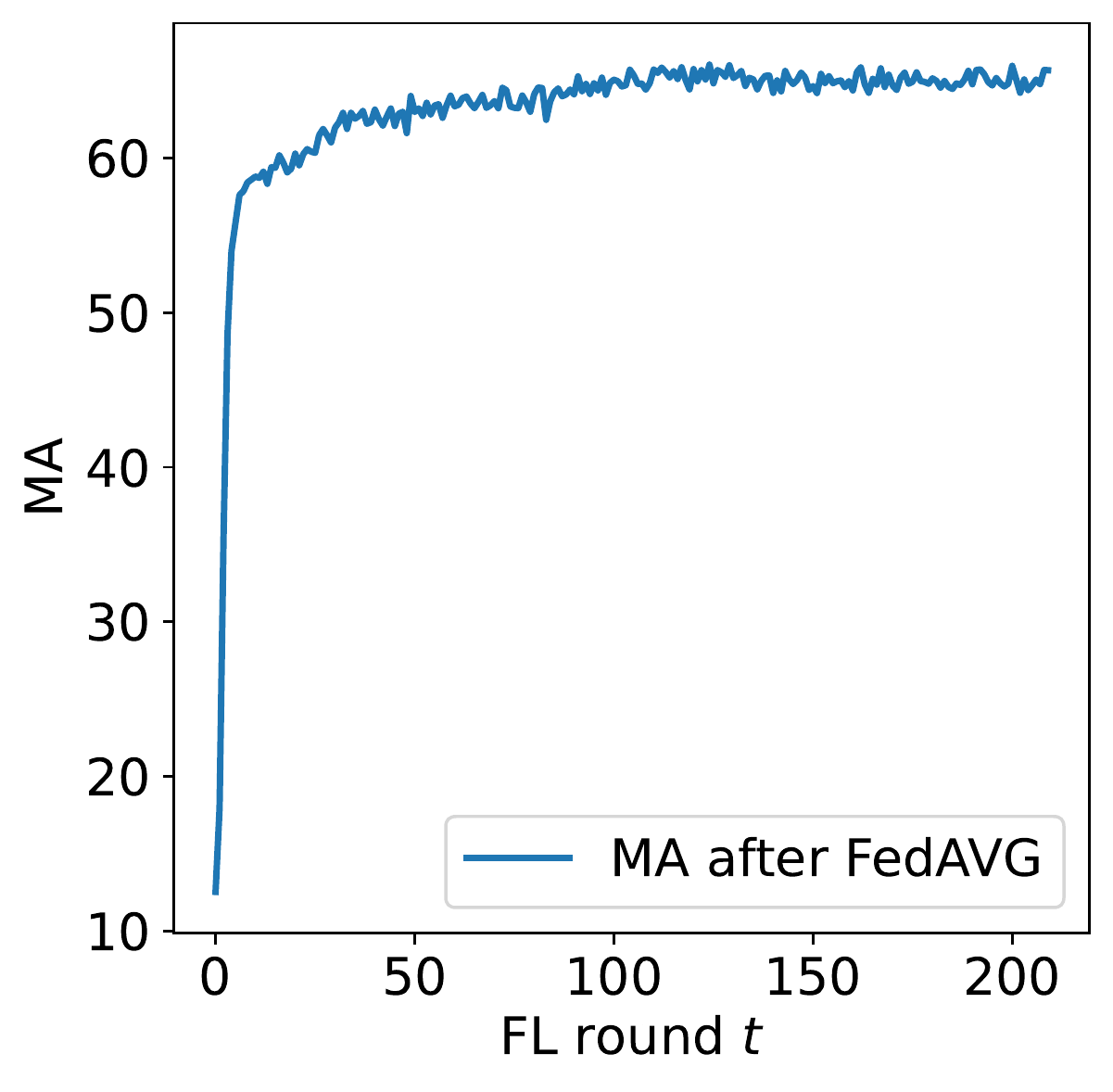}
  \caption{MA for all benign FL rounds in the default scenario with the following parameters: \cifar~\cite{cifar}, \iid distributed data, \clientCount = 20.}
\label{fig:acc:a}
\end{figure}
In our experiments in \hyperref[sec:eval]{\sect\ref{sec:eval}}, we use the parameters of several \pretrained models to initialize the global model. In \hyperref[fig:acc:a]{\fig\ref{fig:acc:a}}, we provide the accuracy in the main task for the model in the default scenario with the following parameters: 2560 samples per client, LR = 0.01 (SGD optimizer, momentum = 0.9, decay = 0.005, \clientCountSelected = 20), $seed_{rand} = 42$. After 50 rounds, the \ma is already high and stable, but increases untill round 125, before a clear overfitting of the model can be observed and \mbox{hyper-parameters} of the federation should be changed. Therefore, we select round 50 as the model in the default scenario.

\section{Additional details on \ourname}
\label{app:defense}
In this section, we provide additional information about \ourname, that helpful to understand the intuition and facilitates reproducibility of the defense.

\textit{Model Distances.} \hyperref[fig:motivation:update]{\fig\ref{fig:motivation:update}} depicts, how locally trained \fl models can vary withing the Euclidean or Cosine distance. We denote the locally trained models as \localModelRoundNext and the original global model, that served as a base for \localModelRoundNext is defined as \globalModelRound. Further we show that scaling of the model parameters after training effects the \metricCos, thus is not a stealthy model poisoning method for an adversary in \fl settings.

\textit{Metric Formulas.} We provide the formulas for the metrics of \ourname in \hyperref[eq:cos]{\equ\ref{eq:cos}} - \hyperref[eq:max]{\equ\ref{eq:max}}. $flatten$ denominates, that all model parameters are arranged in a one dimensional list and $nz$ is an abbreviation for the nonzero function. Each metric can be computed once per round $r$ for each client \clientIndexSelected as well as for each layer and the model as a whole.

\begin{equation} \label{eq:cos}
    \metricCos_\clientIndexSelected^\flroundNextNoSpace = 1 - cosine\_similarity(\flattenFunc(\localModelRoundNext - \globalModelRound))
\end{equation}
\begin{equation} \label{eq:eucl}
    \metricEucl_\clientIndexSelected^\flroundNextNoSpace = euclidean\_distance(\flattenFunc(\localModelRoundNext - \globalModelRound))
\end{equation}
\begin{equation} \label{eq:count}
    \metricCount_\clientIndexSelected^\flroundNextNoSpace = sum(\reluFunc(\signFunc(\flattenFunc(\localModelRoundNext - \globalModelRound))))
\end{equation}
\begin{equation} \label{eq:var}
    \metricVar_\clientIndexSelected^\flroundNextNoSpace = var(\flattenFunc(\localModelRoundNext))
\end{equation}
\begin{equation} \label{eq:min}
    \metricMin_\clientIndexSelected^\flroundNextNoSpace = min(\nonzeroFunc(abs(\localModelRoundNext - \globalModelRound)))
\end{equation}
\begin{equation} \label{eq:max}
    \metricMax_\clientIndexSelected^\flroundNextNoSpace = max(abs(\localModelRoundNext - \globalModelRound))
\end{equation}

\textit{Parameters.} The significance level for \ourname is set to 0.0001 for \iid, to 0.001 for \intraNonIid, and to 0.03 for \interNonIid scenarios.

\begin{table*}
\fontsize{7pt}{8pt}\selectfont
  \caption{\ma for different poisoning methods without adaptive adversary in percent.}
  \label{tab:acc:saa:trigger:ma}
  \begin{tabular}{l c|c|c|c|c|c|c|c|c|c}
    \toprule
    \multicolumn{2}{c|}{\multirow{3}{*}{Aggregation / Defenses}} &\multicolumn{9}{c}{\ma}\\
    & & Pixel Trigger & Clean-Label & Semantic & Edge Case & Label Flip &  Pervasive  & Random Flip & Sign Flip & Noising\\
    & & \cite{badnets} & \cite{turner2019label} & \cite{bagdasaryan2020backdoorfl} & \cite{wang2020attackontails} & \cite{biggio2013poisoninglabelflip, cao2019labelflipdistributed} & \cite{chen2017targetedblend}  & \hyperref[app:triggermethods:random]{\sect\ref{app:triggermethods:random}} & \hyperref[app:triggermethods:sign]{\sect\ref{app:triggermethods:sign}} & \hyperref[app:triggermethods:noise]{\sect\ref{app:triggermethods:noise}}\\
    \midrule
    1:& Global model \globalModelRound& 62.99 & 62.99 & 62.99 & 62.99 & 62.99 & 62.99 & 62.99 & 62.99& 62.99 \\
    \hline
    2:& Average of benign local models & 57.58 & 57.58 & 57.58 & 57.58 & 57.58 & 57.58 & 57.58 & 57.58& 57.58 \\
    3:& Average of poisoned local models & 57.84 & 54.49 & 54.37 & 58.69 & 47.87 & 53.69 & 53.69 & 10.00 & 46.65\\
    \hline
    4:& \fedavg with benign local models & 63.57 & 63.57 & 63.57 & 63.57 & 63.57 & 63.57 & 63.57 & 63.57& 63.57 \\
    5:& \fedavg with poisoned local models & 64.92 & 61.79 & 65.49 & 66.55 & 58.31 & 63.66 & 52.55 & 10.00 & 62.73 \\
    \hline
    6:& \fedavg with all local models & 63.81 & 64.20 & 64.66 & 64.52 & 57.09 & 63.51 & 57.03 & 10.00 &  63.07\\
    \hline
    7:& \naiveBig \clustering & 65.06 & 63.57 & 65.02 & 65.65 & 57.63 & 63.83 & 53.99 & 63.57 &  63.48\\
    8:& \foolsgold~\cite{fung2020FoolsGold} & 63.57 & 63.57 & 63.59 & 63.57 & 63.57 & 63.66 & 60.41 & 63.57 &  63.07\\
    9:& \krum~\cite{blanchard17Krum} & 59.75 & 55.18 & 58.72 & 59.86 & 58.38 & 58.38 & 58.38 & 58.38 &  58.38\\
    10:& \multikrum~\cite{blanchard17Krum} & 64.18 & 61.65 & 65.94 & 66.14 & 64.15 & 65.26 & 64.15 & 64.15 &  64.15\\
    11:& Clip~\cite{mcmahan2018iclrClippingLanguage} & 63.80 & 64.21 & 64.52 & 64.48 & 56.99 & 63.39 & 54.01 & 10.00 &  63.58\\
    12:& Clip\&Noise~\cite{mcmahan2018iclrClippingLanguage} & 50.78 & 59.94 & 57.60 & 57.85 & 50.04 & 54.86 & 49.95 & 10.00 &  57.81\\
    13:& \flame~\cite{nguyen22Flame} & 60.96 & 60.03 & 62.13 & 64.27 & 57.11 & 59.15 & 60.99 & 60.99 &  62.60\\
    14:& T-Mean~\cite{yin2018trimmedMeanMedian} & 63.51 & 64.08 & 64.17 & 64.20 & 56.96 & 63.04 & 56.77 & 10.00 &  63.27\\
    15:& T-Median~\cite{yin2018trimmedMeanMedian} & 51.22 & 53.64 & 52.11 & 55.13 & 48.36 & 49.40 &44.69 & 10.00 &  51.53\\
    16:& FLTrust~\cite{cao2020fltrust} & 63.49 & 63.37 & 63.75 & 64.05 & 62.16 & 63.25 & 63.16 & 26.97 & 63.32\\
    17:& \textbf{\ourname} & 63.57 & 62.18 & 63.36 & 63.15 & 63.15 & 62.82 & 62.88 & 63.57 &  63.57\\
  \bottomrule
\end{tabular}
\end{table*}

\section{Detailed Experimental Results}
\label{app:additionalresults}
In this section, we provide further results for our experiments, that could not be included in the main part of the paper due to space limitations. We selected the most interesting and representative results to be part of the main part, but report the rest below for completeness.

\hyperref[tab:acc:saa:trigger:ma]{\tab\ref{tab:acc:saa:trigger:ma}} reports the \mas corresponding to the \bas and \accs in \hyperref[tab:acc:saa:trigger:ba]{\tab\ref{tab:acc:saa:trigger:ba}} for different poisoning methods without adaptive adversaries. Note, that \ourname provides high \ma independent of the applied poisoning attack. The table shows, that \ourname does not negatively impact the \ma and is also effective against untargeted attacks.

\hyperref[tab:acc:saa:plain:scaled]{\tab\ref{tab:acc:saa:plain:scaled}} shows the results for our default scenario with scaled poisoned models regarding the Euclidean distance of updates. The results confirm, that scaling can increase the \ba. \ourname is already efficient for the unscaled version visualized in scenario \circled{1} in \hyperref[tab:test]{\tab\ref{tab:test}}, hence is also effective for scaled models due to the intuition visualized in \hyperref[fig:motivation:update]{\fig\ref{fig:motivation:update}}. \hyperref[tab:acc:saa:plain:pdr03:scaled]{\tab\ref{tab:acc:saa:plain:pdr03:scaled}} depicts the results for the default scenario with a \pdr of 0.3 for scaled poisoned models.

\begin{table}
\fontsize{7pt}{8pt}\selectfont
  \caption{\ma and \ba in the default scenario with scaled poisoned models regarding the Euclidean distance of updates in percent.}
  \label{tab:acc:saa:plain:scaled}
  \begin{tabular}{l c|c|c}
    \toprule
    \multicolumn{2}{c|}{Accuracies without defenses} &\ma&\ba\\
    \midrule
    1:& Global model \globalModelRound& 62.99 & 1.90\\
    \hline
    2:& Average of benign local models & 57.58 & 4.56\\
    3:& Average of poisoned local models & 57.84 & 85.13\\
    \hline
    4:& \fedavg with benign local models & 63.57 & 1.85\\
    5:& \fedavg with poisoned local models & 64.49 & 86.45\\
    \hline
    6:& \fedavg with all local models & 63.61& \textbf{51.15}\\
    \bottomrule
    \toprule
    \multicolumn{2}{c|}{Global model accuracies after applying defenses} &\ma&\ba\\
    \midrule
    7:& \naiveBig \clustering & 63.67 & \textbf{60.85}\\
    8:& \foolsgold~\cite{fung2020FoolsGold} & 63.57 & \textbf{1.85}\\
    9:& \krum~\cite{blanchard17Krum} & 58.38 & \textbf{3.98}\\
    10:& \multikrum~\cite{blanchard17Krum} & 64.24 & \textbf{56.23}\\
    11:& Clip~\cite{mcmahan2018iclrClippingLanguage} & 63.61 & \textbf{60.33}\\
    12:& Clip\&Noise~\cite{mcmahan2018iclrClippingLanguage} & 57.63&\textbf{60.66}\\
    13:& \flame~\cite{nguyen22Flame} & 60.40 & \textbf{71.02}\\
    14:& T-Mean~\cite{yin2018trimmedMeanMedian} & 63.35 & \textbf{51.52}\\
    15:& T-Median~\cite{yin2018trimmedMeanMedian} & 49.89&\textbf{44.34}\\
    16:& FLTrust~\cite{cao2020fltrust} & 63.13 & \textbf{23.24}\\
    17:& \textbf{\ourname} & 63.36 & \textbf{1.95}\\
  \bottomrule
\end{tabular}
\end{table}

\begin{table}
\fontsize{7pt}{8pt}\selectfont
  \caption{\ma and \ba in the default scenario with \pdr of 0.3 and scaled poisoned models regarding the Euclidean distance of updates in percent.}
  \label{tab:acc:saa:plain:pdr03:scaled}
  \begin{tabular}{l c|c|c}
    \toprule
    \multicolumn{2}{c|}{Accuracies without defenses} &\ma&\ba\\
    \midrule
    1:& Global model \globalModelRound& 62.99 & 1.90\\
    \hline
    2:& Average of benign local models & 57.58 & 4.56\\
    3:& Average of poisoned local models & 54.58 & 93.15\\
    \hline
    4:& \fedavg with benign local models & 63.57 & 1.85\\
    5:& \fedavg with poisoned local models & 58.49 & 97.46\\
    \hline
    6:& \fedavg with all local models & 62.95& \textbf{75.81}\\
    \bottomrule
    \toprule
    \multicolumn{2}{c|}{Global model accuracies after applying defenses} &\ma&\ba\\
    \midrule
    7:& \naiveBig \clustering & 64.57 & \textbf{86.86}\\
    8:& \foolsgold~\cite{fung2020FoolsGold} & 63.57 & \textbf{1.85}\\
    9:& \krum~\cite{blanchard17Krum} & 52.22 & \textbf{95.97}\\
    10:& \multikrum~\cite{blanchard17Krum} & 63.90 & \textbf{92.72}\\
    11:& Clip~\cite{mcmahan2018iclrClippingLanguage} & 63.85 & \textbf{61.86}\\
    12:& Clip\&Noise~\cite{mcmahan2018iclrClippingLanguage} & 57.81&\textbf{70.87}\\
    13:& \flame~\cite{nguyen22Flame} & 60.08 & \textbf{91.34}\\
    14:& T-Mean~\cite{yin2018trimmedMeanMedian} & 63.54 & \textbf{63.98}\\
    15:& T-Median~\cite{yin2018trimmedMeanMedian} & 51.18&\textbf{57.37}\\
    16:& FLTrust~\cite{cao2020fltrust} & 63.13 & \textbf{23.25}\\
    17:& \textbf{\ourname} & 63.36 & \textbf{1.95}\\
  \bottomrule
\end{tabular}
\end{table}

We conducted an attack leveraging our strong adaptive adversary against \foolsgold~\cite{fung2020FoolsGold} and report the result in \hyperref[tab:acc:saa:fixationlast]{\tab\ref{tab:acc:saa:fixationlast}}. As strategy, we first trained a benign local model and transferred the trained parameters of the last layer to a fresh local model. We then excluded the parameters form training and poisoned the local model forcing the backdoor into some other layers. In \hyperref[tab:acc:saa:fixationlast:scaled]{\tab\ref{tab:acc:saa:fixationlast:scaled}} we additionally applied scaling. We can observe, that the \ba of \foolsgold can be increased with those methods.

\begin{table}
\fontsize{7pt}{8pt}\selectfont
  \caption{\ma and \ba in the default scenario with fixation of the last layer to benign trained parameters in percent.}
  \label{tab:acc:saa:fixationlast}
  \begin{tabular}{l c|c|c}
    \toprule
    \multicolumn{2}{c|}{Accuracies without defenses} &\ma&\ba\\
    \midrule
    1:& Global model \globalModelRound& 62.99 & 1.90\\
    \hline
    2:& Average of benign local models & 57.58 & 4.56\\
    3:& Average of poisoned local models & 58.20 & 84.90\\
    \hline
    4:& \fedavg with benign local models & 63.57 & 1.85\\
    5:& \fedavg with poisoned local models & 64.29 & 83.96\\
    \hline
    6:& \fedavg with all local models & 63.74& \textbf{42.22}\\
    \bottomrule
    \toprule
    \multicolumn{2}{c|}{Global model accuracies after applying defenses} &\ma&\ba\\
    \midrule
    7:& \naiveBig \clustering & 63.68 & \textbf{45.95}\\
    8:& \foolsgold~\cite{fung2020FoolsGold} & 63.74 & \textbf{42.22}\\
    9:& \krum~\cite{blanchard17Krum} & 59.69 & \textbf{83.21}\\
    10:& \multikrum~\cite{blanchard17Krum} & 63.90 & \textbf{92.72}\\
    11:& Clip~\cite{mcmahan2018iclrClippingLanguage} & 63.81 & \textbf{42.23}\\
    12:& Clip\&Noise~\cite{mcmahan2018iclrClippingLanguage} & 52.58&\textbf{62.80}\\
    13:& \flame~\cite{nguyen22Flame} & 60.80 & \textbf{76.58}\\
    14:& T-Mean~\cite{yin2018trimmedMeanMedian} & 63.43 & \textbf{43.50}\\
    15:& T-Median~\cite{yin2018trimmedMeanMedian} & 51.94&\textbf{36.75}\\
    16:& FLTrust~\cite{cao2020fltrust} & 63.66 & \textbf{20.14}\\
    17:& \textbf{\ourname} & 63.57 & \textbf{1.85}\\
  \bottomrule
\end{tabular}
\end{table}

\begin{table}
\fontsize{7pt}{8pt}\selectfont
  \caption{\ma and \ba in the default scenario with fixation of the last layer to benign trained parameters and scaled poisoned models regarding the Euclidean distance of updates in percent.}
  \label{tab:acc:saa:fixationlast:scaled}
  \begin{tabular}{l c|c|c}
    \toprule
    \multicolumn{2}{c|}{Accuracies without defenses} &\ma&\ba\\
    \midrule
    1:& Global model \globalModelRound& 62.99 & 1.90\\
    \hline
    2:& Average of benign local models & 57.58 & 4.56\\
    3:& Average of poisoned local models & 58.20 & 84.90\\
    \hline
    4:& \fedavg with benign local models & 63.57 & 1.85\\
    5:& \fedavg with poisoned local models & 63.96 & 87.35\\
    \hline
    6:& \fedavg with all local models & 63.66& \textbf{50.44}\\
    \bottomrule
    \toprule
    \multicolumn{2}{c|}{Global model accuracies after applying defenses} &\ma&\ba\\
    \midrule
    7:& \naiveBig \clustering & 63.29 & \textbf{56.94}\\
    8:& \foolsgold~\cite{fung2020FoolsGold} & 63.61 & \textbf{50.44}\\
    9:& \krum~\cite{blanchard17Krum} & \textbf{58.38} & 3.98\\
    10:& \multikrum~\cite{blanchard17Krum} & 64.26 & \textbf{53.96}\\
    11:& Clip~\cite{mcmahan2018iclrClippingLanguage} & 53.30 & \textbf{58.45}\\
    12:& Clip\&Noise~\cite{mcmahan2018iclrClippingLanguage} & 57.63&\textbf{60.66}\\
    13:& \flame~\cite{nguyen22Flame} & 62.67 & \textbf{71.26}\\
    14:& T-Mean~\cite{yin2018trimmedMeanMedian} & 63.27 & \textbf{50.56}\\
    15:& T-Median~\cite{yin2018trimmedMeanMedian} & 51.76&\textbf{39.64}\\
    16:& FLTrust~\cite{cao2020fltrust} & 63.45 & \textbf{20.37}\\
    17:& \textbf{\ourname} & 63.36 & \textbf{1.95}\\
  \bottomrule
\end{tabular}
\end{table}

\begin{figure}[tb]
  \centering
  \includegraphics[width=0.5\linewidth]{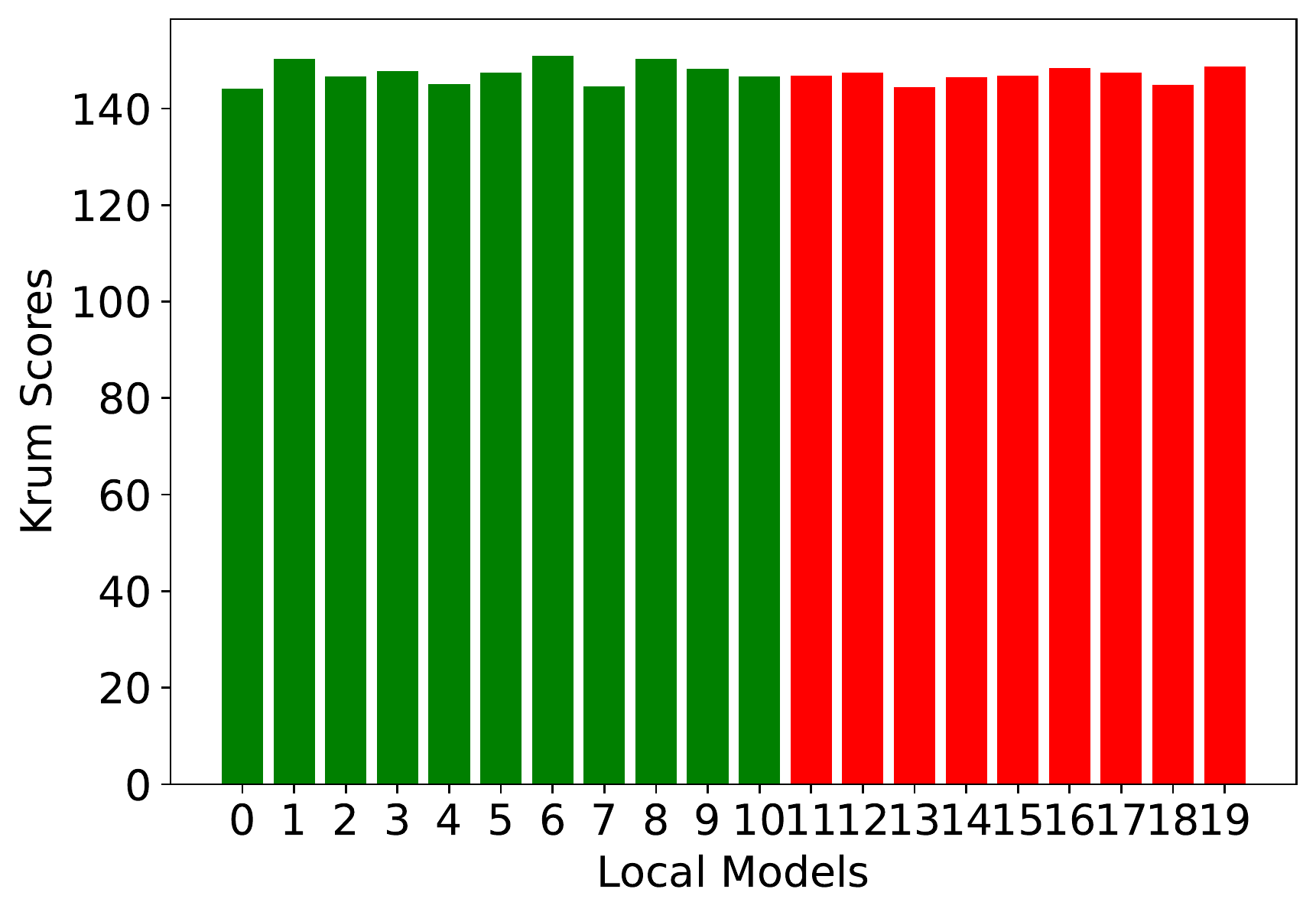}
  \caption{\krum scores for the default scenario. The erasure of the backdoor in the default scenario is based on the fact, that a benign score is the most central one. Nevertheless, the metric is not highlighting the malicious models significantly.}
\label{fig:krum}
\end{figure}

Scenario \circled{3} of \hyperref[tab:test]{\tab\ref{tab:test}} and \hyperref[tab:acc:saa:euclclassic:scaled]{\tab\ref{tab:acc:saa:euclclassic:scaled}} shows the result after adapting to \krum scores~\cite{blanchard17Krum} in an unscaled and scaled version. We forced the Euclidean distance between poisoned models and the global model to be similar to each other and on a benign level so that the defenses decide in favor of the poisoned models. \hyperref[fig:krum]{\fig\ref{fig:krum}} depicts the \krum scores for the default scenario associated with the default scenario in \hyperref[tab:test]{\tab\ref{tab:test}} to show, that the effective backdoor is based on coincidence and not intentionally forced by the attacker. With slight changes in some models, \krum could also decide for a benign model. However, after adaption, we can introduce a high \ba intentionally in scenario \circled{3} of \hyperref[tab:test]{\tab\ref{tab:test}} and \hyperref[tab:acc:saa:euclclassic:scaled]{\tab\ref{tab:acc:saa:euclclassic:scaled}}.

\begin{table}
\fontsize{7pt}{8pt}\selectfont
  \caption{\ma and \ba in the default scenario with adaption of the Euclidean distance between local models and the global model to benign values and scaled poisoned models regarding the Euclidean distance of updates in percent.}
  \label{tab:acc:saa:euclclassic:scaled}
  \begin{tabular}{l c|c|c}
    \toprule
    \multicolumn{2}{c|}{Accuracies without defenses} &\ma&\ba\\
    \midrule
    1:& Global model \globalModelRound& 62.99 & 1.90\\
    \hline
    2:& Average of benign local models & 57.58 & 4.56\\
    3:& Average of poisoned local models & 51.23 & 89.82\\
    \hline
    4:& \fedavg with benign local models & 63.57 & 1.85\\
    5:& \fedavg with poisoned local models & 40.23 & 93.54\\
    \hline
    6:& \fedavg with all local models & 48.90& \textbf{83.93}\\
    \bottomrule
    \toprule
    \multicolumn{2}{c|}{Global model accuracies after applying defenses} &\ma&\ba\\
    \midrule
    7:& \naiveBig \clustering & 46.98 & \textbf{85.76}\\
    8:& \foolsgold~\cite{fung2020FoolsGold} & 63.57 & \textbf{1.85}\\
    9:& \krum~\cite{blanchard17Krum} & 51.46 & \textbf{87.88}\\
    10:& \multikrum~\cite{blanchard17Krum} & 43.19 & \textbf{95.25}\\
    11:& Clip~\cite{mcmahan2018iclrClippingLanguage} & 49.08 & \textbf{83.83}\\
    12:& Clip\&Noise~\cite{mcmahan2018iclrClippingLanguage} & 43.95&\textbf{87.44}\\
    13:& \flame~\cite{nguyen22Flame} & 46.11 & \textbf{92.83}\\
    14:& T-Mean~\cite{yin2018trimmedMeanMedian} & 50.85 & \textbf{85.88}\\
    15:& T-Median~\cite{yin2018trimmedMeanMedian} & 39.62&\textbf{74.82}\\
    16:& FLTrust~\cite{cao2020fltrust} & 55.07 & \textbf{74.72}\\
    17:& \textbf{\ourname} & 63.57 & \textbf{1.85}\\
  \bottomrule
\end{tabular}
\end{table}

\hyperref[tab:acc:noniid:oneclass05]{\tab\ref{tab:acc:noniid:oneclass05}} and \hyperref[tab:acc:noniid:oneclass05:scaled]{\tab\ref{tab:acc:noniid:oneclass05:scaled}} show the results for a classical \intraNonIid scenario crafted by \nonIidOneClass with $\noniidsign = 0.5$. In both cases, \ourname reduces the \ba reliably with only one \fp, while other defenses allow the attacker to embed 34.72\% to 49.47\% \ba. \hyperref[tab:acc:noniid:oneclass05:pdr03]{\tab\ref{tab:acc:noniid:oneclass05:pdr03}} depicts the results for an increased \pdr of 0.3, allowing the adversary to reach a \ba of 57.68\% without defense and up to 92.86\% under active defenses. \ourname still removes the poisoned models most effectively with only two \fps.

\begin{table}
\fontsize{7pt}{8pt}\selectfont
  \caption{\ma and \ba in the default scenario for \nonIidOneClass with $\noniidsign = 0.5$ in percent.}
  \label{tab:acc:noniid:oneclass05}
  \begin{tabular}{l c|c|c}
    \toprule
    \multicolumn{2}{c|}{Accuracies without defenses} &\ma&\ba\\
    \midrule
    1:& Global model \globalModelRound& 62.99 & 1.93\\
    \hline
    2:& Average of benign local models & 47.15 & 6.82\\
    3:& Average of poisoned local models & 45.88 & 84.42\\
    \hline
    4:& \fedavg with benign local models & 65.92 & 1.40\\
    5:& \fedavg with poisoned local models & 64.07 & 83.42\\
    \hline
    6:& \fedavg with all local models & 65.48& 43.96\\
    \bottomrule
    \toprule
    \multicolumn{2}{c|}{Global model accuracies after applying defenses} &\ma&\ba\\
    \midrule
    7:& \naiveBig \clustering & 64.76 & 47.02\\
    8:& \foolsgold~\cite{fung2020FoolsGold} & 50.13 & \textbf{3.45}\\
    9:& \krum~\cite{blanchard17Krum} & \textbf{49.88} & 5.27\\
    10:& \multikrum~\cite{blanchard17Krum} & 60.98 & 52.57\\
    11:& Clip~\cite{mcmahan2018iclrClippingLanguage} & 65.50 & 41.33\\
    12:& Clip\&Noise~\cite{mcmahan2018iclrClippingLanguage} & 59.53&52.31\\
    13:& \flame~\cite{nguyen22Flame} & 61.46 & 34.37\\
    14:& T-Mean~\cite{yin2018trimmedMeanMedian} & 64.83 & 47.27\\
    15:& T-Median~\cite{yin2018trimmedMeanMedian} & 54.84 & 47.46\\
    16:& FLTrust~\cite{cao2020fltrust} & 69.96 & \textbf{4.64}\\
    17:& \textbf{\ourname} & 64.70 & \textbf{2.13}\\
  \bottomrule
\end{tabular}
\end{table}

\begin{table}
\fontsize{7pt}{8pt}\selectfont
  \caption{\ma and \ba in the default scenario for \nonIidOneClass with $\noniidsign = 0.5$ and scaled
poisoned models regarding the Euclidean distance of updates in percent.}
  \label{tab:acc:noniid:oneclass05:scaled}
  \begin{tabular}{l c|c|c}
    \toprule
    \multicolumn{2}{c|}{Accuracies without defenses} &\ma&\ba\\
    \midrule
    1:& Global model \globalModelRound& 62.99 & 1.93\\
    \hline
    2:& Average of benign local models & 47.15 & 6.82\\
    3:& Average of poisoned local models & 45.88 & 84.42\\
    \hline
    4:& \fedavg with benign local models & 65.92 & 1.40\\
    5:& \fedavg with poisoned local models & 64.46 & 84.48\\
    \hline
    6:& \fedavg with all local models & 65.31& 45.94\\
    \bottomrule
    \toprule
    \multicolumn{2}{c|}{Global model accuracies after applying defenses} &\ma&\ba\\
    \midrule
    7:& \naiveBig \clustering & 65.59 & 49.47\\
    8:& \foolsgold~\cite{fung2020FoolsGold} & 50.13 & \textbf{3.45}\\
    9:& \krum~\cite{blanchard17Krum} & \textbf{49.88} & 5.27\\
    10:& \multikrum~\cite{blanchard17Krum} & 63.59 & 7.21\\
    11:& Clip~\cite{mcmahan2018iclrClippingLanguage} & 65.39 & 43.36\\
    12:& Clip\&Noise~\cite{mcmahan2018iclrClippingLanguage} & 58.89&52.54\\
    13:& \flame~\cite{nguyen22Flame} & 58.86 & 34.72\\
    14:& T-Mean~\cite{yin2018trimmedMeanMedian} & 64.61 & 48.96\\
    15:& T-Median~\cite{yin2018trimmedMeanMedian} & 54.51 & 48.37\\
    16:& FLTrust~\cite{cao2020fltrust} & 63.92 & \textbf{4.63}\\
    17:& \textbf{\ourname} & 64.69 & \textbf{2.24}\\
  \bottomrule
\end{tabular}
\end{table}

\begin{table}
\fontsize{7pt}{8pt}\selectfont
  \caption{\ma and \ba in the default scenario with a \pdr of 0.3 and for \nonIidOneClass with $\noniidsign = 0.5$ in percent.}
  \label{tab:acc:noniid:oneclass05:pdr03}
  \begin{tabular}{l c|c|c}
    \toprule
    \multicolumn{2}{c|}{Accuracies without defenses} &\ma&\ba\\
    \midrule
    1:& Global model \globalModelRound& 62.99 & 1.93\\
    \hline
    2:& Average of benign local models & 47.15 & 6.82\\
    3:& Average of poisoned local models & 43.74 & 91.32\\
    \hline
    4:& \fedavg with benign local models & 65.92 & 1.40\\
    5:& \fedavg with poisoned local models & 61.48 & 92.92\\
    \hline
    6:& \fedavg with all local models & 65.64& 57.68\\
    \bottomrule
    \toprule
    \multicolumn{2}{c|}{Global model accuracies after applying defenses} &\ma&\ba\\
    \midrule
    7:& \naiveBig \clustering & 62.49 & 83.70\\
    8:& \foolsgold~\cite{fung2020FoolsGold} & 57.60 & 39.05\\
    9:& \krum~\cite{blanchard17Krum} & 49.43 & 92.86\\
    10:& \multikrum~\cite{blanchard17Krum} & 57.51 & 85.95\\
    11:& Clip~\cite{mcmahan2018iclrClippingLanguage} & 64.60 & 56.81\\
    12:& Clip\&Noise~\cite{mcmahan2018iclrClippingLanguage} & 57.77&70.20\\
    13:& \flame~\cite{nguyen22Flame} & 60.55 & 45.36\\
    14:& T-Mean~\cite{yin2018trimmedMeanMedian} & 63.81 & 62.77\\
    15:& T-Median~\cite{yin2018trimmedMeanMedian} & 52.96 & 66.78\\
    16:& FLTrust~\cite{cao2020fltrust} & 63.73 & \textbf{8.31}\\
    17:& \textbf{\ourname} & 65.49 & \textbf{1.46}\\
  \bottomrule
\end{tabular}
\end{table}

In \hyperref[tab:acc:noniid:oneclass05:pdr03:cos]{\tab\ref{tab:acc:noniid:oneclass05:pdr03:cos}} and \hyperref[tab:acc:noniid:oneclass05:pdr03:cos]{\tab\ref{tab:acc:noniid:oneclass05:pdr03:cos}} we adapt all malicious models to a central benign value regarding the Cosine distance to the global model, which has the effect, that the models are inconspicuous in \krum scores, hence can circumvent \krum and \multikrum~\cite{blanchard17Krum}, while \ourname still erases the backdoor with only two \fps.

\begin{table}
\fontsize{7pt}{8pt}\selectfont
  \caption{\ma and \ba in the default scenario for \nonIidOneClass with $\noniidsign = 0.5$, adaption to benign values regarding the Cosine distance to the global model in percent.}
  \label{tab:acc:noniid:oneclass05:pdr03:cos}
  \begin{tabular}{l c|c|c}
    \toprule
    \multicolumn{2}{c|}{Accuracies without defenses} &\ma&\ba\\
    \midrule
    1:& Global model \globalModelRound& 62.99 & 1.93\\
    \hline
    2:& Average of benign local models & 47.15 & 6.82\\
    3:& Average of poisoned local models & 61.27 & 78.68\\
    \hline
    4:& \fedavg with benign local models & 65.92 & 1.40\\
    5:& \fedavg with poisoned local models & 70.12 & 78.05\\
    \hline
    6:& \fedavg with all local models & 66.70& 23.35\\
    \bottomrule
    \toprule
    \multicolumn{2}{c|}{Global model accuracies after applying defenses} &\ma&\ba\\
    \midrule
    7:& \naiveBig \clustering & 65.92 & 1.40\\
    8:& \foolsgold~\cite{fung2020FoolsGold} & 65.79 & 1.58\\
    9:& \krum~\cite{blanchard17Krum} & 63.26 & 69.94\\
    10:& \multikrum~\cite{blanchard17Krum} & 68.45 & 79.21\\
    11:& Clip~\cite{mcmahan2018iclrClippingLanguage} & 66.79 & 24.16\\
    12:& Clip\&Noise~\cite{mcmahan2018iclrClippingLanguage} & 56.22 & 25.71\\
    13:& \flame~\cite{nguyen22Flame} & 64.12 & 27.76\\
    14:& T-Mean~\cite{yin2018trimmedMeanMedian} & 65.96 & 27.62\\
    15:& T-Median~\cite{yin2018trimmedMeanMedian} & 51.60 & 40.07\\
    16:& FLTrust~\cite{cao2020fltrust} & 63.57 & \textbf{18.75}\\
    17:& \textbf{\ourname} & 66.47 & \textbf{1.45}\\
  \bottomrule
\end{tabular}
\end{table}

\begin{table}
\fontsize{7pt}{8pt}\selectfont
  \caption{\ma and \ba in the default scenario for \nonIidOneClass with $\noniidsign = 0.5$, adaption to benign values regarding the Cosine distance to the global model and scaled poisoned models regarding the Euclidean distance of updates in percent.}
  \label{tab:acc:noniid:oneclass05:pdr03:cos:scaled}
  \begin{tabular}{l c|c|c}
    \toprule
    \multicolumn{2}{c|}{Accuracies without defenses} &\ma&\ba\\
    \midrule
    1:& Global model \globalModelRound& 62.99 & 1.93\\
    \hline
    2:& Average of benign local models & 47.15 & 6.82\\
    3:& Average of poisoned local models & 61.27 & 78.68\\
    \hline
    4:& \fedavg with benign local models & 65.92 & 1.40\\
    5:& \fedavg with poisoned local models & 42.53 & 95.03\\
    \hline
    6:& \fedavg with all local models & 60.38& 55.28\\
    \bottomrule
    \toprule
    \multicolumn{2}{c|}{Global model accuracies after applying defenses} &\ma&\ba\\
    \midrule
    7:& \naiveBig \clustering & 65.92 & 1.44\\
    8:& \foolsgold~\cite{fung2020FoolsGold} & 64.86 & 1.75\\
    9:& \krum~\cite{blanchard17Krum} & 23.13 & 84.58\\
    10:& \multikrum~\cite{blanchard17Krum} & 35.39 & 90.44\\
    11:& Clip~\cite{mcmahan2018iclrClippingLanguage} & 60.84 & 53.96\\
    12:& Clip\&Noise~\cite{mcmahan2018iclrClippingLanguage} & 40.94 & 80.35\\
    13:& \flame~\cite{nguyen22Flame} & 62.33 & 4.07\\
    14:& T-Mean~\cite{yin2018trimmedMeanMedian} & 58.91 & 57.82\\
    15:& T-Median~\cite{yin2018trimmedMeanMedian} & 33.96 & 50.51\\
    16:& FLTrust~\cite{cao2020fltrust} & 59.83 & \textbf{21.46}\\
    17:& \textbf{\ourname} & 66.47 & \textbf{1.45}\\
  \bottomrule
\end{tabular}
\end{table}

\hyperref[tab:acc:noniid:random:r1]{\tab\ref{tab:acc:noniid:random:r1}} and \hyperref[tab:acc:noniid:random:r1:scaled]{\tab\ref{tab:acc:noniid:random:r1:scaled}} show the results in a \interNonIid scenario based on our \randomgen strategy for a model in \fl round one and highlights, that \ourname outperforms other defenses in reducing the \ba of the new global model. 
\hyperref[tab:acc:noniid:100]{\tab\ref{tab:acc:noniid:100}} and \hyperref[tab:acc:noniid:100:scaled]{\tab\ref{tab:acc:noniid:100:scaled}} show the results for a setting in round 50,w here 100 clients are part of the federation and 20 clients are selected randomly in each round for training. Due to the later rounds, \ourname is even more effective than other defenses and reduces the \ba to a minimum.

\begin{table}
\fontsize{7pt}{8pt}\selectfont
  \caption{\ma and \ba in the default scenario with \interNonIid based on our \randomgen strategy with a model in \fl round one in percent.}
  \label{tab:acc:noniid:random:r1}
  \begin{tabular}{l c|c|c}
    \toprule
    \multicolumn{2}{c|}{Accuracies without defenses} &\ma&\ba\\
    \midrule
    1:& Global model \globalModelRound& 59.52 & 8.17\\
    \hline
    2:& Average of benign local models & 35.05 & 14.38\\
    3:& Average of poisoned local models & 34.29 & 82.94\\
    \hline
    4:& \fedavg with benign local models & 36.09 & 37.97\\
    5:& \fedavg with poisoned local models & 21.49 & 98.72\\
    \hline
    6:& \fedavg with all local models & 32.57& 80.85\\
    \bottomrule
    \toprule
    \multicolumn{2}{c|}{Global model accuracies after applying defenses} &\ma&\ba\\
    \midrule
    7:& \naiveBig \clustering & 32.68 & 54.84\\
    8:& \foolsgold~\cite{fung2020FoolsGold} & 30.14 & 87.66\\
    9:& \krum~\cite{blanchard17Krum} & 19.28 & 80.82\\
    10:& \multikrum~\cite{blanchard17Krum} & 10.05 & 99.96\\
    11:& Clip~\cite{mcmahan2018iclrClippingLanguage} & 32.88 & 79.82\\
    12:& Clip\&Noise~\cite{mcmahan2018iclrClippingLanguage} & 25.63 & 88.33\\
    13:& \flame~\cite{nguyen22Flame} & 10.23 & 99.66\\
    14:& T-Mean~\cite{yin2018trimmedMeanMedian} & 33.21 & 76.28\\
    15:& T-Median~\cite{yin2018trimmedMeanMedian} & 21.10 & 62.05\\
    16:& FLTrust~\cite{cao2020fltrust} & 57.27 & \textbf{23.02}\\
    17:& \textbf{\ourname} & 35.08 & \textbf{41.64}\\
  \bottomrule
\end{tabular}
\end{table}

\begin{table}
\fontsize{7pt}{8pt}\selectfont
  \caption{\ma and \ba in the default scenario with \interNonIid based on our \randomgen strategy with a model in \fl round one and scaled
poisoned models regarding the Euclidean distance of updates in percent.}
  \label{tab:acc:noniid:random:r1:scaled}
  \begin{tabular}{l c|c|c}
    \toprule
    \multicolumn{2}{c|}{Accuracies without defenses} &\ma&\ba\\
    \midrule
    1:& Global model \globalModelRound& 59.52 & 8.17\\
    \hline
    2:& Average of benign local models & 35.05 & 14.38\\
    3:& Average of poisoned local models & 34.29 & 82.94\\
    \hline
    4:& \fedavg with benign local models & 36.09 & 37.97\\
    5:& \fedavg with poisoned local models & 17.38 & 99.24\\
    \hline
    6:& \fedavg with all local models & 30.19 & 85.41\\
    \bottomrule
    \toprule
    \multicolumn{2}{c|}{Global model accuracies after applying defenses} &\ma&\ba\\
    \midrule
    7:& \naiveBig \clustering & 31.59 & 56.90\\
    8:& \foolsgold~\cite{fung2020FoolsGold} & 26.93 & 92.20\\
    9:& \krum~\cite{blanchard17Krum} & 24.27 & 38.53\\
    10:& \multikrum~\cite{blanchard17Krum} & 10.00 & 100.00\\
    11:& Clip~\cite{mcmahan2018iclrClippingLanguage} & 30.61 & 84.32\\
    12:& Clip\&Noise~\cite{mcmahan2018iclrClippingLanguage} & 23.47 & 94.51\\
    13:& \flame~\cite{nguyen22Flame} & 17.16 & 36.53\\
    14:& T-Mean~\cite{yin2018trimmedMeanMedian} & 31.89 & 78.20\\
    15:& T-Median~\cite{yin2018trimmedMeanMedian} & 20.38 & 60.84\\
    16:& FLTrust~\cite{cao2020fltrust} & 57.02 & \textbf{23.05}\\
    17:& \textbf{\ourname} & 47.29 & \textbf{15.58}\\
  \bottomrule
\end{tabular}
\end{table}

\begin{table}
\fontsize{7pt}{8pt}\selectfont
  \caption{\ma and \ba in the default scenario with \interNonIid based on our \randomgen strategy with 100 clients in the federation in percent.}
  \label{tab:acc:noniid:100}
  \begin{tabular}{l c|c|c}
    \toprule
    \multicolumn{2}{c|}{Accuracies without defenses} &\ma&\ba\\
    \midrule
    1:& Global model \globalModelRound &59.26 & 9.54\\
    \hline
    2:& Average of benign local models & 33.44 & 11.53\\
    3:& Average of poisoned local models & 34.51 & 83.70\\
    \hline
    4:& \fedavg with benign local models & 40.47 & 15.14\\
    5:& \fedavg with poisoned local models & 37.07 & 88.38\\
    \hline
    6:& \fedavg with all local models & 46.00& 70.58\\
    \bottomrule
    \toprule
    \multicolumn{2}{c|}{Global model accuracies after applying defenses} &\ma&\ba\\
    \midrule
    7:& \naiveBig \clustering & 27.81 & 48.08\\
    8:& \foolsgold~\cite{fung2020FoolsGold} & 51.16 & 74.58\\
    9:& \krum~\cite{blanchard17Krum} & 17.21 &88.17\\
    10:& \multikrum~\cite{blanchard17Krum} & 18.16 & 93.85\\
    11:& Clip~\cite{mcmahan2018iclrClippingLanguage} & 46.16 & 67.88\\
    12:& Clip\&Noise~\cite{mcmahan2018iclrClippingLanguage} & 26.36 & 77.95\\
    13:& \flame~\cite{nguyen22Flame} & 22.38 & 91.66\\
    14:& T-Mean~\cite{yin2018trimmedMeanMedian} & 46.29 & 67.70\\
    15:& T-Median~\cite{yin2018trimmedMeanMedian} & 22.60 & 51.86\\
    16:& FLTrust~\cite{cao2020fltrust} & 47.00 & \textbf{24.26}\\
    17:& \textbf{\ourname} & 40.95 & \textbf{2.00}\\
  \bottomrule
\end{tabular}
\end{table}

\begin{table}
\fontsize{7pt}{8pt}\selectfont
  \caption{\ma and \ba in the default scenario with \interNonIid based on our \randomgen strategy with 100 clients in the federation and scaled
        poisoned models regarding the Euclidean distance of updates in percent.}
  \label{tab:acc:noniid:100:scaled}
  \begin{tabular}{l c|c|c}
    \toprule
    \multicolumn{2}{c|}{Accuracies without defenses} &\ma&\ba\\
    \midrule
    1:& Global model \globalModelRound &59.26 & 9.54\\
    \hline
    2:& Average of benign local models & 33.44 & 11.53\\
    3:& Average of poisoned local models & 34.51 & 83.70\\
    \hline
    4:& \fedavg with benign local models & 40.47 & 15.14\\
    5:& \fedavg with poisoned local models & 21.92 & 95.46\\
    \hline
    6:& \fedavg with all local models & 35.14& 87.10\\
    \bottomrule
    \toprule
    \multicolumn{2}{c|}{Global model accuracies after applying defenses} &\ma&\ba\\
    \midrule
    7:& \naiveBig \clustering & 26.41 & 92.11\\
    8:& \foolsgold~\cite{fung2020FoolsGold} & 37.38 & 91.50\\
    9:& \krum~\cite{blanchard17Krum} & 23.44 & 33.20\\
    10:& \multikrum~\cite{blanchard17Krum} & 19.64 & 73.38\\
    11:& Clip~\cite{mcmahan2018iclrClippingLanguage} & 35.65 & 86.22\\
    12:& Clip\&Noise~\cite{mcmahan2018iclrClippingLanguage} & 25.07 & 95.75\\
    13:& \flame~\cite{nguyen22Flame} & 10.00 & 100.00\\
    14:& T-Mean~\cite{yin2018trimmedMeanMedian} & 41.72 & 76.07\\
    15:& T-Median~\cite{yin2018trimmedMeanMedian} & 20.13 & 54.57\\
    16:& FLTrust~\cite{cao2020fltrust} & 49.16 & \textbf{25.30}\\
    17:& \textbf{\ourname} & 46.70 & \textbf{0.08}\\
  \bottomrule
\end{tabular}
\end{table}

\hyperref[tab:acc:cnn]{\tab\ref{tab:acc:cnn}} and \hyperref[tab:acc:squeeze]{\tab\ref{tab:acc:squeeze}} show the experiments results with a CNN training on \mnist~\cite{mnist} and \squeezenet~\cite{iandola2016squeezenet} training on \cifar\cite{cifar}. The CNN consists of two convolutional layers, the first with 32 output layers, the second with 64 output layers, both applying a kernel size of 5. The output of the convolutional layers traverse a ReLU~\cite{agarap2018deep_relu} and a 2D pooling layer, before being fed into three fully connected layers with output size 512, 256 and 10 output respectively. In both experiments, we used a \mbox{self-pre-trained} model as global model. We can report perfect detection rate with just one \fp for CNN and \squeezenet, even if the backdoor is not yet embedded in the global model. Hence, a stronger adaption by the adversary would strengthen the detection capabilities of \ourname. Other defenses instead can be circumvented by the adaptive adversary.

\begin{table}
\fontsize{7pt}{8pt}\selectfont
  \caption{\ma and \ba in the default scenario with a CNN trained on \mnist~\cite{mnist} with a \pdr of 0.3 in percent.}
  \label{tab:acc:cnn}
  \begin{tabular}{l c|c|c}
    \toprule
    \multicolumn{2}{c|}{Accuracies without defenses} &\ma&\ba\\
    \midrule
    1:& Global model \globalModelRound &76.74 & 2.05\\
    \hline
    2:& Average of benign local models & 84.87 & 0.57\\
    3:& Average of poisoned local models & 60.73 & 39.59\\
    \hline
    4:& \fedavg with benign local models & 86.51 & 0.54\\
    5:& \fedavg with poisoned local models & 63.04 & 37.77\\
    \hline
    6:& \fedavg with all local models & 85.31& 2.35\\
    \bottomrule
    \toprule
    \multicolumn{2}{c|}{Global model accuracies after applying defenses} &\ma&\ba\\
    \midrule
    7:& \naiveBig \clustering & 86.51 & 0.54\\
    8:& \foolsgold~\cite{fung2020FoolsGold} & 85.31 & 2.35\\
    9:& \krum~\cite{blanchard17Krum} & 83.79 & 0.51\\
    10:& \multikrum~\cite{blanchard17Krum} & 86.45 & 0.59\\
    11:& Clip~\cite{mcmahan2018iclrClippingLanguage} & 85.13 & 2.03\\
    12:& Clip\&Noise~\cite{mcmahan2018iclrClippingLanguage} & 84.13 & 2.75\\
    13:& \flame~\cite{nguyen22Flame} & 86.50 & 0.50\\
    14:& T-Mean~\cite{yin2018trimmedMeanMedian} & 85.13 & 2.19\\
    15:& T-Median~\cite{yin2018trimmedMeanMedian} & 85.13 & 2.19\\
    16:& FLTrust~\cite{cao2020fltrust} & 79.26 & 1.90\\
    17:& \textbf{\ourname} & 86.59 & \textbf{0.53}\\
  \bottomrule
\end{tabular}
\end{table}

\begin{table}
\fontsize{7pt}{8pt}\selectfont
  \caption{\ma and \ba in the default scenario with a \squeezenet~\cite{iandola2016squeezenet} trained on \cifar\cite{cifar} in percent.}
  \label{tab:acc:squeeze}
  \begin{tabular}{l c|c|c}
    \toprule
    \multicolumn{2}{c|}{Accuracies without defenses} &\ma&\ba\\
    \midrule
    1:& Global model \globalModelRound &53.06 & 8.3\\
    \hline
    2:& Average of benign local models & 56.04 & 5.67\\
    3:& Average of poisoned local models & 52.21 & 40.03\\
    \hline
    4:& \fedavg with benign local models & 61.03 & 5.82\\
    5:& \fedavg with poisoned local models & 56.33 & 38.85\\
    \hline
    6:& \fedavg with all local models & 60.20& 10.32\\
    \bottomrule
    \toprule
    \multicolumn{2}{c|}{Global model accuracies after applying defenses} &\ma&\ba\\
    \midrule
    7:& \naiveBig \clustering & 61.30 & 5.82\\
    8:& \foolsgold~\cite{fung2020FoolsGold} & 60.21 & 10.32\\
    9:& \krum~\cite{blanchard17Krum} & 55.93 & 5.44\\
    10:& \multikrum~\cite{blanchard17Krum} & 58.75 & 16.17\\
    11:& Clip~\cite{mcmahan2018iclrClippingLanguage} & 60.18 & 10.27\\
    12:& Clip\&Noise~\cite{mcmahan2018iclrClippingLanguage} & 55.24 & 4.73\\
    13:& \flame~\cite{nguyen22Flame} & 60.78 & 5.45\\
    14:& T-Mean~\cite{yin2018trimmedMeanMedian} & 60.15 & 10.04\\
    15:& T-Median~\cite{yin2018trimmedMeanMedian} & 59.68 & 8.40\\
    16:& FLTrust~\cite{cao2020fltrust} & 55.99 & 8.44\\
    17:& \textbf{\ourname} & 60.22 & \textbf{10.80}\\
  \bottomrule
\end{tabular}
\end{table}

\clearpage

\subsection{Setting Independence or \ourname}
\label{app:hyperindependence}
All randomness within the system was seeded with 42 within our experiments, but we conducted spot tests with $seed_{rand} = \{0,1,13\}$ and found similar results, hence, the seed does not influence our findings.

We changed \lr of the default scenario to $\lr = \{0.1, 0.01, 0.001\}$ and found, that 0.01 is the best choice for benign and adversarial training regarding the local and global \ma and \ba, hence a valid choice for our experiments. A \lr or 0.1 is too big destructing the adversarial models to \naive classifiers and reducing the \ma of benign clients to 30\% on average. For LR 0.001, it depends on the round \flround, where it is used. In early rounds, 0.01 is the better choice to speed up the federations training process, but in advanced \fl rounds a lower \lr naturally increases the accuracies, as in every machine learning scenario. Hence, the \ma can be increased, but it is also more difficult for the adversary to adapt some metrics within the defined epochs. Nevertheless, \ourname achieved the same detection \acc with both settings, thus detecting the \naive classifiers for \lr 0.01, which behave similar to a untargeted poisoning attacks, and the models with better accuracies in the 0.001 \lr setting. We set the \lr fixed to 0.01 as a good \mbox{trade-off} between both scenarios.

In all our experiments, we keep the \pmr as high as possible without violating the majority assumption of \hyperref[sec:approach:threatmodel]{\sect\ref{sec:approach:threatmodel}}. Since \ourname does not remove poisoned models with a single test, but prunes different poisonings gradually, we automatically test lower \pmrs within range \mbox{$[0.0, 0.5[$}, demonstrating the independence of \ourname to \pmrs.

Since \ourname does not leverage the plain \ma values, we set \alphasign as low as possible, so that the adversary still achieves a high \ba while simultaneously applying a maximum adaption level. We tested \mbox{$\alphasign = [0.1, 0.2, ..., 0.9]$} and found \mbox{$\alphasign = 0.3$} being the most beneficial choice for \adversary.\footnote{Besides adapting to all \ourname metrics, we conducted experiments starting with only adapting to \metricCos and then adding the other metircs \mbox{step-wise} to find a valid \alphasign, since adapting to all metrics of \ourname simultaneously is not possible in the end.} For higher values, the anomaly to a benign model increases and any defense leveraging the respective metrics detects the attack even clearer, for lower values, the model completely focuses on adapting to metrics and ignores the \ba, thus does not enable the backdoor. Consequently, in parallel to the \ba, the \ma is low having the same effect as an untargeted poisoning attack. In such scenarios an adaption to all metrics of \ourname appears to be very difficult allowing \ourname to be effective. Thus \ourname is independent of \alphasign.

\hyperref[tab:acc:mnist]{\tab\ref{tab:acc:mnist}} and \hyperref[tab:acc:gtsrb]{\tab\ref{tab:acc:gtsrb}} show results for experiments with \mnist~\cite{mnist} and \gtsrb~\cite{gtsrb} respectivelly, showing that \ourname is also effective with varying datasets. \ourname detects the poisoned models with one fp for \mnist and 100\% \acc for \gtsrb even if the backdoor is not yet strong enough to poison the new global model. Further strengthening of the \ba by the adversary would increase the significance within the metrics of \ourname.

\begin{table}
\fontsize{7pt}{8pt}\selectfont
  \caption{\ma and \ba in the default scenario with \mnist~\cite{mnist} as a dataset and \pdr of 0.3 and scaled
poisoned models regarding the Euclidean distance of updates in percent.}
  \label{tab:acc:mnist}
  \begin{tabular}{l c|c|c}
    \toprule
    \multicolumn{2}{c|}{Accuracies without defenses} &\ma&\ba\\
    \midrule
    1:& Global model \globalModelRound &97.60 & 0.43\\
    \hline
    2:& Average of benign local models & 94.30 & 0.40\\
    3:& Average of poisoned local models & 91.73 & 100.00\\
    \hline
    4:& \fedavg with benign local models & 97.22 & 0.45\\
    5:& \fedavg with poisoned local models & 97.20 & 100.00\\
    \hline
    6:& \fedavg with all local models & 97.24& 2.92\\
    \bottomrule
    \toprule
    \multicolumn{2}{c|}{Global model accuracies after applying defenses} &\ma&\ba\\
    \midrule
    7:& \naiveBig \clustering & 97.22 & 0.45\\
    8:& \foolsgold~\cite{fung2020FoolsGold} & 97.22 & 0.45\\
    9:& \krum~\cite{blanchard17Krum} & 95.31 &100.00\\
    10:& \multikrum~\cite{blanchard17Krum} & 97.26 & 46.93\\
    11:& Clip~\cite{mcmahan2018iclrClippingLanguage} & 97.26 & 1.74\\
    12:& Clip\&Noise~\cite{mcmahan2018iclrClippingLanguage} & 86.73 & 48.05\\
    13:& \flame~\cite{nguyen22Flame} & 97.45 & 3.03\\
    14:& T-Mean~\cite{yin2018trimmedMeanMedian} & 97.35 & 1.91\\
    15:& T-Median~\cite{yin2018trimmedMeanMedian} & 96.69 & 2.15\\
    16:& FLTrust~\cite{cao2020fltrust} & 97.34 & \textbf{0.62}\\
    17:& \textbf{\ourname} & 97.18 & \textbf{0.42}\\
  \bottomrule
\end{tabular}
\end{table}

\begin{table}
\fontsize{7pt}{8pt}\selectfont
  \caption{\ma and \ba in the default scenario with \gtsrb~\cite{gtsrb} as a dataset and \pdr of 0.3 and scaled
poisoned models regarding the Euclidean distance of updates in percent.}
  \label{tab:acc:gtsrb}
  \begin{tabular}{l c|c|c}
    \toprule
    \multicolumn{2}{c|}{Accuracies without defenses} &\ma&\ba\\
    \midrule
    1:& Global model \globalModelRound &86.62 & 0.96\\
    \hline
    2:& Average of benign local models & 78.43 & 0.59\\
    3:& Average of poisoned local models & 62.42 & 94.38\\
    \hline
    4:& \fedavg with benign local models & 85.77 & 1.06\\
    5:& \fedavg with poisoned local models & 83.00 & 90.81\\
    \hline
    6:& \fedavg with all local models & 86.19& 8.01\\
    \bottomrule
    \toprule
    \multicolumn{2}{c|}{Global model accuracies after applying defenses} &\ma&\ba\\
    \midrule
    7:& \naiveBig \clustering & 85.57 & 31.65\\
    8:& \foolsgold~\cite{fung2020FoolsGold} & 85.12 & 0.70\\
    9:& \krum~\cite{blanchard17Krum} & 88.57 &1.15\\
    10:& \multikrum~\cite{blanchard17Krum} & 88.08 & 0.92\\
    11:& Clip~\cite{mcmahan2018iclrClippingLanguage} & 86.21 & 4.39\\
    12:& Clip\&Noise~\cite{mcmahan2018iclrClippingLanguage} & 14.98 & 91.72\\
    13:& \flame~\cite{nguyen22Flame} & 84.29 & 13.42\\
    14:& T-Mean~\cite{yin2018trimmedMeanMedian} & 86.30 & 4.26\\
    15:& T-Median~\cite{yin2018trimmedMeanMedian} & 74.22 & 1.61\\
    16:& FLTrust~\cite{cao2020fltrust} & 86.73 & \textbf{1.04}\\
    17:& \textbf{\ourname} & 85.12 & \textbf{0.70}\\
  \bottomrule
\end{tabular}
\end{table}

\begin{figure}[tb]
  \centering
  \includegraphics[width=\linewidth]{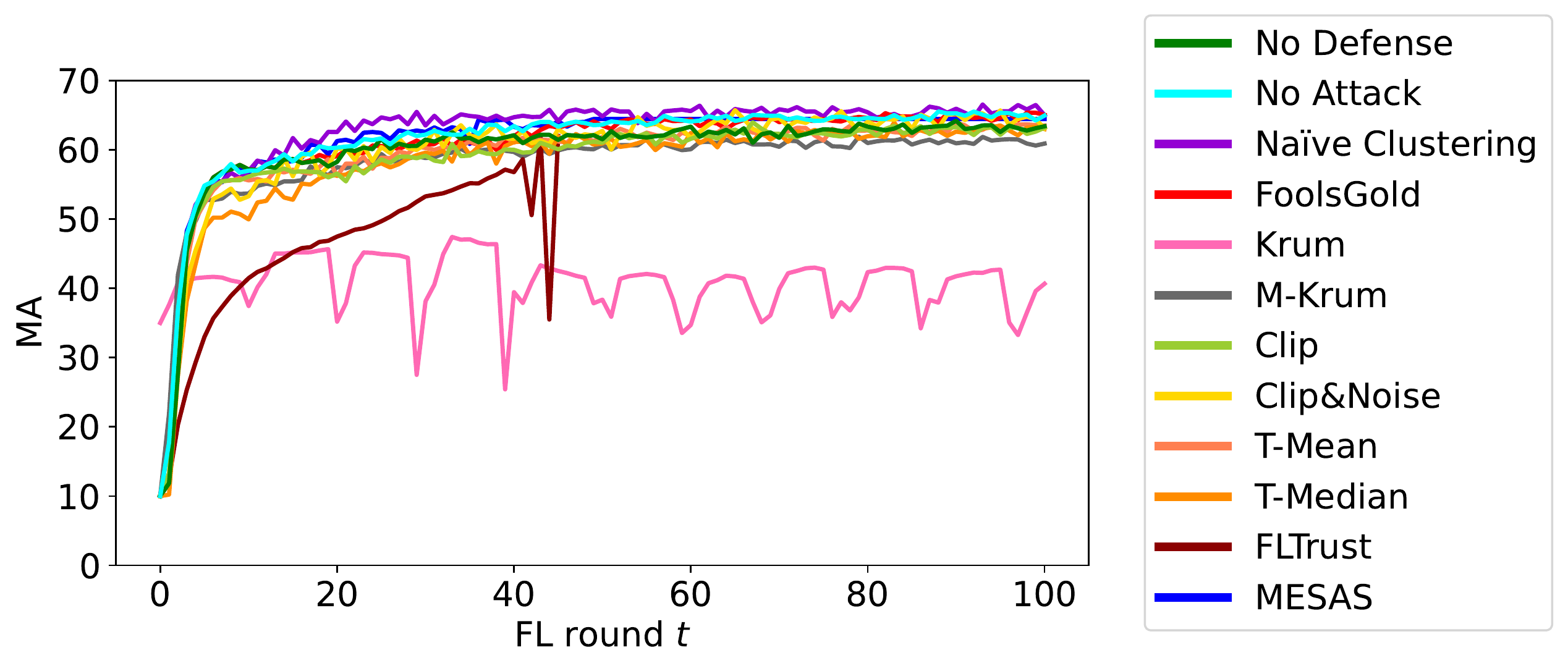}
  \caption{\mas for different defenses over multiple \fl rounds \flround in the default scenario.}
\label{fig:longrun:ma}
\end{figure}

\begin{figure}[tb]
  \centering
  \includegraphics[width=\linewidth]{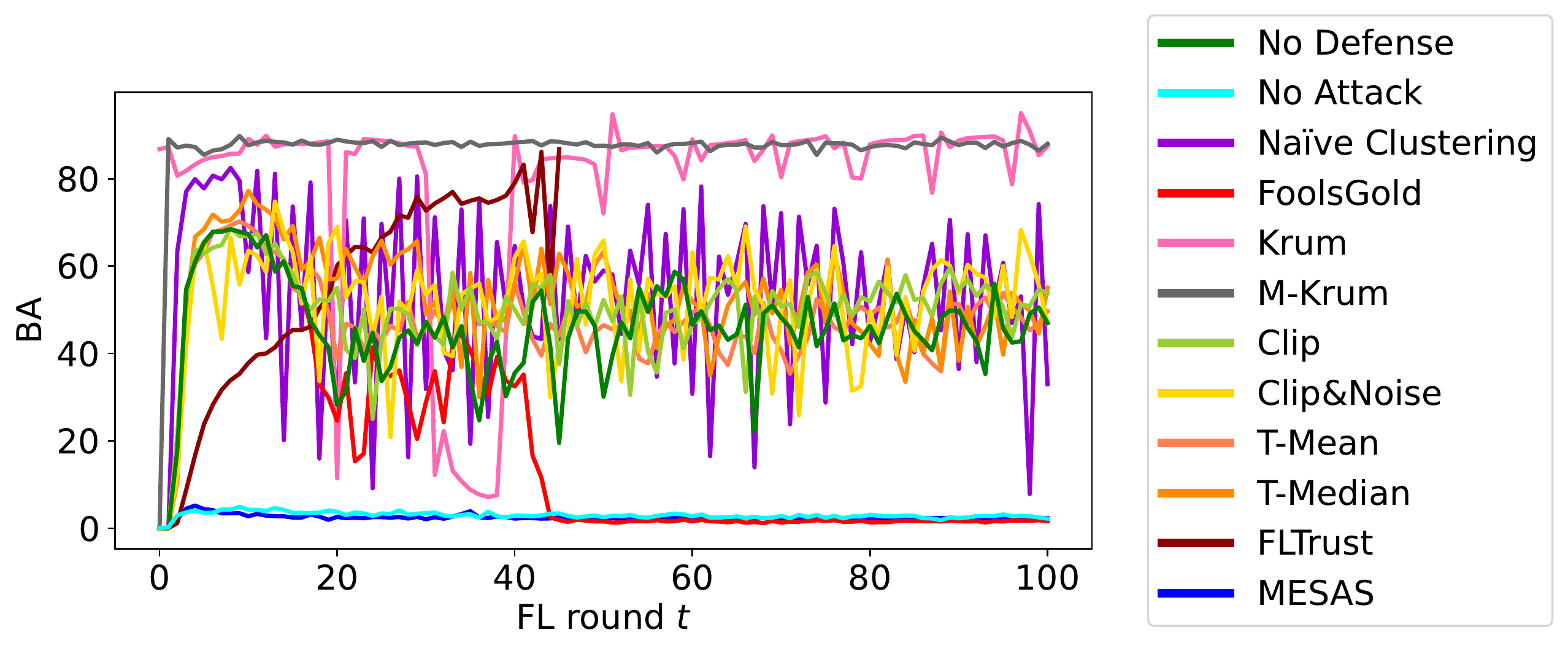}
  \caption{\bas for different defenses over multiple \fl rounds \flround in the default scenario.}
\label{fig:longrun:ba}
\end{figure}

Furthermore, we conducted additional experiments, where models were trained starting from a randomly initialized model for 100 rounds until reaching stability. The performance of the defense mechanisms, along with scenarios encompassing no defense and no attack, was analyzed and depicted in \hyperref[fig:longrun:ma]{\fig\ref{fig:longrun:ma}} and \hyperref[fig:longrun:ba]{\fig\ref{fig:longrun:ba}}.\footnote{We do not report results for \flame~\cite{nguyen22Flame}, since for our setup, the applied noise did influence the model so that the training process stopped at round five. Further, FLTrust is only reported till round 45, since afterward weights of zero are assigned to each update leading to a \naive model.} Notably, the results revealed that \ourname consistently yielded the low \ba values similar to the no-attack scenario, indicating that the backdoor was effectively prevented from being embedded in the final global model under \ourname. In contrast, other defense approaches exhibited relatively higher \bas. Only \foolsgold~\cite{fung2020FoolsGold} could reach low \bas in certain rounds, too. Note, that these \ba values were obtained without the incorporation of adaptation mechanisms. Consequently, with the inclusion of adaptability in defense strategies (\cf~\hyperref[sec:eval:saa:defenses]{\sect\ref{sec:eval:saa:defenses}}), the \bas for adaptive attacks could be further increased. Additionally, \ourname did not compromise the overall \ma of the system, thereby avoiding any significant downsides in its application. The preservation of \ma further underscores the efficacy and advantages of \ourname.

\section{Hyper-Parameters of Experiments}
\label{app:hyperparams}

To provide a detailed and complete overview of our experimental settings, we will list some \mbox{hyper-parameters} of the defenses in the following: For \flame~\cite{nguyen22Flame}, the noising level is set to 0.001, as noted by the authors within the paper. For \tmean~\cite{yin2018trimmedMeanMedian}, we trim the upper and lower 5\% to get rid of outliers. The noise level of our differential privacy defense was set to 0.01. The threshold for both, \krum and \multikrum~\cite{blanchard17Krum} was set to 0.7 and the rate of clients considered for \multikrum is 0.3.

\section{Assignment of \nonIidBig Distributions}
\label{app:distributions}

\subsection{\intraBigNonIid}
\label{app:distributions:intra}
\nonIidOneClass assigns one main label class to the client, which has more samples than the remaining classes. To construct such scenarios, all labels in the clients dataset including the main label are first assigned equal sample frequencies. Then, the \textit{\nonIid rate \noniidsignFormular} controls how many samples removed from all classes equally and reassigned to the main label to create a focus on this class. For \ensuremath{\noniidsign = 0} all samples are uniformly distributed, hence an \iid setting is created. For \ensuremath{\noniidsign = 1} only samples from the main label are contained in the dataset. An example of \nonIidOneClass is visualized in the classic \nonIid scenario in \hyperref[fig:noniid]{\fig\ref{fig:noniid}}. \nonIidTwoClass works like \nonIidOneClass, but assigns two main labels simultaneously.
Distribution \nonIid defines the sampling frequency for each label with respect to a distribution. We leverage Dirichlet~\cite{minka2000dirichlet} and normal distribution and assign the biggest value to the main label.

\subsection{\interBigNonIid}
\label{app:distributions:inter}
We generate \interNonIid datasets by assigning arbitrary datasets to clients. The \textit{\randomgen} strategy first randomly decides for each label if it is contained in client's local dataset by coin flip. Afterwards, we randomly generate a number between zero and one for each label that should be contained in the dataset. Then, we sum those random values and assign the relative percentage of the sum to the each label. Finally, those values can be converted to real sample frequencies by multiplying the percentage with the desired overall sample count of the client. This results in \interNonIid datasets even with disjoint data. The sample distribution of the setup within this paper is listed in \hyperref[tab:distribution:random]{\tab\ref{tab:distribution:random}}.

Certainly, it is also possible to leverage different \intraNonIid for each client's dataset to generate \interNonIid scenarios, if one needs more control over the distributions.

\begin{table}
\fontsize{7pt}{8pt}\selectfont
  \caption{Sample frequencies for each label in the clients' datasets for our \randomgen strategy.}
  \label{tab:distribution:random}
  \begin{tabular}{c|c|c|c|c|c|c|c|c|c|c}
    \toprule
    \multirow{2}{*}{Client} &\multicolumn{10}{c}{Label} \\
    &0 & 1&2 &3 &4 &5 &6 &7 &8 &9\\
    \midrule
    0 & 598 & 0 & 325 & 0 & 259 & 404 & 511 & 463 & 0 & 0\\
    1 & 0 & 777 & 0 & 494 & 0 & 623 & 666 & 0 & 0 & 0\\
    2 & 0 & 0 & 919 & 433 & 0 & 0 & 770 & 438 & 0 & 0\\
    3 & 0 & 745 & 0 & 1344 & 392 & 0 & 0 & 0 & 0 & 79\\
    4 & 355 & 95 & 0 & 232 & 814 & 0 & 0 & 0 & 683 & 381\\
    5 & 0 & 203 & 543 & 0 & 0 & 599 & 308 & 400 & 507 & 0\\
    6 & 295 & 0 & 827 & 0 & 0 & 0 & 1438 & 0 & 0 & 0\\
    7 & 0 & 1116 & 84 & 0 & 0 & 0 & 0 & 1360 & 0 & 0\\
    8 & 408 & 454 & 30 & 0 & 0 & 0 & 279 & 518 & 538 & 333\\
    9 & 0 & 431 & 271 & 0 & 0 & 206 & 788 & 36 & 0 & 828\\
    10 & 715 & 113 & 431 & 0 & 0 & 508 & 0 & 0 & 476 & 317\\
    \hline
    11 & 560 & 424 & 369 & 0 & 343 & 0 & 406 & 89 & 270 & 99\\
    12 & 99 & 2461 & 0 & 0 & 0 & 0 & 0 & 0 & 0 & 0\\
    13 & 0 & 0 & 595 & 257 & 172 & 0 & 568 & 206 & 527 & 235\\
    14 & 0 & 0 & 0 & 2047 & 0 & 0 & 0 & 0 & 513 & 0\\
    15 & 159 & 149 & 199 & 546 & 642 & 0 & 447 & 0 & 404 & 14\\
    16 & 494 & 254 & 486 & 388 & 0 & 523 & 0 & 0 & 0 & 415\\
    17 & 0 & 0 & 315 & 947 & 0 & 0 & 963 & 209 & 0 & 126\\
    18 & 0 & 0 & 0 & 271 & 549 & 509 & 0 & 640 & 0 & 591\\
    19 & 0 & 178 & 0 & 677 & 0 & 0 & 588 & 285 & 832 & 0\\
  \bottomrule
\end{tabular}
\end{table}

\end{document}